\documentclass[journal]{IEEEtran}
\usepackage{lineno,hyperref}
\usepackage{graphicx, epstopdf}
\usepackage{amsmath,amsthm}
\usepackage{array}
\usepackage{epsfig}
\usepackage{amssymb}
\usepackage{algorithm}
\usepackage{algorithmic}
\usepackage{subfigure,rotating}
\usepackage{multirow}
\usepackage{setspace}
\usepackage{booktabs}
\usepackage[table*]{xcolor}
\usepackage{bm}
\xdefinecolor{gray95}{gray}{0.55}
\xdefinecolor{gray25}{gray}{0.8}
\usepackage{colortbl}
\usepackage[utf8]{inputenc}
\begin{document}

\title{Learning from Non-Stationary Stream Data in Multiobjective Evolutionary Algorithm}

\author{Jianyong Sun, Hu Zhang, Aimin Zhou and Qingfu Zhang\thanks{J. Sun is with the School of Computer Science and Electrical Engineering, University of Essex, Colchester CO4 3SQ, U.K. (e-mail: jysun@essex.ac.uk).}
\thanks{H. Zhang is with the Center for Control Theory and Guidance Technology, Harbin Institute of Technology, Harbin, China. (e-mail: jxzhanghu@126.com).}
\thanks{A. Zhou is with the Shanghai Key Laboratory of Multidimensional Information Processing, East China Normal University, 3663 North Zhongshan Road, Shanghai, 200062, China, and the Department of Computer Science and Technology, East China Normal University, 3663 North Zhongshan Road, Shanghai, 200062, China. (e-mail: amzhou@cs.ecnu.edu.cn)}\thanks{Q. Zhang is with the Department of Computer Science, City University of Hong Kong, Hong Kong, and the City University of Hong Kong Shenzhen Research Institute, Shenzhen, 5180057, China.}
\thanks{J. Sun and H. Zhang contributed equally to this work.}}

\maketitle

%misused these machine learning techniques in the sense that the basic i.i.d. assumption, which states that data to be modelled and analysed should be independent and identically distributed, of these techniques is not well checked and actually violated. Specifically, solutions in the population cannot be considered as sampled from the same distribution,

\begin{abstract}
Evolutionary algorithms (EAs) have been well acknowledged as a promising paradigm for solving optimisation problems with multiple conflicting objectives in the sense that they are able to locate a set of diverse approximations of Pareto optimal solutions in a single run. EAs drive the search for approximated solutions through maintaining a diverse population of solutions and by recombining promising solutions selected from the population. Combining machine learning techniques has shown great potentials since the intrinsic structure of the Pareto optimal solutions of an multiobjective optimisation problem can be learned and used to guide for effective recombination. However, existing multiobjective EAs (MOEAs) based on structure learning spend too much computational resources on learning. To address this problem, we propose to use an online learning scheme. Based on the fact that offsprings along evolution are streamy, dependent and non-stationary (which implies that the intrinsic structure, if any, is temporal and scale-variant), an online agglomerative clustering algorithm is applied to adaptively discover the intrinsic structure of the Pareto optimal solution set; and to guide effective offspring recombination. Experimental results have shown significant improvement over five state-of-the-art MOEAs on a set of well-known benchmark problems with complicated Pareto sets and complex Pareto fronts.
\end{abstract}

\begin{IEEEkeywords}
Multiobjective Optimization $|$ Evolutionary Algorithms $|$ Machine Learning $|$ Online Agglomerative Clustering
\end{IEEEkeywords}

\IEEEpeerreviewmaketitle

\section{Introduction}

\IEEEPARstart{I}{n} practice, a decision maker often requires to consider optimising multiple conflicting objectives. This type of optimisation problems are usually referred to as {\em multiobjective optimisation problems (MOPs)}. Since the objectives of the problems usually conflict with each other, there does not exist a unique solution that can optimise all the objectives simultaneously. Therefore, a set of Pareto optimal solutions, named as \emph{Pareto set (PS)}, exists for an MOP~\cite{1999Miettinen}.
%The goal \textcolor[rgb]{1.00,0.00,0.00}{of solving a MOP} is to determine a set of values of model parameters or state variables that provide the best tradeoff (or Pareto optimal) solutions among multiple predefined cost or objective functions.
A solution is considered to be `Pareto optimal' if it is impossible to make any one objective better off without making at least another one worse off. Finding the PS often challenges greatly on computational capacity and algorithm intelligence~\cite{2001Deb}.

In the last three decades, extensive research on \emph{evolutionary algorithms (EAs)} have shown that the EA paradigm is very powerful in handling MOPs, in the sense that a set of solutions that approximates to the PS, named as approximated set, can be obtained in a single run without requiring much computational effort~\cite{2011Zhou}\cite{zhang2015self}. EAs simulate the genetic evolution of a population of individuals to best fit their living environment~\cite{yu2010introduction}. To design an effective EA, effective recombination for fit offspring generation is a key. Research has shown that a problem's domain knowledge, if any, can greatly improve the search efficiency if the knowledge is properly collected or learned during the search process~\cite{2008ZhangZJ}.

For an $m$-objective optimisation problem, it has been proved that the distribution of the PS exhibits an ($m-1$)-dimensional manifold structure under mild conditions~\cite{regularity}. This property is often referred to as the regularity property. From the point view of EA design, an effective EA is expected if the manifold structure can be discovered and applied for offspring generation. Some EAs have been developed to combine {\em machine learning} techniques for the discovery of the intrinsic manifold structure to aid the search for the PS. For examples, in regularity model-based estimation of distribution algorithm (RM-MEDA)~\cite{2008ZhangZJ}, the local principal component analysis (local PCA) approach is applied at each generation. It uses the learned principal components to approximate the manifold structure. Some EAs adopted other machine learning techniques to approximate the manifold structure~\cite{wang2012regularity}. All these algorithms apply the machine learning techniques at {\em every generation}. These learning algorithms often need to visit all data several times (iterations) until converge. Thus, a considerable amount of computational resources is consumed on learning.

To reduce the computational overhead, the \emph{multiobjective EA (MOEA)} proposed by Zhang et al.~\cite{zhang2015self} couples the population evolution and the model inference. In their MOEA, only one iteration of the learning algorithm is applied at each generation. This scheme provides an important development on saving computational resources. The evolution procedure can also be seen as a learning procedure; intrinsic PS's structure of an MOP is expected to be learned dynamically from the changing candidate solutions. However, there is a fundamental issue in this scheme. As well known, one of the main assumptions in machine learning is that sample observations are assumed to be effectively \emph{i.i.d. (independent and identically distributed)} for the purposes of statistical inference. But, under the scheme in~\cite{zhang2015self}, along the evolution procedure, the assumption is largely violated. First, solutions at adjacent generations have rather different qualities in terms of their respective objectives, which indicate that they might not be sampled from the same underlying distribution (i.e. these solutions are not identically distributed). Second, the generation of new solutions at present generation depends on collective information from previous generation, which indicates solutions at adjacent generations are not independent. %In general, we may conclude that the basic i.i.d. assumption is largely violated in the context of learning model based MOEAs.

%existing learning-based EAs apply some machine learning techniques rather arbitrarily without checking whether the input data are i.i.d. or not.
%offspring are created either by sampling from the learned model at previous generation or using some recombination operators.

%Apparently our purpose of employing learning techniques are to reduce computational cost
Look deeply into the data (i.e. offsprings created during the evolution search) we try to learn from, some special characteristics can be observed: 1) the structure to be discovered along evolution is temporal and changing dynamically. In other words, these data are produced by a non-stationary process\footnote{A process is stationary if and only if the joint distribution of the data at different time are the same. Specifically, if we let $t = 1,\cdots$ be the generations of the evolution, and $\mathbf{y}_t$ be a $n$-dimensional solution. The sequence $\mathbf{y}_t$ is a stationary stochastic process if the joint probabilistic distribution of $(\mathbf{y}_{t_1+h}, \cdots, \mathbf{y}_{t_N+h})$ and $(\mathbf{y}_{t_1}, \cdots, \mathbf{y}_{t_N})$ are the same for all $h = 0, 1, \cdots, $ and an arbitrary selection of $t_1, \cdots, t_N$. This is obviously not the case for the stream of offsprings created during the evolution process.}; 2) the structure determined by the data is scale-variant. On a short time scale, the structure is pseudo-stationary, while on a long time scale, the structure has a sequential and converging property. That is, along the evolution process, the underlying structure is similar between adjacent generations, while the structure will finally be converging to the PS's manifold structure of the considered optimisation problem.

In this paper, we present the first-ever MOEA based on an online machine learning\footnote{In computer science, online machine learning methods learn patterns from data which are available in a sequential order as opposed to batch learning techniques which generate the best predictor by learning on the entire training data set at once.} from a stream of non-stationary data. In our algorithm, a modified algorithm to the online agglomerative clustering algorithm presented in~\cite{guedalia1999line} is developed to learn the PS's structure addressing the above mentioned characteristics. Obvious advantages of the proposed \emph{online clustering based evolutionary algorithm (OCEA)} include 1) a perfect match between the search dynamics and the non-stationary structure learning and 2) a significantly reduced computational cost on learning (data need to be visited only once in the context of online learning). To successfully implement the proposed algorithm, we need to address three main issues. First, how to modify the online agglomerative clustering in accordance with the evolution process to discover the underlying structure? Second, how to properly use the learned structure to create offsprings effectively? Finally, how to select the fittest individuals to drive the search towards the PS? These issues will be discussed in the following sections.

The rest of the paper is organised as follows. The background and previous work on multiobjective evolutionary algorithms is introduced in Section~\ref{background}. Section~\ref{algorithm} presents the proposed algorithm in detail. Experimental studies are shown in Section~\ref{expStu} and \ref{furtherDis}. The analysis of parameters effect to algorithmic performance is discussed in Section~\ref{senStudy}. Section~\ref{conclusion} concludes the paper.

\section{Background and Previous Work}\label{background}

A box-constrained continuous MOP can be stated as follows:
\begin{equation}
\begin{array}{ll}
\min & \mathbf{F}(\mathbf{x})=(f_1(\mathbf{x}),\cdots,f_m(\mathbf{x}))^\intercal\\
\mathrm{s.t. } & \mathbf{x}=(x_1,\cdots,x_n)^\intercal \in \Omega
\end{array}
\label{eq.1}
\end{equation}where $\Omega=\prod_{i=1}^n[a_i, b_i] \subseteq \mathbb{R}^n$ defines the decision (search) space; $a_i$ and $b_i$ are the lower and upper boundaries of variable $x_i$, respectively; $\mathbf{x}=(x_1, \cdots, x_n)^\intercal $ is a vector of decision variable; $\mathbf{F}: \Omega \to \mathbb{R}^{m}$ represents the mapping from search space to objective space where $m$ objective functions $f_i(\mathbf{x}), i=1,\ldots,m$ are to be considered.

Suppose that $\mathbf{u}=(u_1,\cdots,u_m)^\intercal, \mathbf{v}=(v_1,\cdots,v_m)^\intercal \in \mathbb{R}^m$ are two vectors. If $u_i \leq v_i$ for all $i\in \{1,\cdots, m\}$, but there exists at least one index $j$, such that $u_j < v_j$, then $\mathbf{u}$ is said to dominate $\mathbf{v}$\footnote{The definition of domination is for minimization. ``Dominate" means ``be better than".}, denoted by $\mathbf{u}\prec \mathbf{v}$. A solution $\mathbf{x}^*\in \Omega$ is called (\emph{globally}) \emph{Pareto optimal} if there is no $\mathbf{x}\in \Omega$ such that $\mathbf{F}(\mathbf{x}) \prec \mathbf{F}(\mathbf{x^*})$. The set of all Pareto optimal solutions, denoted by \emph{PS}, are named as \emph{Pareto set}. The set of the objective vectors of the Pareto optimal solutions is called \emph{Pareto front}, denoted by \emph{PF}. The goal of an MOEA for an MOP is to find a set of approximated solutions whose objective vectors (the objective vectors constitute an approximated front) are as close to the PF as possible (i.e. the convergence requirement), and distribute along the PF as widely and evenly as possible (i.e. the  diversity requirement).

Great efforts have been made to deal with MOPs in the evolutionary computation community~\cite{2011Zhou}. These developed approaches focus either on establishing a mechanism to balance convergence and diversity, or on developing effective recombination.

MOEAs concerning the balance between convergence and diversity basically fall into three categories. In the first category, the Pareto dominance relationship is applied for promising solution selection. The nondominated sorting developed by Deb et al.~\cite{2002DebPAM} is the most known method. Its primary use is to drive the search towards the PF which favours convergence. It needs to incorporate other strategies, such as crowding distance~\cite{2002DebPAM} and K-nearest neighbor method~\cite{zitzler2001spea2}, to preserve the population diversity. It has been found out that dominance-based sorting method is not able to provide enough comparability for many-objective ($\geq 4$ objectives) optimization problems. Typical dominance-based MOEAs include NSGA-II~\cite{2002DebPAM}, SPEA2~\cite{zitzler2001spea2}, PESA-II~\cite{corne2001pesa}, NSGA-III~\cite{deb2014evolutionary}, and others.

In the second category, MOEAs based on performance metrics, such as \emph{hypervolume (HV)}, R2 and $\Delta_p$, were developed. The performance metrics embed the convergence and diversity requirements together so that they can be employed to directly guide the selection of solutions for a good balance of convergence and diversity. Representative MOEAs include SMS-EMOA~\cite{beume2007sms}, HyPE~\cite{2011BaderZ}, R2-IBEA~\cite{phan2013r2} and DDE~\cite{rodriguez2012new}. The computation of the performance metrics becomes much more difficult and time-consuming in dealing with many-objective optimisation problems.

The third category is the decomposition-based MOEAs. In this category, a number of reference vectors in the objective space are used to decompose the problem into a set of single objective subproblems~\cite{2009LiZ}, or several simple multiobjective subproblems~\cite{Liu2013}. The convergence is controlled by the objective values of the subproblems; while the diversity is managed by computing the distances of the solutions to the reference vectors. Representative decomposition-based MOEAs include MOEA/D~\cite{2007ZhangL}, MOEA/D-DE~\cite{2009LiZ}, MOEA/D-STM~\cite{li2014stable}, MOEA/D-M2M~\cite{Liu2013} and others.

Regarding MOEAs focusing on effective recombination, they are almost all designed based on the regularity property of MOPs. The underlying assumption is that the manifold structure could be used to greatly improve the search efficiency since high-quality offsprings can be generated if the regularity structure is properly modelled and learned. % The balance between convergence and diversity is usually assured by existing environmental selection methods, such as non-dominated sorting, hypervolume indicator-based method, etc.
The first work on applying the regularity property in designing MOEA, i.e., aforementioned RM-MEDA, was proposed in 2008~\cite{2008ZhangZJ}, where the manifold structure is approximated by the first $(m-1)$ principal components. This work was improved later by using help from the modelling on the PF~\cite{zhou2009approximating}. Various regularity based MOEAs have been developed since then, such as a reducing redundant cluster based RM-MEDA~\cite{wang2012regularity}, a RM-MEDA with local learning strategy~\cite{li2013improved}, evolutionary multiobjective optimisation via manifold learning~\cite{Li2014learning}, and others. Moreover, in~\cite{zhang2015self}, a self-organising map method is incorporated within the evolution procedure to search for the manifold PS structure.

%\textcolor{blue}{Include more literature on regularity property based MOEAs.}

\section{The Algorithm}\label{algorithm}

%It is hoped that the projection of the approximated set, called approximated front, found by the evolutionary search in a single run converges to the PF well, and distributes along the whole PF evenly. One of the important issues in designing a MOEA is on how to create promising individuals, which is not only critical for search effectiveness but also for efficiency. To address this issue, we propose to incorporate the regularity property to guide offsprings generation.

%According to the regularity property, the PS exhibits a manifold structure. It is expected that a manifold structure will be emerging along the evolution process, and be finally converging to the PS.

%take the selected solutions as the training data at each generation. Data points are usually required to be visited a number of times to make sure the convergence of the learning algorithms. This is to account for a reliable establishment of the learned structure of the MOP. Apparently, this will result in a high computational cost.

As discussed previously, existing regularity based MOEAs usually spend a high computational cost on learning. To reduce the consumption of computational resources, we propose to adopt an online machine learning scheme. Offsprings are considered as a stream of data since they come in order along the evolution process, and can only be accessed once or a small number of generations. Moreover, it is observed that along the evolution process, the stream of solutions is dependent, and non-stationary. Therefore, the application of online learning algorithm is able to reduce the number of visits and account for the non-stationary nature. This can significantly reduce the computational resources.

Note that a finite mixture of Gaussian clusters can be used to well approximate the distribution of a set of data points statistically.\footnote{It is well acknowledged that mixtures of Gaussian distributions are dense in the set of probability distributions with respect to weak topology~\cite{bacharoglou10}.} This motives us to approximate the manifold structure by using an online {\em clustering} algorithm. The cluster statistics, including the number of clusters, cluster mean and variance-covariance, will evolve over time. To model this non-stationary process, we propose to modify an online agglomerative clustering algorithm called AddC~\cite{guedalia1999line} and use it to dynamically estimate the cluster statistics. %The incorporation of AddC into MOEA is summ an algorithmic framework, named as \emph{online clustering based evolutionary algorithm (OCEA)}. The details of OCEA are as follows.

In the following, we first describe the online agglomerative clustering algorithm developed in~\cite{guedalia1999line} and discuss how it should be modified to adapt to the evolution process of MOEAs. The other details of the developed algorithm are then presented.

\subsection{Online Agglomerative Clustering}

AddC, presented by Guedalia et al.~\cite{guedalia1999line} in 1998, is developed for clustering a stream of non-stationary data. AddC's clustering procedures are shown in Alg.~\ref{online_clustering}. From line~\ref{oac.1} to \ref{oac.2}, an arriving new data point $\mathbf{y}$ is assigned to the cluster that is closest to it at first. This step attempts to minimise the within cluster variance. Afterwards, from line~\ref{oac.3} to~\ref{oac.6}, if there are less than $K_{\max}$ clusters, $\mathbf{y}$ is employed as a centroid to create a new cluster; otherwise, from line~\ref{oac.4} to~\ref{oac.6}, two redundant clusters which are closest to each other are merged, and $\mathbf{y}$ is also treated as a centroid to create a new cluster for replacing the redundant cluster (i.e. $\mathcal{C}^\delta$ in line~\ref{oac.6}). The merging operation is aimed to maximise the distances between the centroids and to remove redundant clusters. The creation of new clusters is to consider the temporal changes in the distribution of the data. In line \ref{oac.7}, if there still exist data points to be clustered, the clustering operations are repeated. Otherwise, a post process is conducted to remove clusters with a negligible number ($\epsilon$) of data in line \ref{oac.8}. The post process is to eliminate outliers if any.

\begin{algorithm}[htbp]
\caption{Online Agglomerative Clustering AddC}\label{online_clustering}
\begin{algorithmic}[1]
\REQUIRE  an arriving new data point $\mathbf{y}$, centroids $\mathbf{z}^k$ and counters $c^k$ of $m$ existing clusters $\mathcal{C}^1, \cdots, \mathcal{C}^m$, $1\leq k \leq m$, and the maximum number of clusters allowed $K_{\max}$.
\ENSURE a new set of clusters.
\STATE The centroid which is closest to the data point $\mathbf{y}$ is defined as the winner, \[j = \arg\min\limits_{1\leq k \leq m} ||\mathbf{y} - \mathbf{z}^k||.\]\label{oac.1}\\
\STATE Update the closest centroid and its counter, \[\mathbf{z}^j = \mathbf{z}^j +\frac{\mathbf{y} - \mathbf{z}^j}{c^j};~c^j = c^j + 1,\]where $c^j$ is the number of data points in $\mathcal{C}^j$.\label{oac.2}\\
\STATE If $m < K_{\max}$, set $m = m + 1$ and $\delta = m$. Goto step~\ref{oac.6}.\label{oac.3}\\
\STATE Find a pair of closest (redundant) centroids, \[(\gamma, \delta) = \arg\min\limits_{\gamma, \delta, \gamma \neq \delta} ||\mathbf{z}^\gamma -\mathbf{z}^\delta||.\]\label{oac.4}\\
\STATE Merge redundant clusters and update the cluster statistics, \[\mathbf{z}^\gamma = \frac{\mathbf{z}^\gamma c^\gamma + \mathbf{z}^\delta c^\delta}{c^\gamma + c^\delta};~c^\gamma = c^\gamma + c^\delta.\]\label{oac.5}\\
\STATE Initialise a new cluster $\mathcal{C}^\delta$, $\mathbf{z}^\delta = \mathbf{y}$ and $c^\delta = 0$.\label{oac.6}\\
\STATE If there still exist data points to be clustered, take a new point $\mathbf{y}$ and goto Step~\ref{oac.1}.\label{oac.7}\\
\STATE Post process: $\forall k$, if $c^k < \epsilon$, perform steps \ref{oac.6} and \ref{oac.7}.\label{oac.8}\\
\end{algorithmic}
\end{algorithm}

\subsection{Algorithmic Framework}

The framework of OCEA is presented in Alg.~\ref{framework}. In line~\ref{fw.1} to~\ref{fw.2}, an initial population $\cal P$ is yielded, an external archive $\mathcal{A}$ is initialised to be the same as $\cal P$. In the first generation, each solution is considered as a cluster where itself is initialised to be the centroid $\mathbf{z}^i = \mathbf{x}^i$ and counter $c^i = 1$, $i=1,\cdots,N$. Afterwards, at each generation, an offspring $\mathbf{y}^i$ is generated around each solution $\mathbf{x}^i$ (lines~\ref{fw.7} to \ref{fw.9}). To generate $\mathbf{y}^i$, a mating control parameter $\beta \in [0,1]$ is applied to balance exploration and exploitation. With $\beta$, the solution generation will be in favour of exploitation. That is, the reference (or parent) solutions are chosen from the cluster that $\mathbf{x}^i$ locates. With $1-\beta$, the reference individuals are chosen from the global mating pool specified in line~\ref{fw.5}. This is to favour exploration. After recombination, the generated offspring $\mathbf{y}^i$ is then used to update external archive and current population by environmental selection, and the clustering information (lines~\ref{fw.9} and~\ref{fw.10}).

The solution generation and the updating procedures for population and clusters will be described in the following subsections.

\begin{algorithm}[htbp]
\caption{OCEA framework}\label{framework}
\begin{algorithmic}[1]
\REQUIRE population size $N$, maximum evolutionary generations $T$, mating control parameter $\beta$.
\ENSURE population $\cal P$.
\STATE Intialization $\mathcal{P} = \{ {\mathbf{x}^1}, \cdots ,{\mathbf{x}^N}\}$ and an external archive $\mathcal{A} = \mathcal{P}$.\label{fw.1}\\
\STATE Take each $\mathbf{x}^i \in \mathcal{P}$ as a cluster ${\cal C}^i$ with centroid $\mathbf{z}^i = \mathbf{x}^i$ and counter $c^i = 1$.\label{fw.2}\\
\FOR{$t\leftarrow 1$ to $T$}\label{fw.3}
\STATE Set $m = $\#clusters.\label{fw.4}\\
\STATE Construct a global mating pool $\mathcal{M}$ by randomly choosing a solution from a $\mathcal{C}^i, 1\leq i \leq m$.\label{fw.5}\\
\FOR{$i \leftarrow 1$ to $N$\label{fw.6}}
\STATE Construct a mating pool $\mathcal{Q}^i$ for each $\mathbf{x}^i$ as follows: \begin{equation}\mathcal{Q}^i = \left\{ {\begin{array}{ll}
{\mathcal{C}^{k_i}}&{\mbox{if}{\kern 1pt} {\kern 1pt} rand() < \beta }\\
{\mathcal{M}}&{\mbox{otherwise}}\nonumber
\end{array}} \right.,\end{equation}where $\mathcal{C}^{k_i}$ represents that $\mathbf{x}^i$ loactes in $\mathcal{C}^k$, $rand()$ is a random number generator in $[0,1]$.\label{fw.7}\\
\STATE Generate ${\mathbf{y}^i}$ = \textsc{SolGen} $({\mathcal{Q}^i},{\mathbf{x}^i})$.\label{fw.8}\\
\STATE Update and clustering $[\mathcal{A}, \mathcal{C}]$ = \textsc{Esoc}~$(\mathcal{A}, \mathbf{y}^i, \mathcal{C})$.\label{fw.9}\\
\ENDFOR
\STATE Set $\mathcal{P}=\mathcal{A}$ and pass the clustering results of $\mathcal{A}$ to $\mathcal{P}$.\label{fw.10}\\
\ENDFOR
\end{algorithmic}
\end{algorithm}

\subsection{New Solution Generation}

In this paper, the \emph{differential evolution (DE)} and \emph{polynomial mutation (PM)} operators are adopted to generate offsprings as presented in Alg.~\ref{sg}. The recombination operator takes the current solution $\mathbf{x}$ and its mating pool $\mathcal{Q}$ as input and outputs an offspring $\mathbf{y}$. DE~\cite{price2006differential} is firstly used to generate a trial solution (line~\ref{sg.2}), a repair mechanism is employed to correct any component that is outside the search boundary of that component (line~\ref{sg.3}). After repair, the PM~\cite{2001Deb} operator is applied to generate a new solution (line~\ref{sg.4}). The new solution is repaired again if necessary and the final solution is returned (line~\ref{sg.5}).

In Alg.~\ref{sg}, $F$ and $CR$ are the two control parameters for the DE operator, $p_m$ and $\eta_m$ are the parameters for the PM operator. If $CR=1$, the DE operator in Alg.~\ref{sg} is rotation invariant, which is of advantage to deal with complicated PS~\cite{2009LiZ}. Therefore DE is selected to generate new offsprings in OCEA. Obviously, the use of other recombination operators is not limited; e.g. we could use the recombination operators in~\cite{XQiu2015}.

\begin{algorithm}[htbp]
\caption{Solution generation (\textsc{SolGen}) operator}\label{sg}
\begin{algorithmic}[1]
\REQUIRE a current solution $\mathbf{x}$ and its mating pool $\cal Q$
\ENSURE a trial solution $\mathbf{y}$
\STATE Choose randomly two distinct parent individuals $\mathbf{x}^1$ and $\mathbf{x}^2$ from $\mathcal{Q}$\label{sg.1}\\
\STATE Generate $\mathbf{y}^{'}=(y^{'}_1,\cdots,y^{'}_n)^\intercal$ as follows:
\[
y^{'}_i = \left \{
\begin{array}{ll}
x_i+F \times (x_i^{1}-x_i^{2})&\text{if}~rand()\leq CR\\
x_i&\text{otherwise}\\
\end{array}
\right..
\]\label{sg.2}\\
\STATE Repair $\mathbf{y}^{'}$,
\[
y^{''}_i = \left \{
\begin{array}{ll}
a_i & \text{if}~y^{'}_i < a_i\\
b_i & \text{if}~y^{'}_i > b_i\\
y^{'}_i & \text{otherwise}
\end{array}
\right.,
\]where $\mathbf{x}_i \in [a_i, b_i]$.\label{sg.3}\\
\STATE Mutate $\mathbf{y}^{''}$,
\[
y_i = \left \{
\begin{array}{ll}
y^{''}_i+\delta_i \times (b_i-a_i) & \text{if}~rand()< p_m\\
y^{''}_i & \text{otherwise}
\end{array}
\right.,
\]where $r=rand()$ if a uniform random generator in [0,1], and{\small \[
\delta_i = \left \{
\begin{array}{ll}
\left[2r+(1-2r)(\frac{b_i-y_i^{''}}{b_i-a_i})^{\eta_m+1}\right]^{\frac{1}{\eta_m+1}}-1 & \text{if}~r<0.5,\\
1-\left[2-2r+(2r-1)(\frac{y_i^{''}-a_i}{b_i-a_i})^{\eta_m+1}\right]^{\frac{1}{\eta_m+1}} & \text{otherwise}
\end{array}
\right.
\]}\label{sg.4}\\
\STATE If necessary, repair $\mathbf{y}^{''}\rightarrow\mathbf{y}$\label{sg.5}\\
\end{algorithmic}
\end{algorithm}

\subsection{Updating on Population and Clusters}

In Alg.~\ref{framework} line~\ref{fw.9}, function $\textsc{Esco}$ is applied to carry out environmental selection and clustering updating. OCEA adopts the environmental selection method proposed in SMS-EMOA~\cite{beume2007sms} which is based on the hypervolume metric. The hypervolume metric is the only known unitary metric that is Pareto compliant~\cite{zitzler2003performance}. It has shown better performance over decomposition-based and Pareto dominance-based environmental selection approaches~\cite{2011BaderZ}.

Regarding cluster updating, we modify the online agglomerative clustering algorithm AddC (Alg.~\ref{online_clustering}) so that it can be fitted into the evolutionary search mechanism. The modified AddC is fused in OCEA to update/refine the clusters to adaptively learn the PS's structure.

Alg.~\ref{es} presents the details of \textsc{Esoc}. For each new solution $\mathbf{y}$, $\mathcal{A}$ is updated by the hypervolume metric based environmental selection. Specifically, the fast non-dominanted sorting approach proposed in NSGA-II~\cite{2002DebPAM} is applied to partition the external archive ${\mathcal{A}} \cup \{\mathbf{y}\}$ into $L$ non-dominanted fronts $\{\mathcal{B}^1,\cdots,\mathcal{B}^L\}$, where $\mathcal{B}^1$ is the best front and $\mathcal{B}^L$ is the worst one (line~\ref{es.1}). $L>1$ which indicates that there are more than one front in ${\mathcal{A}} \cup \{\mathbf{y}\}$. If it is the case, the solution $\mathbf{x}^*$ in $\mathcal{B}^L$ with the largest $d(\mathbf{x},{\mathcal{A}} \cup \{\mathbf{y}\})$ value is removed, where $d(\mathbf{x},\mathcal{A} \cup \{\mathbf{y}\})$ denotes the number of solutions in $\mathcal{A} \cup \{\mathbf{y}\}$ that dominates $\mathbf{x}$. Otherwise, if $L=1$, the solution $\mathbf{x}^*$ that least contributes to the hypervolume, i.e. $\Delta_{\varphi}(\mathbf{x},\mathcal{B}^1)$ (line~\ref{es.3} to \ref{es.4}, and \ref{es.9}), is excluded. The calculation of $\Delta_{\varphi}$ can be found in~\cite{beume2007sms}.

\begin{algorithm}[ht]
\caption{The updating procedure (\textsc{Esoc}).}\label{es}
\begin{algorithmic}[1]
\REQUIRE a new solution $\mathbf{y}$, external archive $\mathcal{A}$, centroids $\mathbf{z}^k$ and counters $c^k$ of current existing clusters $\mathcal{C}^1, \cdots, \mathcal{C}^m$, $1\leq k \leq m$, and the maximum number of clusters allowed $K_{\max}$.
\ENSURE External archive ${\mathcal{A}}$ and its cluster information.
\STATE Apply the fast non-dominanted sorting approach on ${\mathcal{A}} \cup \{\mathbf{y}\}$ to obtain $L$ fronts $\{\mathcal{B}^1,\cdots,\mathcal{B}^L\}$.\label{es.1}\\
\IF{$L>1$\label{es.2}}
	\STATE  Determine the worst solution, \[\mathbf{x}^* = \arg\max\limits_{\mathbf{x}\in \mathcal{B}^L}d(\mathbf{x},{\mathcal{A}} \cup \{\mathbf{y}\}).\]\label{es.3}
	\ELSE
	\STATE Determine the worst solution, \[\mathbf{x}^* = \arg\min\limits_{\mathbf{x}\in {\mathcal{A}}\cup \{\mathbf{y} \}}\Delta_{\varphi}(\mathbf{x},\mathcal{B}^1).\]\label{es.4}
\ENDIF
\IF{$\mathbf{x}^*\neq \mathbf{y}$}\label{es.17}
	\STATE If $\mathbf{x}^* \in \mathcal{C}^k, k\in\{1,\cdots,m\}$, then remove $\mathbf{x}^*$ from $\mathcal{C}^k$: $\mathcal{C}^k=\mathcal{C}^k\backslash\{\mathbf{x}^*\}$. \label{es.5}\\
		\IF{$\mathcal{C}^k=\emptyset$\label{es.6}}
		\STATE Remove cluster $\mathcal{C}^k$, set $m=m-1$.\label{es.7}\\
	\ELSE
		\STATE Update cluster $\mathcal{C}^k$: \[c^k=c^k-1, \mathbf{z}^k = \mathbf{z}^k - \frac{\mathbf{x}^*-\mathbf{z}^k}{c^k}.\]\label{es.8}
	\ENDIF\label{es.15}
	\STATE Delete the worst solution ${\mathcal{A}}= {\mathcal{A}}\cup \{\mathbf{y}\}\backslash\{\mathbf{x}^*\}$.\label{es.9}
	\STATE Set $m=m+1$, construct a new cluster $\mathcal{C}^m$, set $c^m=1$, $\mathbf{z}^m=\mathbf{y}$.\label{es.10}\\
	\IF{$m>K_{\max}$}\label{es.11}
		\STATE Find two closest clusters, \[(\gamma, \delta) = \arg\min\limits_{\gamma, \delta, \gamma \neq \delta} ||\mathbf{z}^\gamma -\mathbf{z}^\delta\|.\]\label{es.12}
		\STATE Merge the two clusters, \[\mathbf{z}^\gamma = \frac{\mathbf{z}^\gamma c^\gamma + \mathbf{z}^\delta c^\delta}{c^\gamma + c^\delta},~c^\gamma = c^\gamma + c^\delta.\]\label{es.13}
	\ENDIF\label{es.16}
\ELSE \label{es.18}
\STATE Delete the worst solution in ${\mathcal{A}}= {\mathcal{A}}\cup \{\mathbf{y}\}\backslash\{\mathbf{x}^*\}$.\label{es.14}\\
\ENDIF
\end{algorithmic}
\end{algorithm}

If $\mathbf{y}$ is kept in $\mathcal{A}$ after environmental selection, i.e., $\mathbf{x}^*\neq \mathbf{y}$, the online clustering operation is invoked. First, $\mathbf{x}^*$ is removed from its cluster ${\cal C}^*$, and its cluster's centroid and counter are updated following equations in line~\ref{es.8}. It differs from AddC where no data points are to be removed during the online clustering process. Then $\mathbf{y}$ is taken as a new centroid to construct a new cluster (line~\ref{es.10}). If there are more than $K_{\max}$ clusters in $\mathcal{A}$, two clusters that are closest to each other are emerged (lines~\ref{es.11} to \ref{es.13}) to complete the clustering operation.

\subsection{Notes on OCEA}

It is necessary to emphasize that:

\begin{itemize}
\item The evolution procedure of OCEA is also an online clustering procedure working on a stream of offsprings which are created and updated during the evolution process. We would expect that the clustering structure is to be gradually emerged during evolution and finally gets well shaped at termination.

\item Different from the original AddC (Alg.~\ref{online_clustering}), (a) the clustering procedure in OCEA starts from the $N$ initial clusters composed of the $N$ solutions in the initial population (line~\ref{fw.2} in Alg.~\ref{framework}); (b) During the evolution, some solutions are dominated and need to be removed. An extra operation is added to account for the removal of solutions, including the updating of cluster statistics and the discarding of any empty cluster (lines~\ref{es.5} to \ref{es.15} in Alg.~\ref{es}); (c) In our online clustering procedure, $\mathbf{y}$ is not assigned to its closest cluster as opposed to Alg.~\ref{online_clustering} where a new data need to be assigned to its closest cluster (line~\ref{es.10} in Alg.~\ref{es}).

\item OCEA incorporates the online clustering tightly within the evolution search. The online clustering discovers adaptively the PS structure along with the evolution. New solutions are created taking the cluster information into account at each generation. As a result, it can be seen that the online clustering closely adapts to the search procedure; and accounts for the non-stationary of the evolution dynamics.

\item Different from existing regularity model-based MOEAs in which the learning at each generation has a time complexity linearly to the number of training iterations. The number of generations should be large enough to make sure the convergence of the learning algorithm. On the contrary, in our scheme, each solution is visited only once. This can significantly reduce the computational burden.

\item In our scheme, we do not require a post-process which is different from the original AddC algorithm.
\end{itemize}

\section{Experimental Study}\label{expStu}

To investigate the performance of OCEA, it is compared with two decomposition-based MOEAs (MOEA/D-DE~\cite{2009LiZ} and TMOEA/D~\cite{liu2010t}), one regularity model based MOEA (RM-MEDA~\cite{2008ZhangZJ}), one popular performance metric based MOEA (SMS-EMOA~\cite{beume2007sms}), and one typical Pareto dominance based MOEA (NSGA-II~\cite{2002DebPAM}).

Among these algorithms, MOEA/D-DE decomposes the MOP into a set of single-objective problems with uniformly distributed weights. It might be not able to obtain approximated fronts with good diversity for MOPs with complex PFs. TMOEA/D transforms the objective functions into those that are easy to be addressed by MOEA/D. This is to make MOEA/D perform well on MOPs with complex PFs. RM-MEDA is developed based on the regularity property. It learns some local principle components at each generation, and uses the principle components to approximate the manifold structure. SMS-MOEA uses the hypervolume metric as the selection criterion. NSGA-II, on the other hand, uses the Pareto dominance relationship among individuals and crowding distance to carry out environmental selection. These comparison algorithms cover all the main streams of MOEAs in the literatures.

\subsection{Test Instances and Performance Metrics}

MOPs with complex PF and complicated PS structures are particularly focused in this paper. The GLT test suite from~\cite{zhang2015self} are used in the comparison experiments. The test suite includes a variety of problems with various characteristics that challenge MOEAs greatly. Those characteristics include disconnected PF, convex PF, nonlinear variable linkage, etc.

Two commonly-used performance metrics, \emph{inverted generational distance (IGD)}~\cite{2008ZhangZJ} and \emph{hypervolume (HV)}~\cite{zitzler1999multiobjective}, are employed to measure the algorithm's performance. These two metrics can measure both the convergence and diversity of the final approximated fronts found by MOEAs. Lower IGD and larger HV metric values imply better performance of MOEAs.

To calculate the HV metric value of an approximated front, a reference point which can be dominated by all the objective vectors in the final approximated front need to be set. The reference points chosen for the test instances are as follows: for GLT1, $\mathbf{r}=(2,2)^\intercal$, for GLT2 $\mathbf{r}=(2,11)^\intercal$, for GLT3 $\mathbf{r}=(2,2)^\intercal$ and for GLT4 $\mathbf{r}=(2,3)^\intercal$, for GLT5-GLT6, $\mathbf{r}=(2,2,2)^\intercal$.

\subsection{Experimental Settings}\label{expSet}

It has been well acknowledged that for the GLT test instances, the DE and PM operators are more able to produce promising solutions than other operators~\cite{zhang2015self}. Therefore, to make a fair comparison, the recombination operators in NSGA-II and SMS-EMOA are replaced by the DE and PM operators used in this paper. Furthermore, all parameters in the experiments are adjusted through preliminary experiments for optimal performance on these test instances. All algorithms are implemented in Matlab and tested in the same computer. The parameter settings for these algorithms are as follows:
\begin{itemize}
\item Common parameters:
\begin{itemize}
\item population size: $N=100$ for bi-objective and 105 for tri-objective instances;
\item search space dimension: $n=10$ for GLT1-GLT6;
\item runs: each algorithms independently runs each test instance for 33 times;
\item termination: maximum evolutionary generation $T=300$.
\end{itemize}
\item Parameters for OCEA:
\begin{itemize}
\item maximum number of clusters allowed: $K_{\max}=7$;
\item mating control parameter: $\beta=0.6$;
\item DE control parameters: $F=0.6,~CR=1$;
\item PM control parameters: $p_m=1/n,~\eta_m=20$.
\end{itemize}
\item Parameters for MOEA/D-DE:
\begin{itemize}
\item neighbourhood size: $NS=5$;
\item mating control parameter: $\beta=0.7$;
\item maximum number of solutions to be replaced by an offspring: 2;
\item DE control parameters: $F=0.9,~CR=0.6$;
\item PM control parameters: $p_m=1/n,~\eta_m=20$.
\end{itemize}
\item Parameters for TMOEA/D:
\begin{itemize}
\item neighbourhood size: $NS=30$;
\item generations for the first stage: $T1=T/10$;
\item generations for the second stage: $T2=\alpha T$,\\$\alpha= \{0.01,~0.02,~\cdots,~0.1,~0.1,~0.1,~0.15\}$;
\item DE control parameters: $F=0.5,~CR=1$.
\end{itemize}
\item Parameters for RM-MEDA:
\begin{itemize}
\item number of clusters in local PCA: 5;
\item maximum iterations used in local PCA: 50;
\item sampling extension ratio: 0.25.
\end{itemize}
\item Parameters for NSGA-II and SMS-EMOA:
\begin{itemize}
\item DE control parameters: $F=0.5,~CR=1$;
\item PM control parameters: $p_m=1/n,~\eta_m=20$.
\end{itemize}
\end{itemize}

To get statistically sound conclusions in the experiments, each algorithm independently runs 33 times for each instance, and the comparisons are performed based on the statistics of the performance metric values, i.e., mean and standard deviation values. In the comparison table, the mean IGD and HV metric values for each instance are sorted in an ascending and descending order, respectively, and the ranks are given in the square brackets of the table. The best mean metric values are highlighted in bold face with gray background. The Wilcoxon's rank sum test at a 5\% significance level is also performed to test the significance of differences between the mean metric values of each instance obtained by each pair of algorithms. In the tables, ``$\dag$", ``$\S$", and ``$\approx$" are used to denote that the mean metric values obtained by OCEA is better than, worse than, or similar to those achieved by the comparison algorithm, respectively.

\subsection{Comparison Study}

To study the statistical performance of OCEA, Table~\ref{GLT_Test} shows the statistics of IGD and HV metric values obtained by MOEA/D-DE, TMOEA/D, RM-MEDA, NSGA-II, SMS-EMOA and OCEA on the GLT test suite averaged over 33 independent runs. In general, OCEA obtains 8 out of 12 best mean metric values, while the rest algorithms only obtain 4. According to the mean ranks, the algorithms' performance ranked from the best to the worst are OCEA, RM-MEDA, TMOEA/D, SMS-EMOA, NSGA-II and MOEA/D-DE. Specifically, according to the Wilcoxon's rank sum test, in the 12 comparisons with each of MOEA/D-DE, TMOEA/D, RM-MEDA, NSGA-II and SMS-EMOA, OCEA achieves 12, 11, 11, 12, 11 better, 0, 1, 1, 0, 0 worse, and 0, 0, 0, 0, 1 similar mean metric values, respectively. Table~\ref{GLT_Test} denotes that OCEA performs the best overall on the GLT test suite.

\begin{table*}[!htb]\scriptsize
\centering \caption{Statistics (mean(std. dev.)[rank]) of IGD and HV metric values of final approximated fronts obtained by MOEA/D-DE, TMOEA/D, RM-MEDA, NSGA-II, SMS-EMOA and OCEA algorithms over 33 independent runs on the GLT test suite}\label{GLT_Test}
\begin{tabular}{lcccccc}\toprule
Instance&MOEA/D-DE&TMOEA/D&RM-MEDA&NSGA-II&SMS-EMOA&OCEA\\
\cmidrule{2-7}
&\multicolumn{6}{c}{IGD}\\
\midrule
GLT1&7.042e-03$^\dag_{9.92e-04}$[3]&5.472e-03$^\dag_{1.20e-03}$[2]&1.324e-02$^\dag_{1.87e-02}$[4]&1.550e-02$^\dag_{9.58e-03}$[5]&2.100e-02$^\dag_{1.01e-02}$[6]&\cellcolor{gray25}\textbf{2.041e-03}$_{4.45e-04}$[1]\\
GLT2&3.569e-01$^\dag_{7.33e-02}$[6]&3.636e-02$^\S_{9.69e-03}$[2]&\cellcolor{gray25}\textbf{3.327e-02}$^\S_{1.15e-03}$[1]&4.146e-02$^\dag_{2.66e-03}$[4]&4.955e-02$^\dag_{2.88e-02}$[5]&3.720e-02$_{1.54e-03}$[3]\\
GLT3&3.829e-02$^\dag_{1.12e-02}$[6]&2.212e-02$^\dag_{5.61e-02}$[5]&2.018e-02$^\dag_{1.11e-02}$[4]&1.405e-02$^\dag_{8.67e-03}$[2]&1.929e-02$^\dag_{1.02e-02}$[3]&\cellcolor{gray25}\textbf{6.432e-03}$_{3.13e-03}$[1]\\
GLT4&1.985e-02$^\dag_{3.42e-03}$[2]&4.462e-02$^\dag_{1.14e-01}$[5]&4.147e-02$^\dag_{5.41e-02}$[4]&4.106e-02$^\dag_{4.65e-02}$[3]&5.505e-02$^\dag_{5.33e-02}$[6]&\cellcolor{gray25}\textbf{5.769e-03}$_{9.19e-05}$[1]\\
GLT5&8.079e-02$^\dag_{3.14e-03}$[6]&4.409e-02$^\dag_{9.69e-04}$[3]&5.130e-02$^\dag_{1.95e-03}$[4]&6.419e-02$^\dag_{4.26e-03}$[5]&3.018e-02$^\dag_{3.93e-04}$[2]&\cellcolor{gray25}\textbf{2.942e-02}$_{5.11e-04}$[1]\\
GLT6&5.582e-02$^\dag_{2.18e-02}$[6]&4.059e-02$^\dag_{3.31e-02}$[4]&3.835e-02$^\dag_{2.15e-03}$[3]&5.384e-02$^\dag_{3.97e-03}$[5]&2.225e-02$^\approx_{4.01e-04}$[2]&\cellcolor{gray25}\textbf{2.223e-02}$_{6.98e-04}$[1]\\
\toprule
&\multicolumn{6}{c}{HV}\\
\cmidrule{2-7}
GLT1&3.367e+00$^\dag_{4.98e-03}$[2]&3.366e+00$^\dag_{2.58e-03}$[3]&3.316e+00$^\dag_{3.72e-02}$[4]&3.312e+00$^\dag_{2.31e-02}$[5]&3.297e+00$^\dag_{2.39e-02}$[6]&\cellcolor{gray25}\textbf{3.369e+00}$_{4.31e-03}$[1]\\
GLT2&1.943e+01$^\dag_{6.62e-02}$[6]&1.977e+01$^\dag_{3.93e-02}$[2]&1.970e+01$^\dag_{6.87e-03}$[5]&1.972e+01$^\dag_{7.86e-03}$[3]&1.972e+01$^\dag_{1.13e-01}$[4]&\cellcolor{gray25}\textbf{1.981e+01}$_{1.14e-03}$[1]\\
GLT3&3.941e+00$^\dag_{2.10e-03}$[6]&3.943e+00$^\dag_{1.92e-02}$[5]&3.944e+00$^\dag_{1.88e-03}$[4]&3.946e+00$^\dag_{1.64e-03}$[2]&3.946e+00$^\dag_{2.01e-03}$[3]&\cellcolor{gray25}\textbf{3.948e+00}$_{9.82e-04}$[1]\\
GLT4&4.980e+00$^\dag_{2.51e-03}$[2]&4.869e+00$^\dag_{4.10e-01}$[6]&4.961e+00$^\dag_{3.85e-02}$[3]&4.954e+00$^\dag_{5.36e-02}$[4]&4.953e+00$^\dag_{3.50e-02}$[5]&\cellcolor{gray25}\textbf{4.993e+00}$_{5.82e-04}$[1]\\
GLT5&7.939e+00$^\dag_{1.78e-03}$[5]&7.958e+00$^\dag_{9.53e-04}$[3]&7.951e+00$^\dag_{1.45e-03}$[4]&7.939e+00$^\dag_{3.24e-03}$[6]&7.968e+00$^\dag_{1.96e-04}$[2]&\cellcolor{gray25}\textbf{7.969e+00}$_{2.67e-04}$[1]\\
GLT6&7.937e+00$^\dag_{1.35e-02}$[5]&7.947e+00$^\dag_{1.89e-02}$[4]&7.948e+00$^\dag_{1.52e-03}$[3]&7.933e+00$^\dag_{3.19e-03}$[6]&7.960e+00$^\dag_{2.82e-04}$[2]&\cellcolor{gray25}\textbf{7.961e+00}$_{4.16e-04}$[1]\\
\midrule
Mean Rank&4.583&3.667&3.583&4.167&3.833&1.167\\
$\dag$/$\S$/$\approx$&12/0/0&11/1/0&11/1/0&12/0/0&11/0/1&\\
\bottomrule
\end{tabular}
\end{table*}

\begin{figure*}[!htb]
\centering
\includegraphics[width=0.32\textwidth]{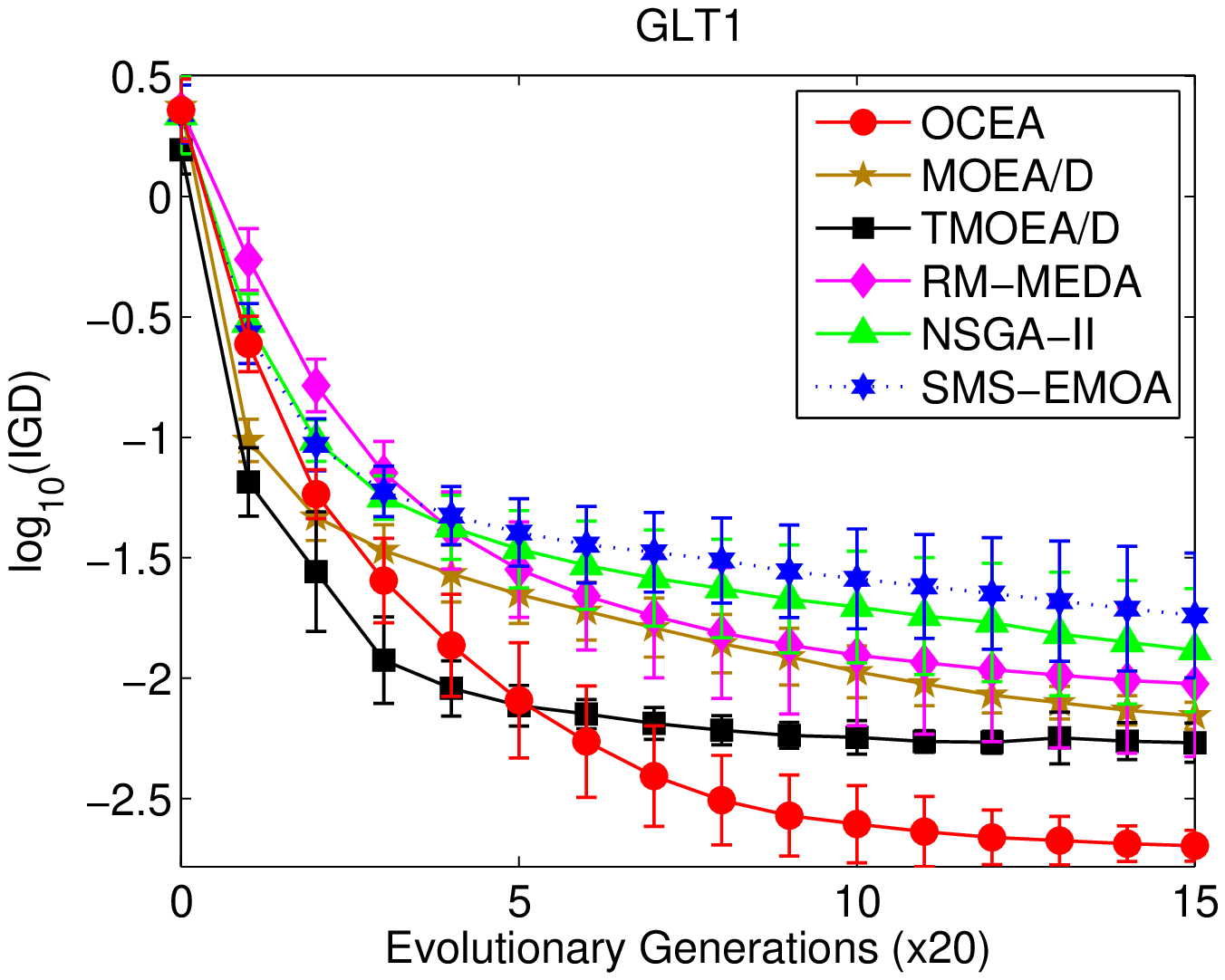}
\includegraphics[width=0.32\textwidth]{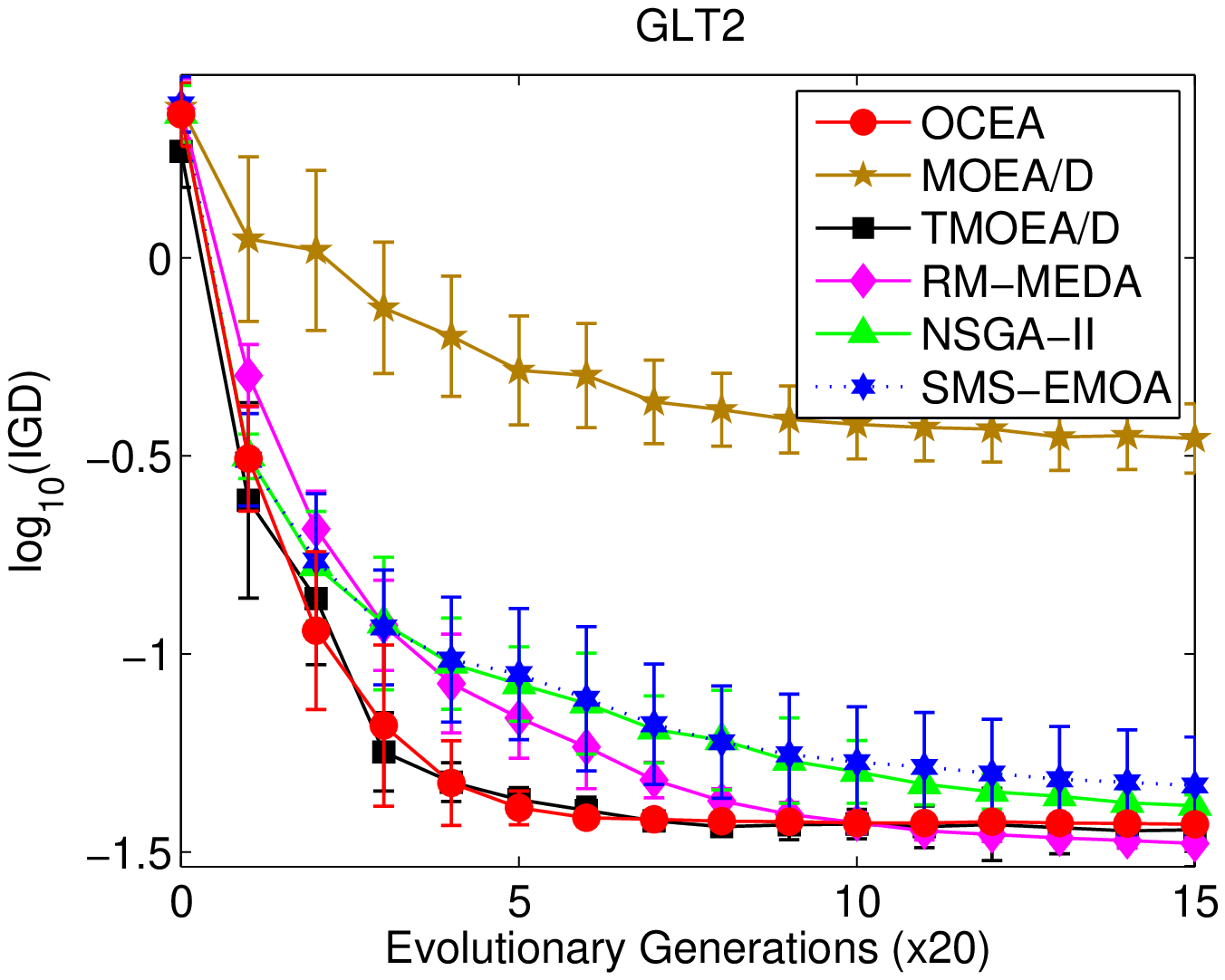}
\includegraphics[width=0.32\textwidth]{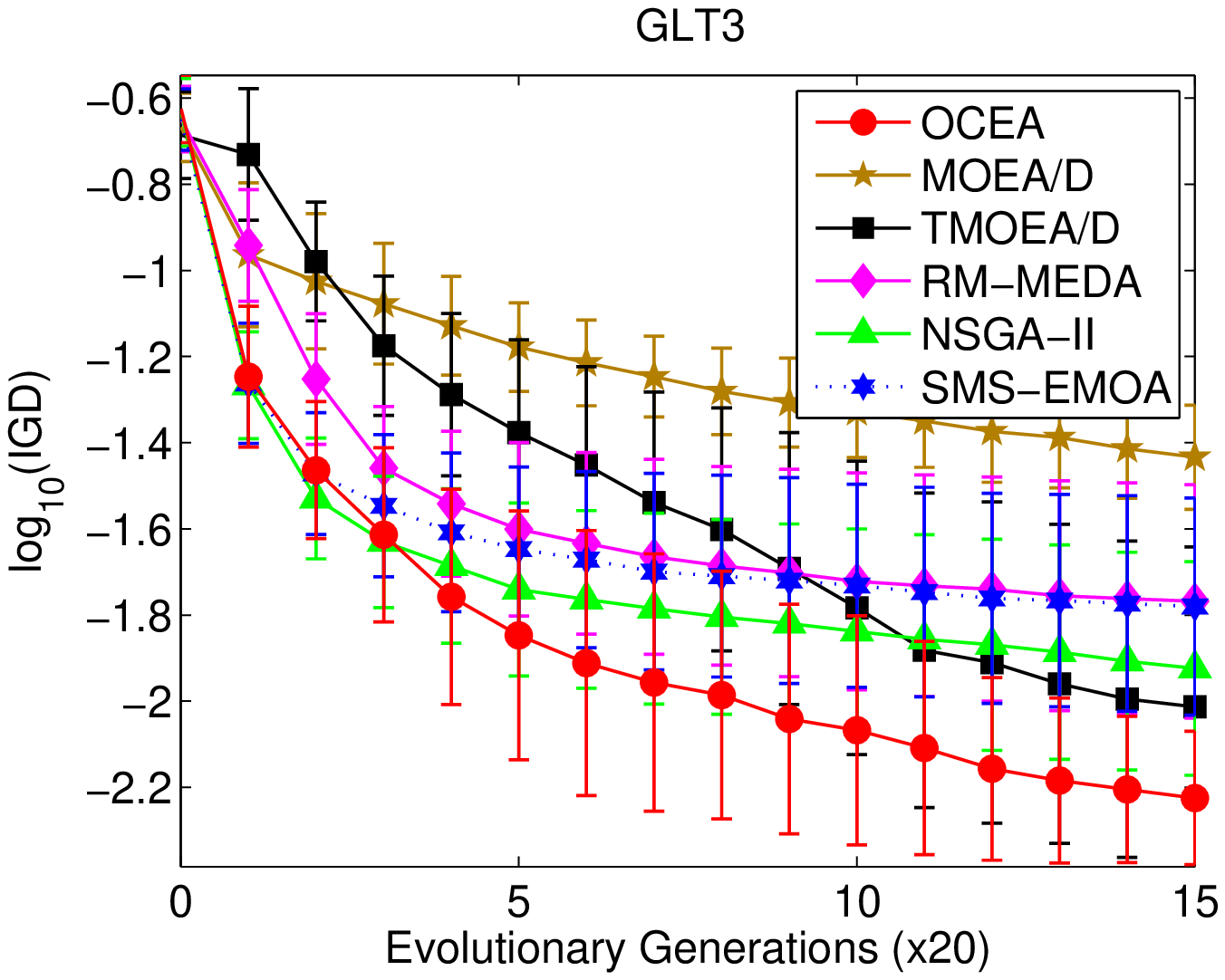}\\
\includegraphics[width=0.32\textwidth]{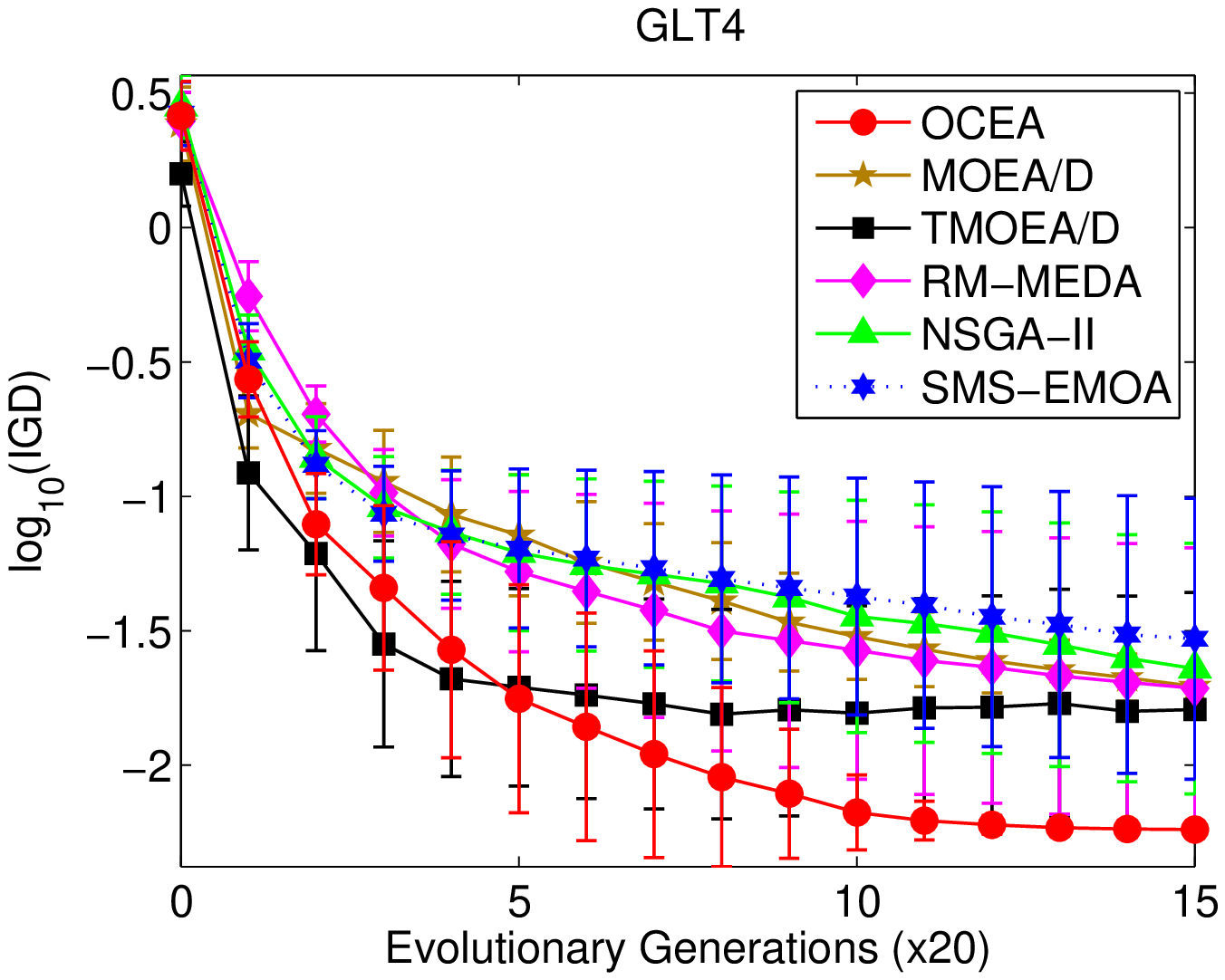}
\includegraphics[width=0.32\textwidth]{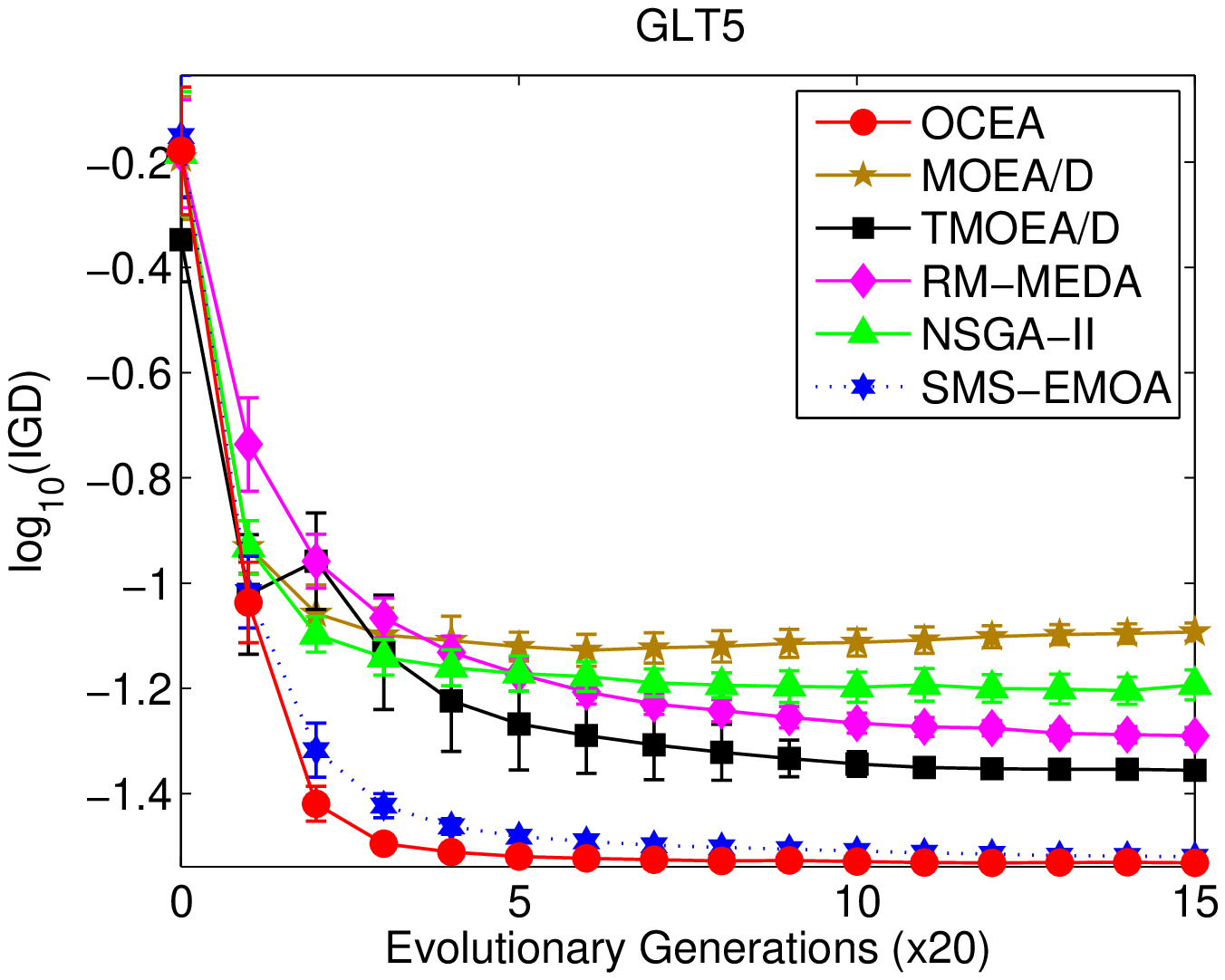}
\includegraphics[width=0.32\textwidth]{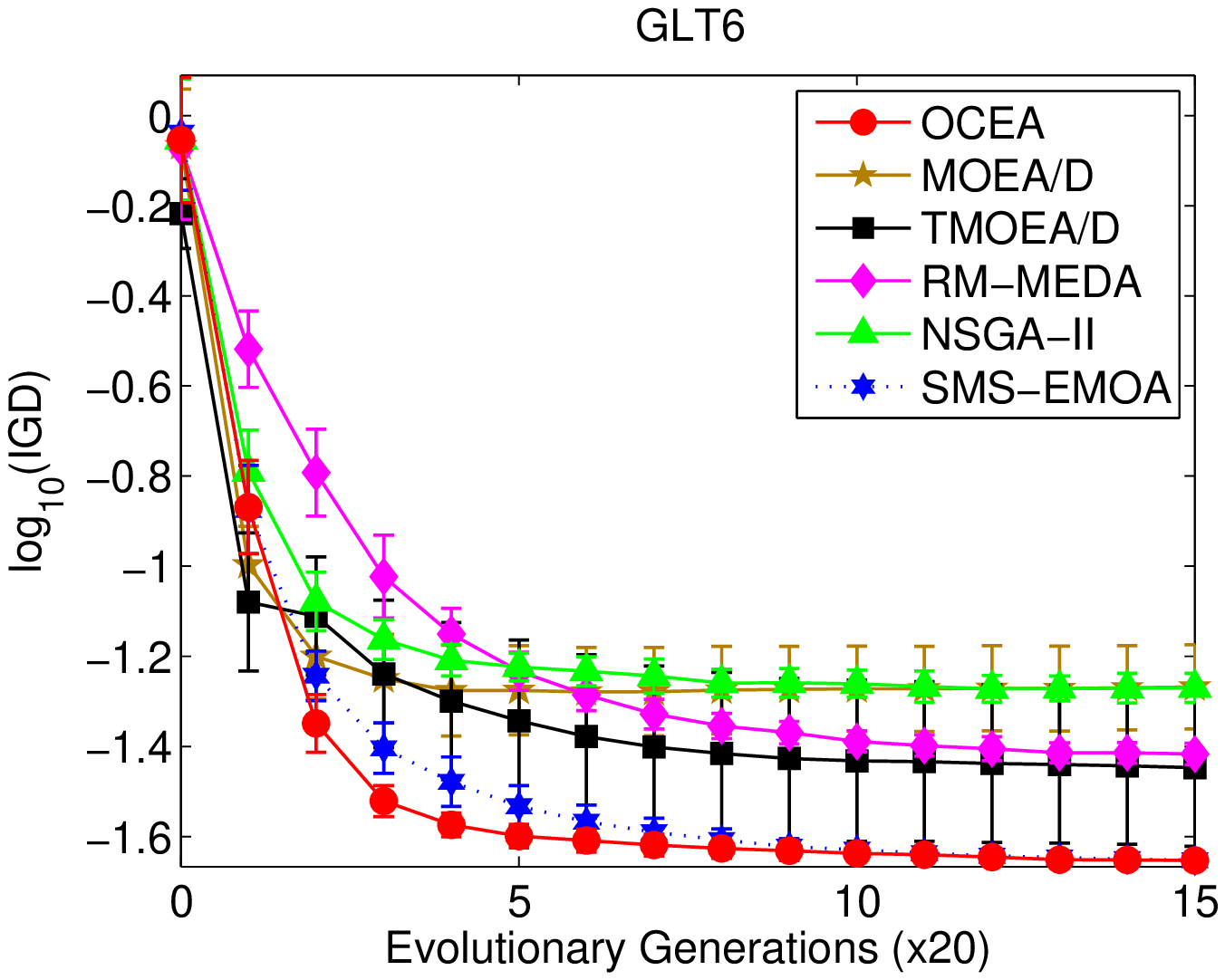}
\caption{Evolution of the statistics of IGD metric values obtained by MOEA/D-DE, TMOEA/D, RM-MEDA, NSGA-II, SMS-EMOA and OCEA on GLT1-GLT6}\label{IGDEvolution}
\end{figure*}

To observe the search efficiency of OCEA, Fig.~\ref{IGDEvolution} shows the evolution of the statistics of the IGD metric values obtained by the six algorithms on GLT1-GLT6. From the figure, it can be seen that for GLT1 and GLT3-GLT6, OCEA reaches the fastest to the lowest mean IGD metric values. For GLT2, OCEA has the slower, similar and faster speed in comparison with RM-MEDA, TMOEA/D and the other algorithms, respectively. Moreover, when dealing with GLT2, OCEA actually performs better than RM-MEDA at the early stage compared with RM-MEDA. From the evolution of the standard deviations of the metrics, it also can be observed that within 300 generations, OCEA has achieved robust performance on all the instances except for GLT3. Fig.~\ref{IGDEvolution} indicates that OCEA approaches the fastest to the PFs and maintains the most diverse populations among the comparison algorithms on average.

\begin{figure*}[htbp]
\centering
\includegraphics[width=0.24\textwidth]{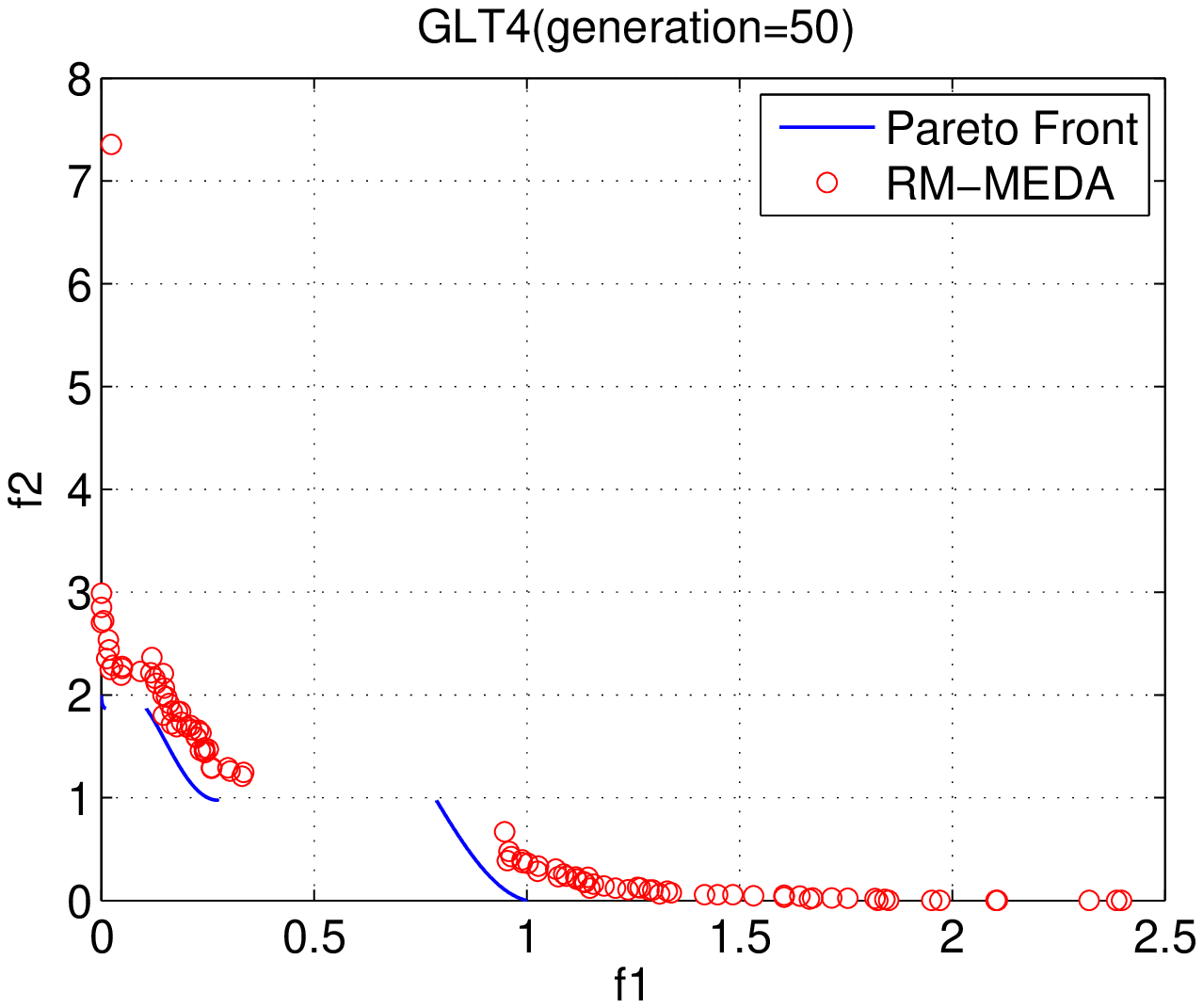}
\includegraphics[width=0.24\textwidth]{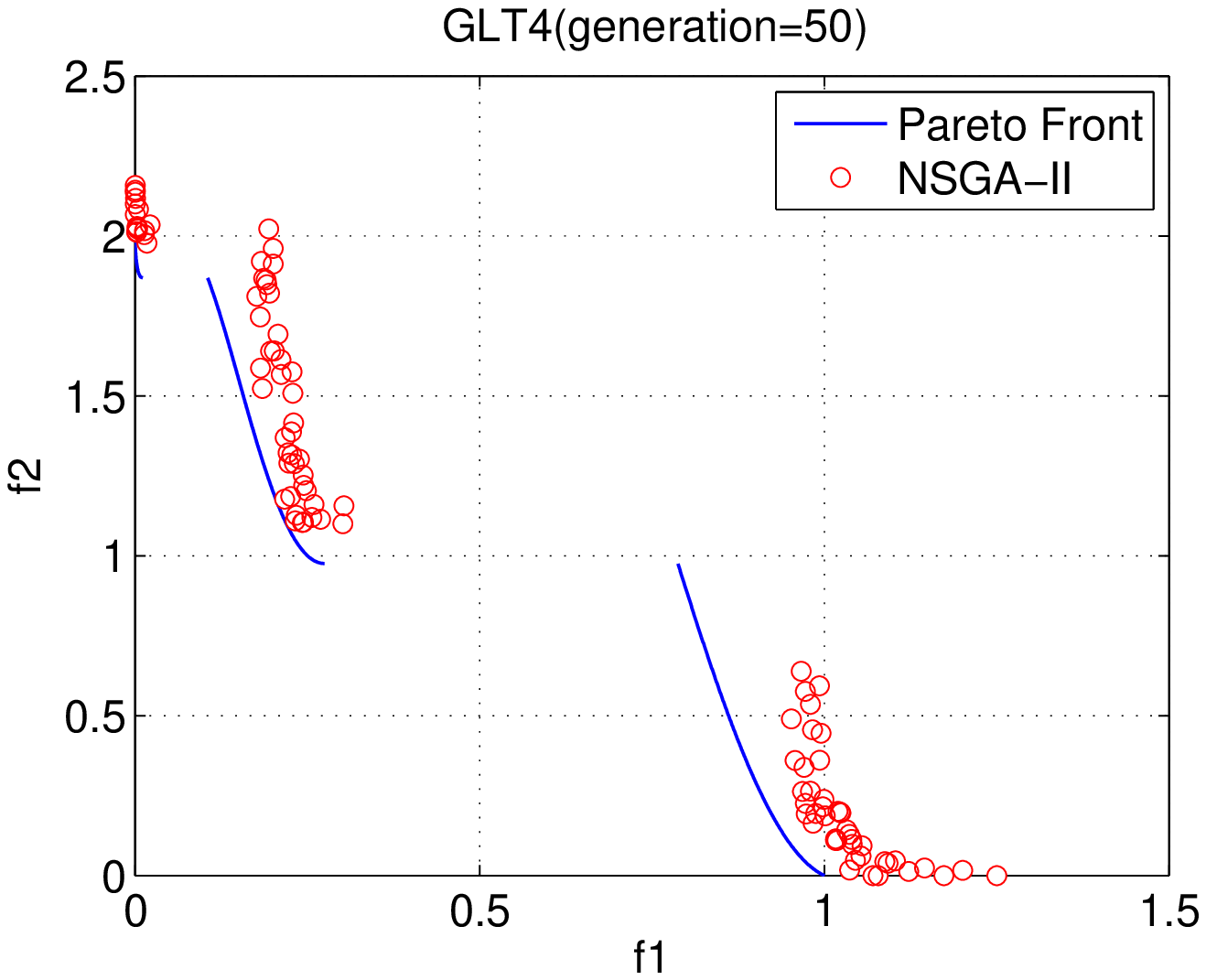}
\includegraphics[width=0.24\textwidth]{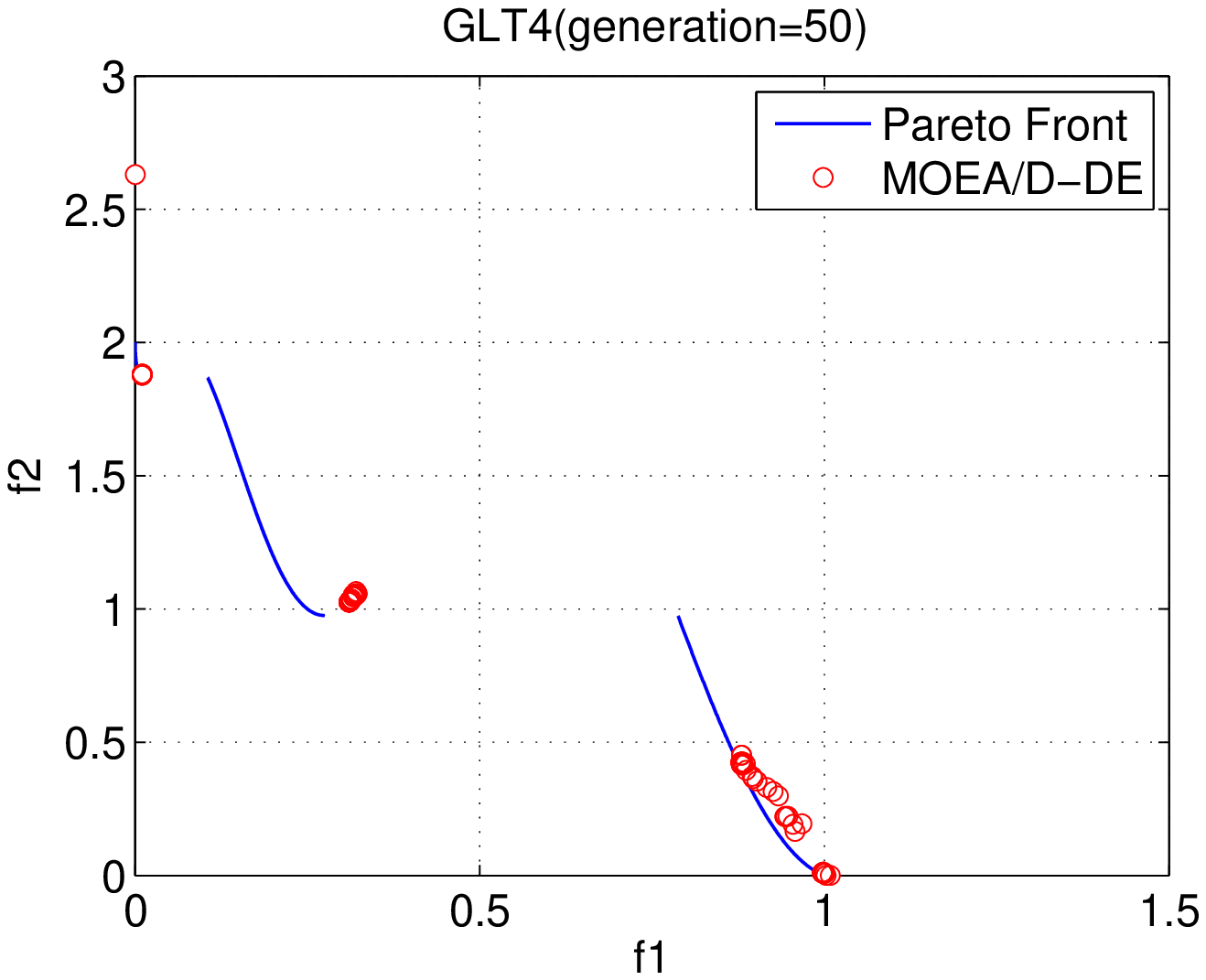}
\includegraphics[width=0.24\textwidth]{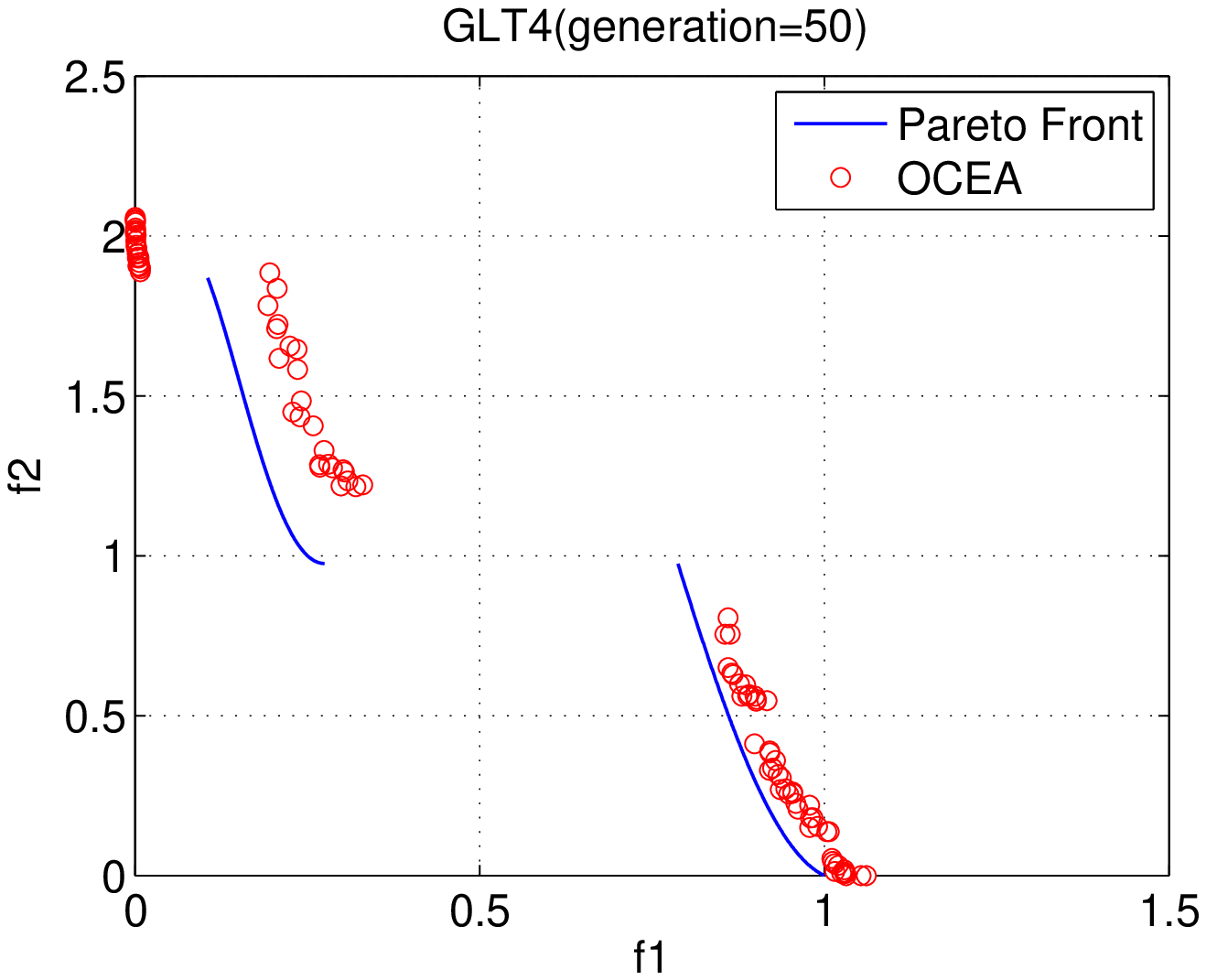}\\
\includegraphics[width=0.24\textwidth]{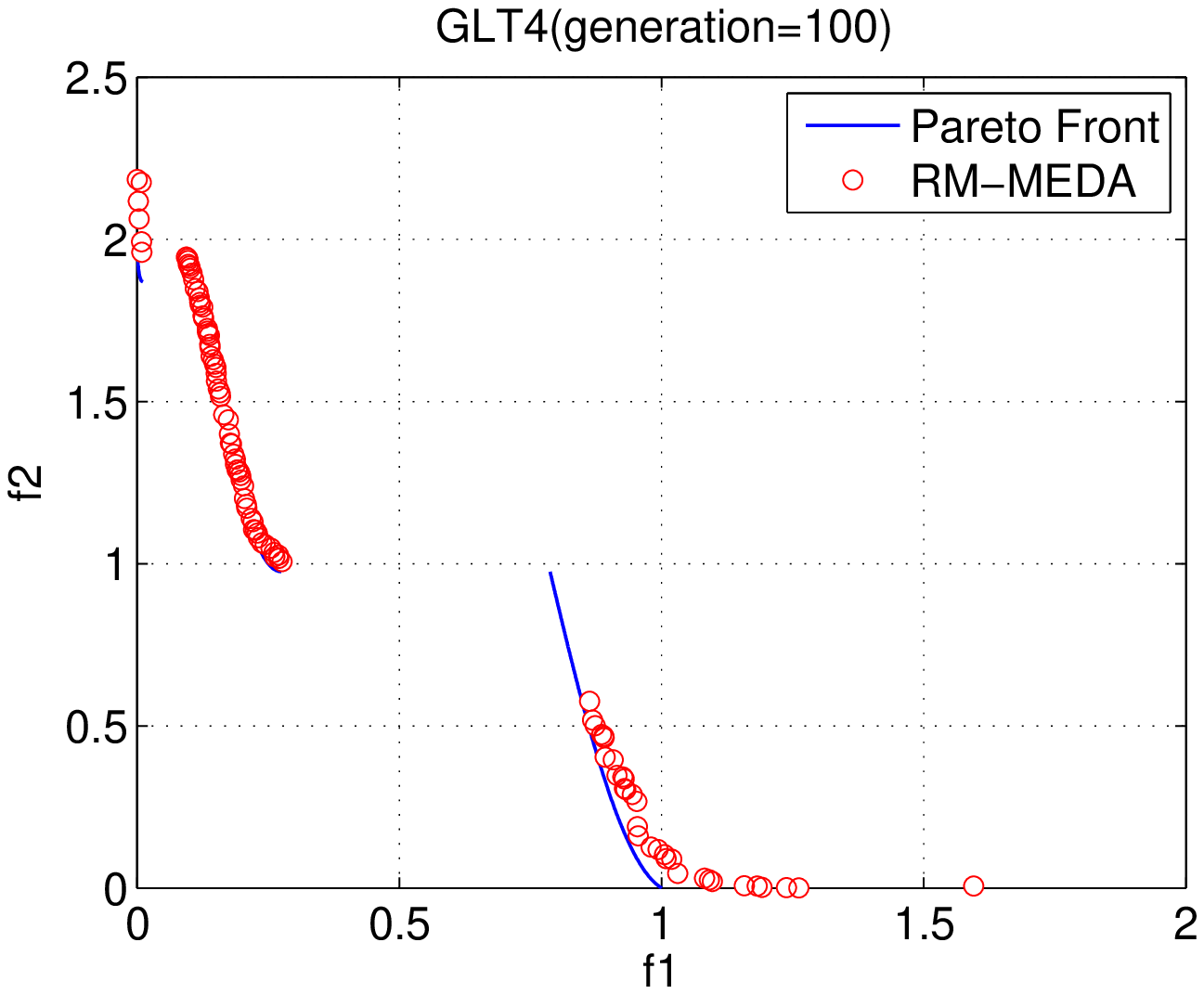}
\includegraphics[width=0.24\textwidth]{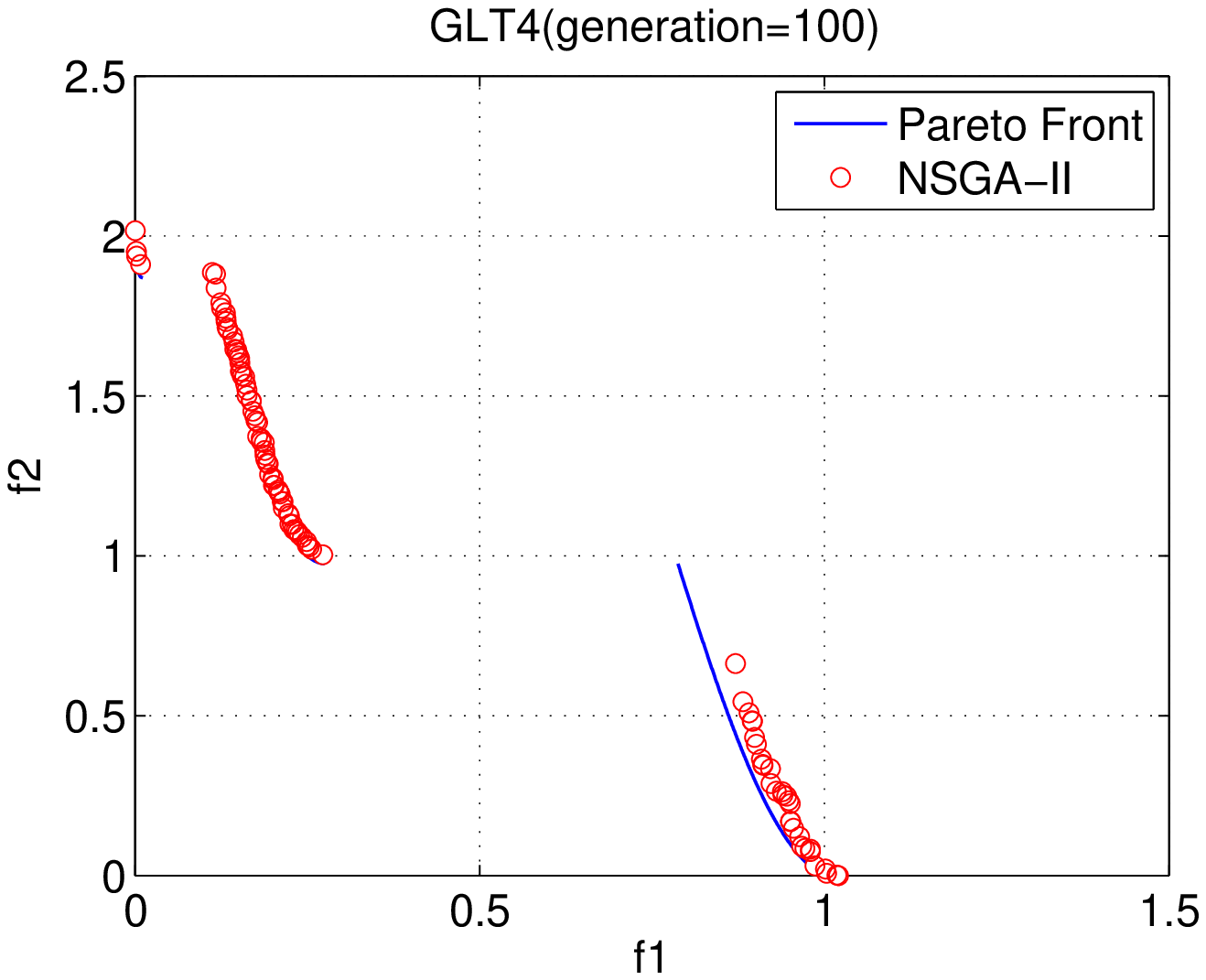}
\includegraphics[width=0.24\textwidth]{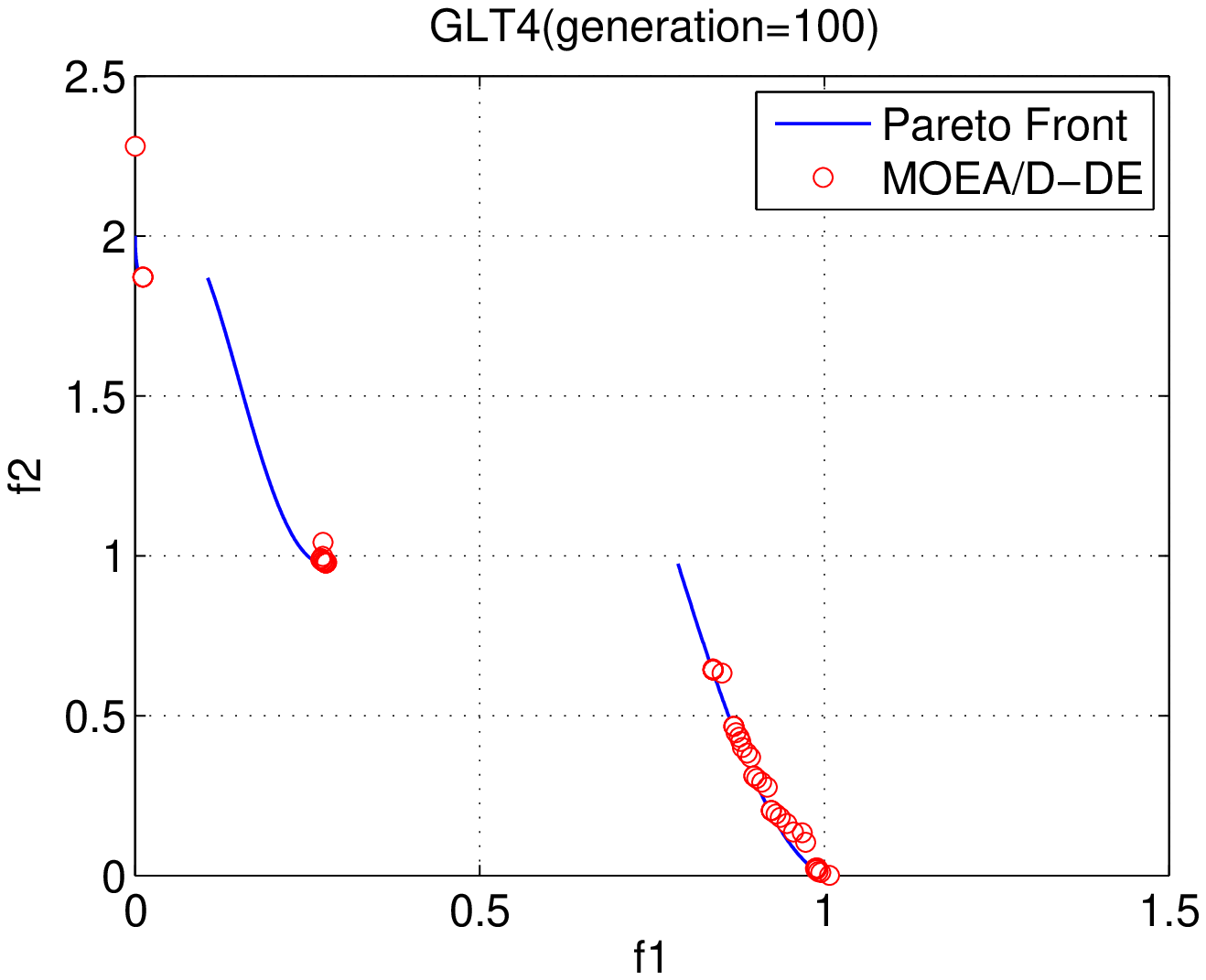}
\includegraphics[width=0.24\textwidth]{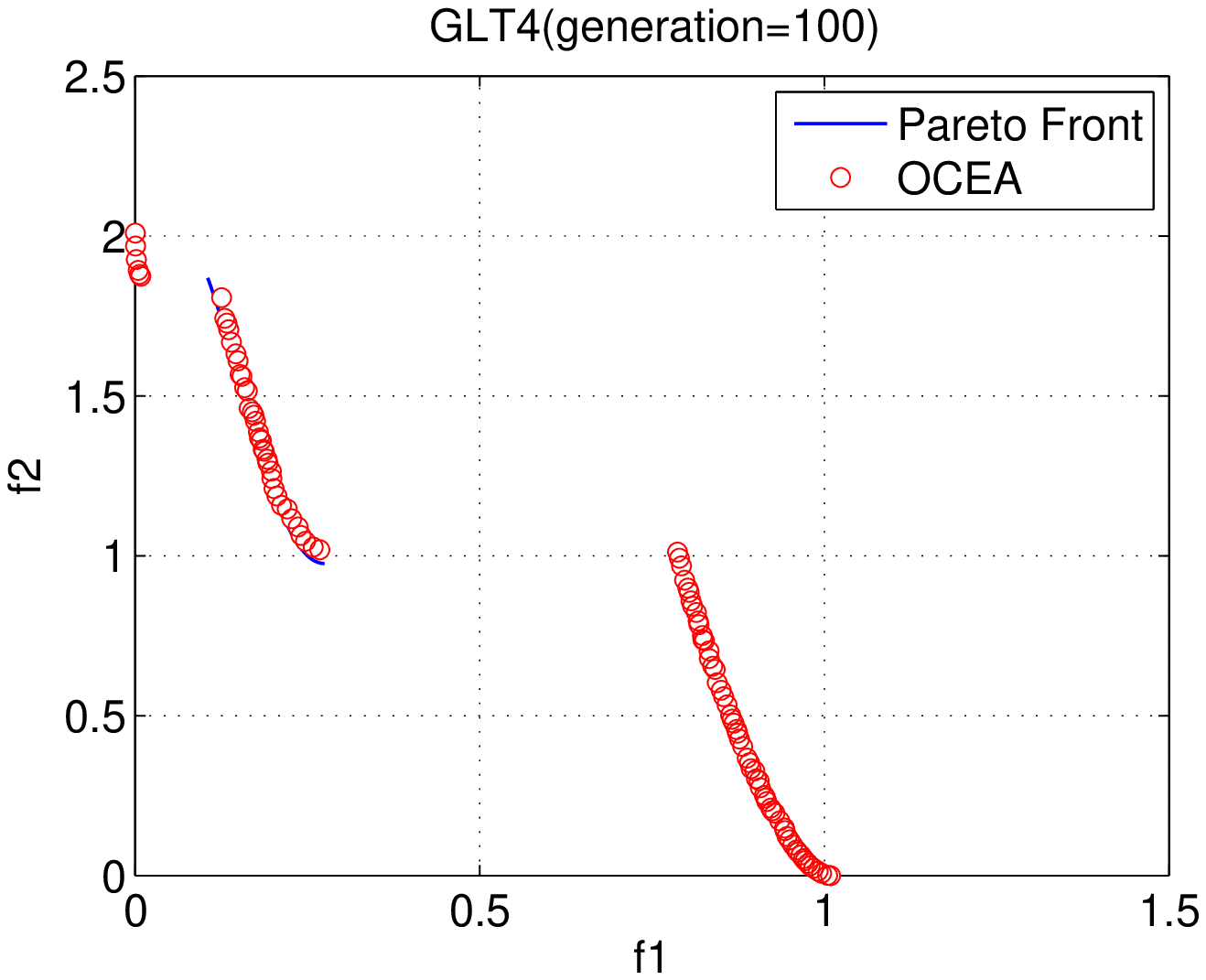}\\
\includegraphics[width=0.24\textwidth]{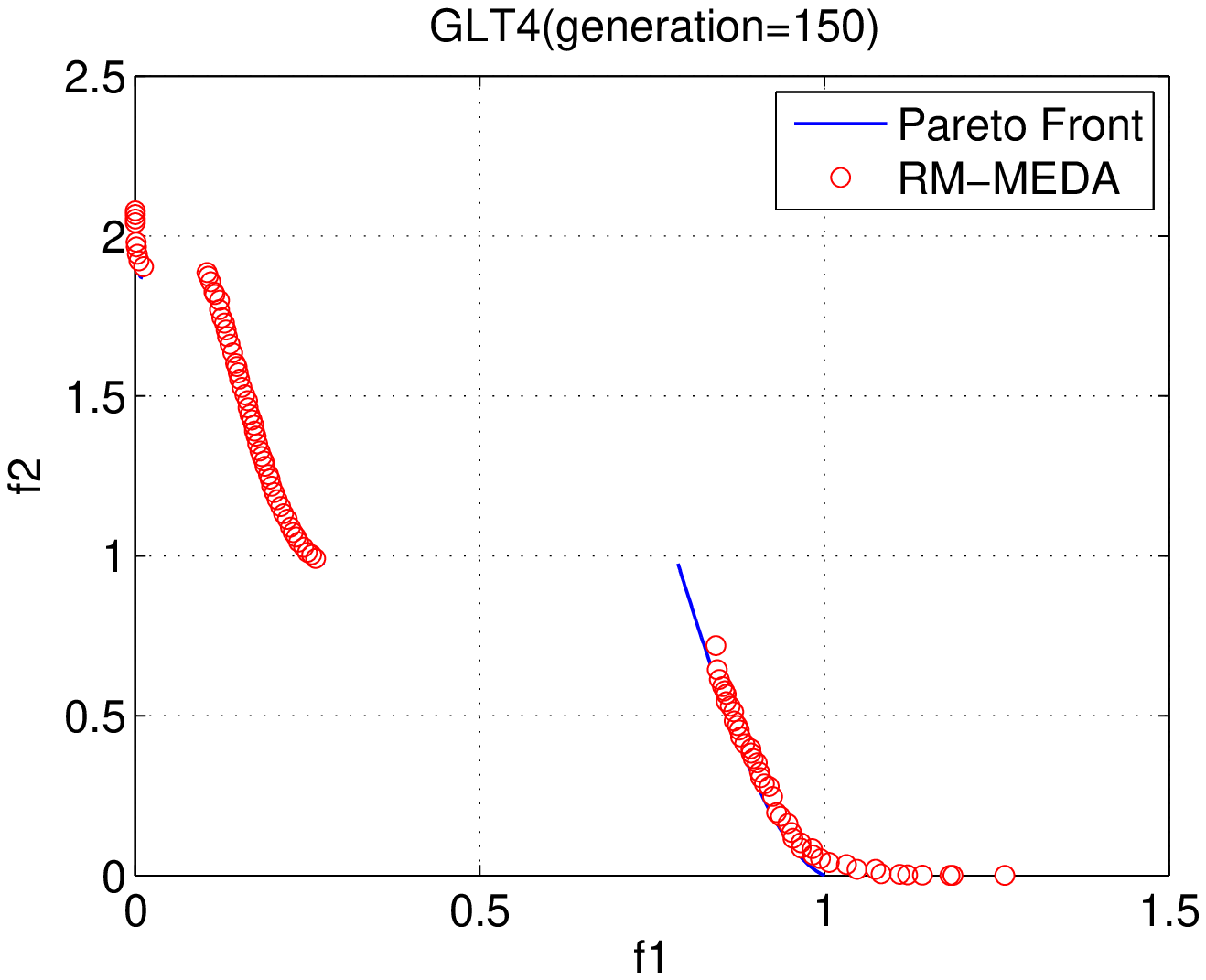}
\includegraphics[width=0.24\textwidth]{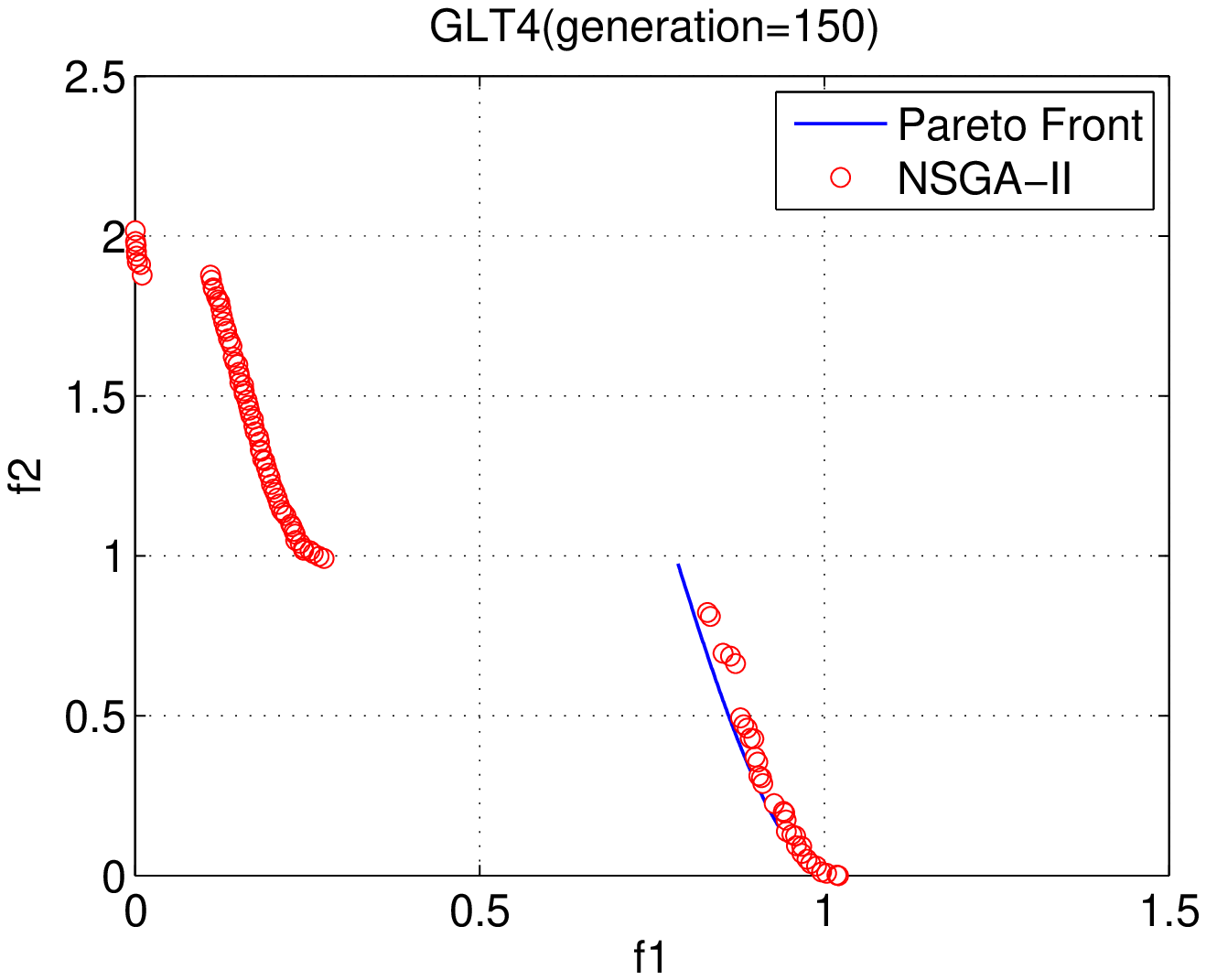}
\includegraphics[width=0.24\textwidth]{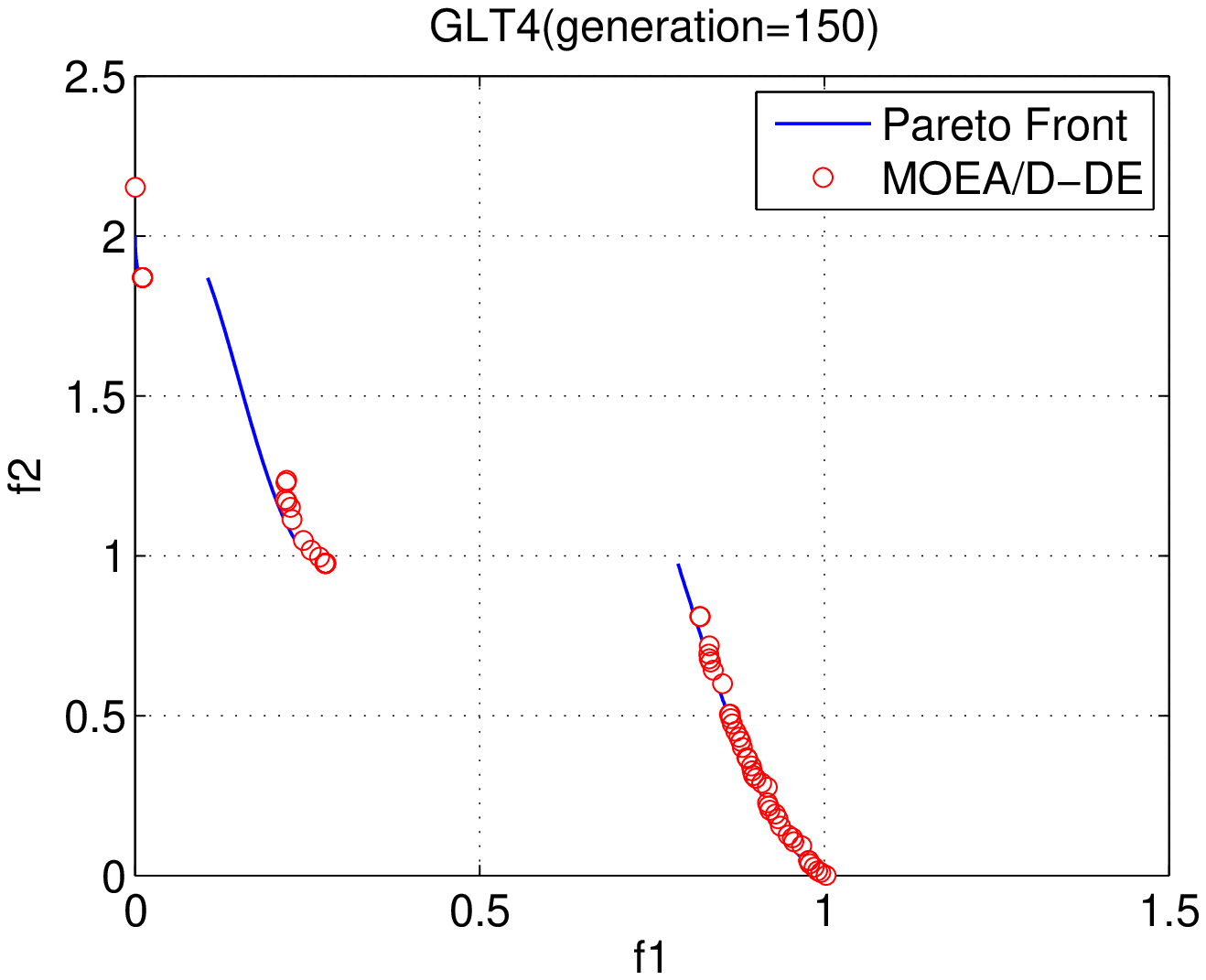}
\includegraphics[width=0.24\textwidth]{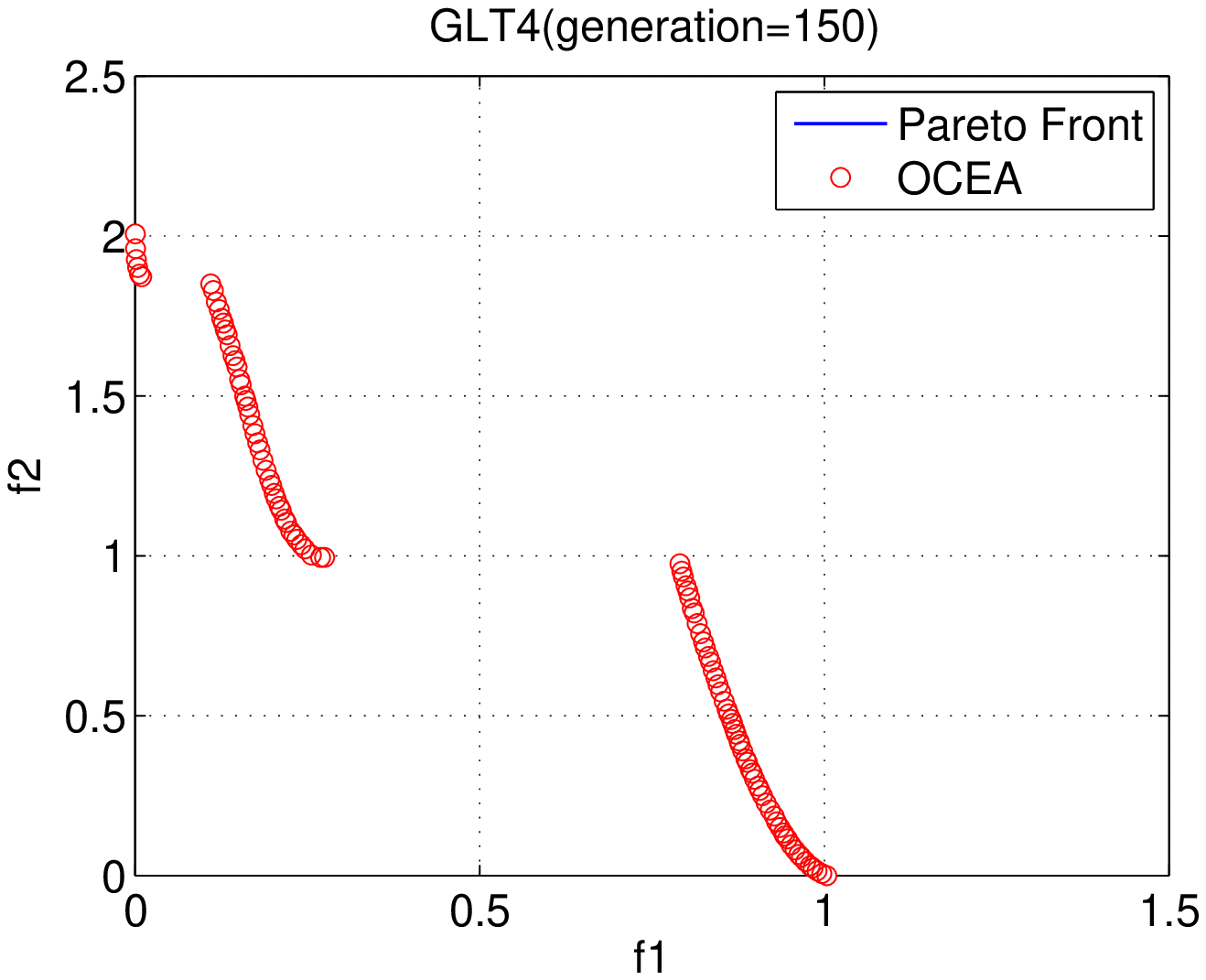}\\
\includegraphics[width=0.24\textwidth]{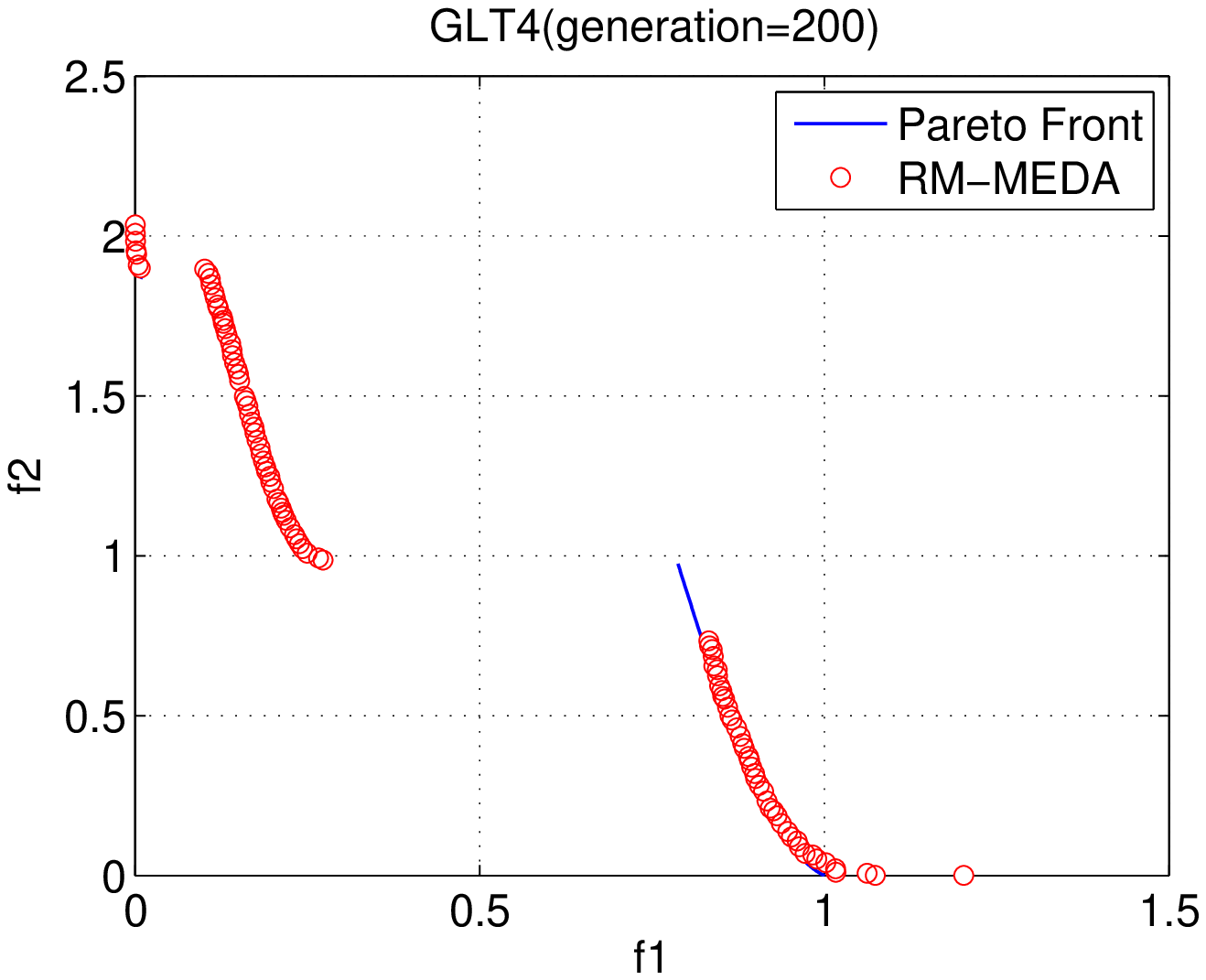}
\includegraphics[width=0.24\textwidth]{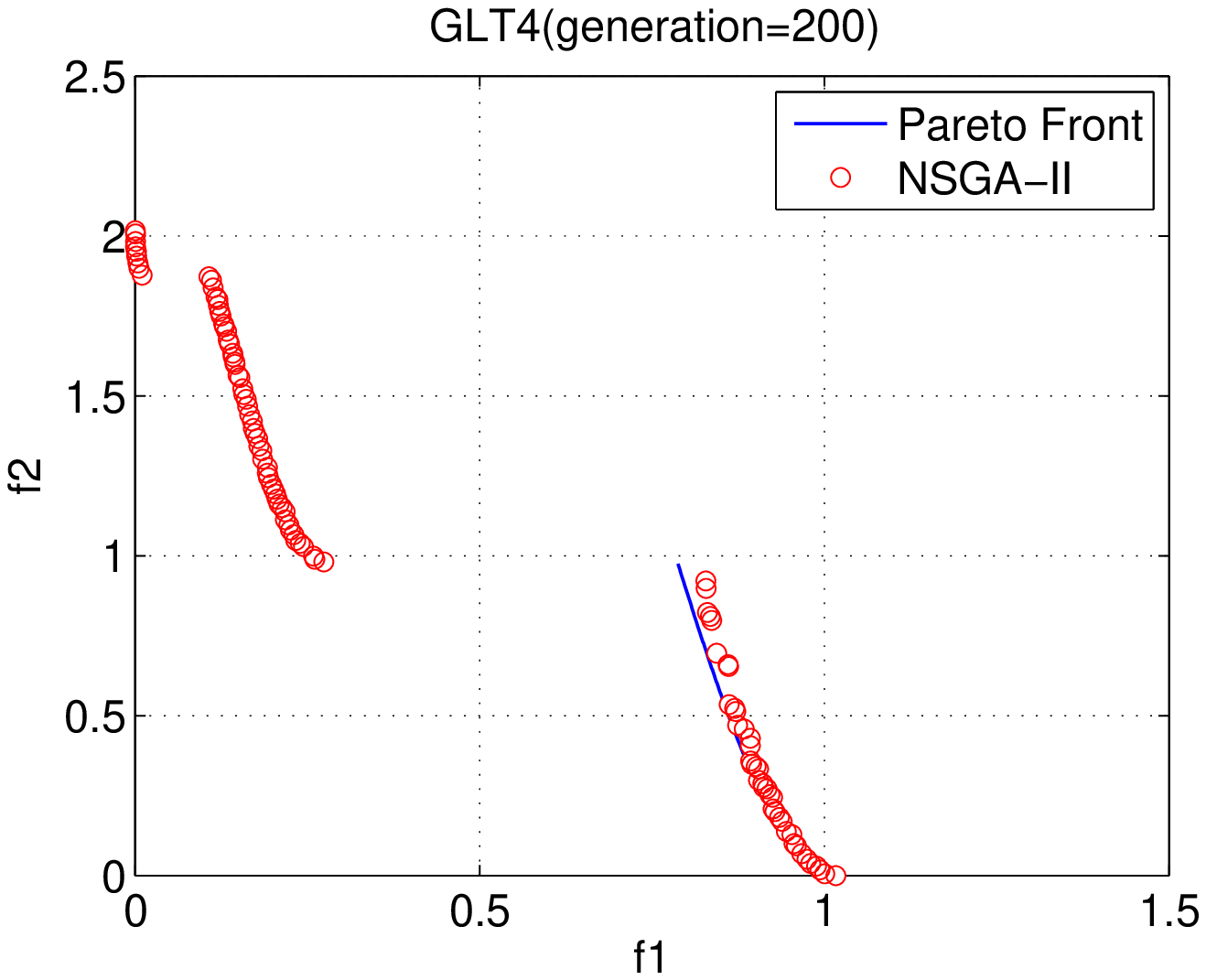}
\includegraphics[width=0.24\textwidth]{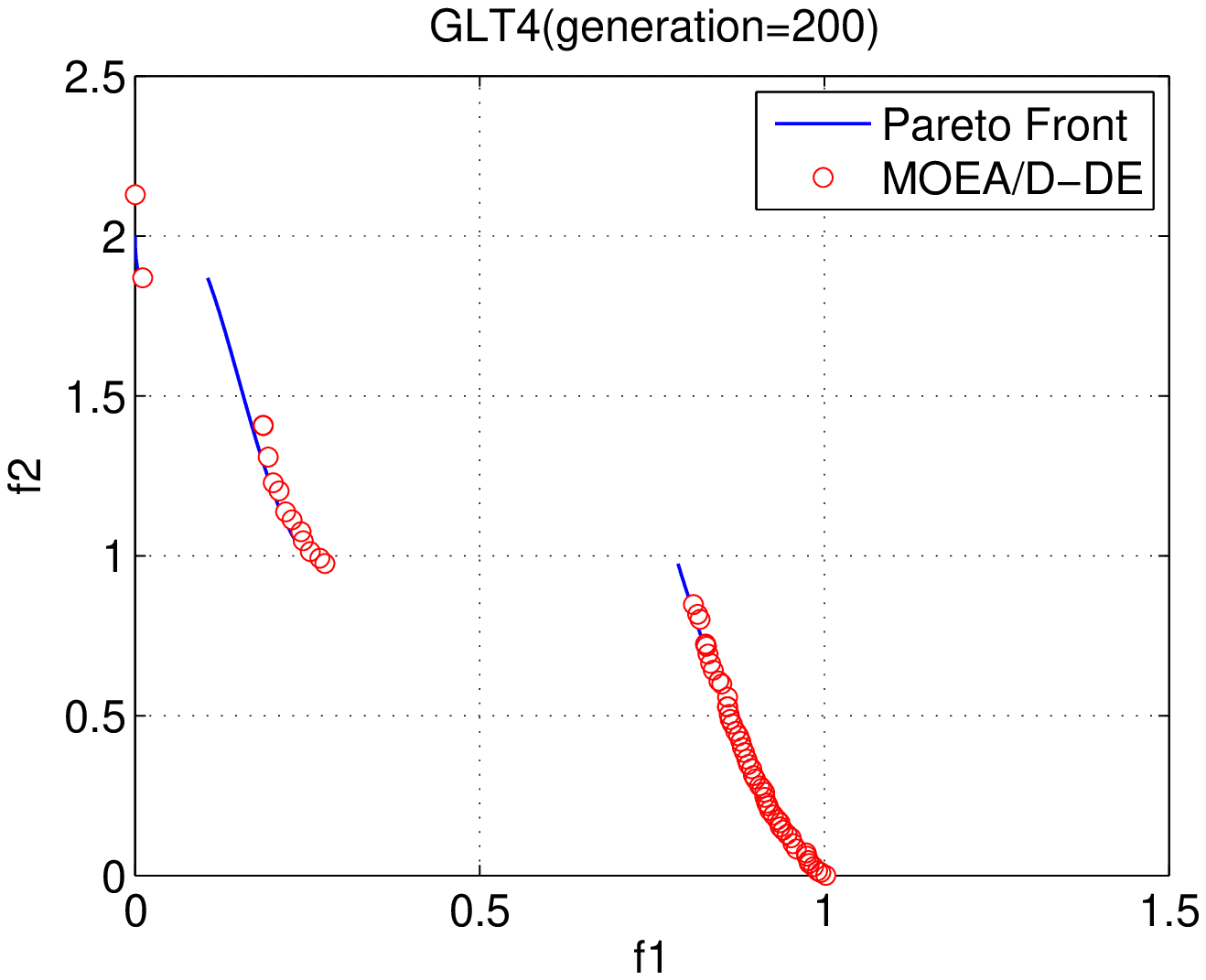}
\includegraphics[width=0.24\textwidth]{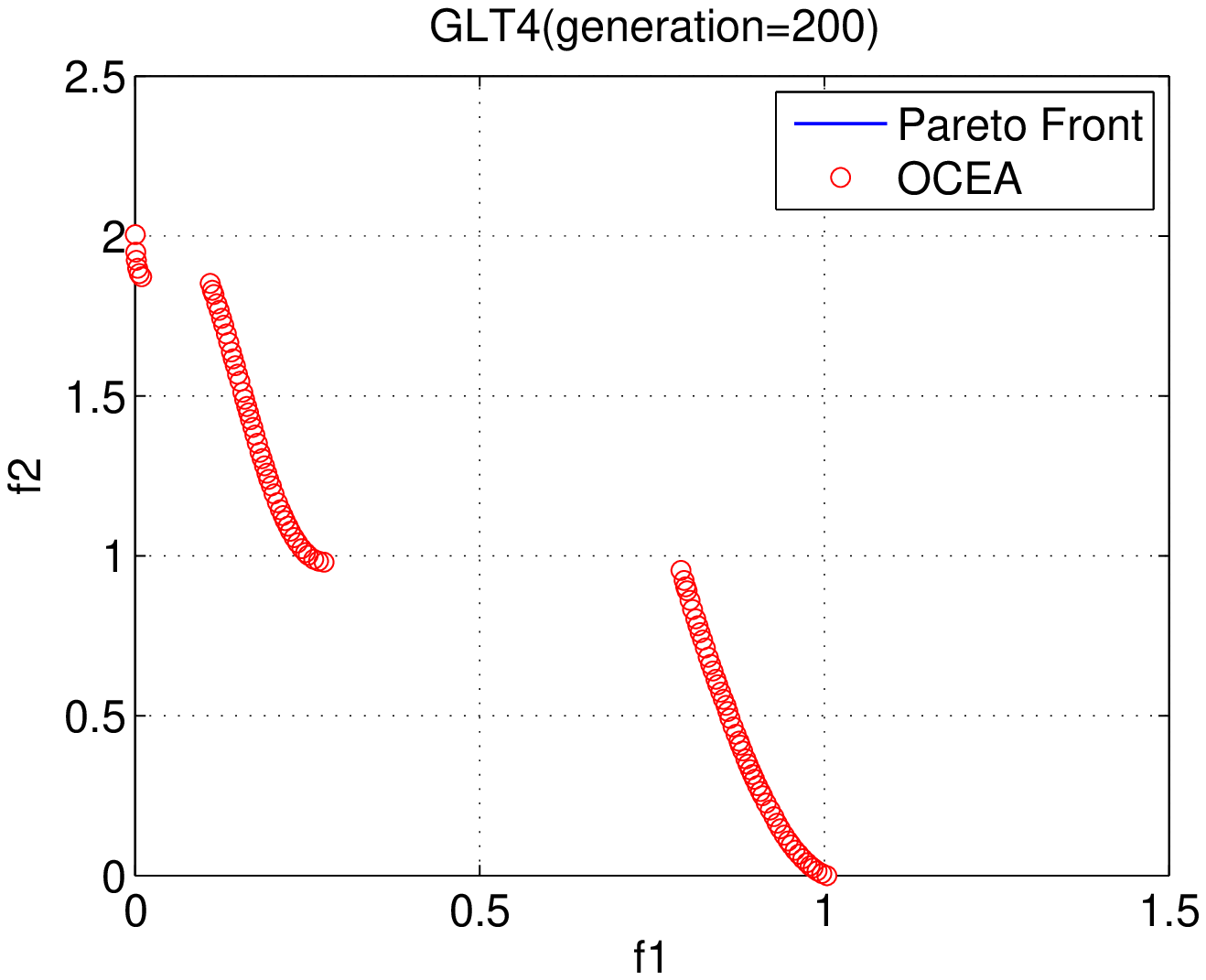}\\
\includegraphics[width=0.24\textwidth]{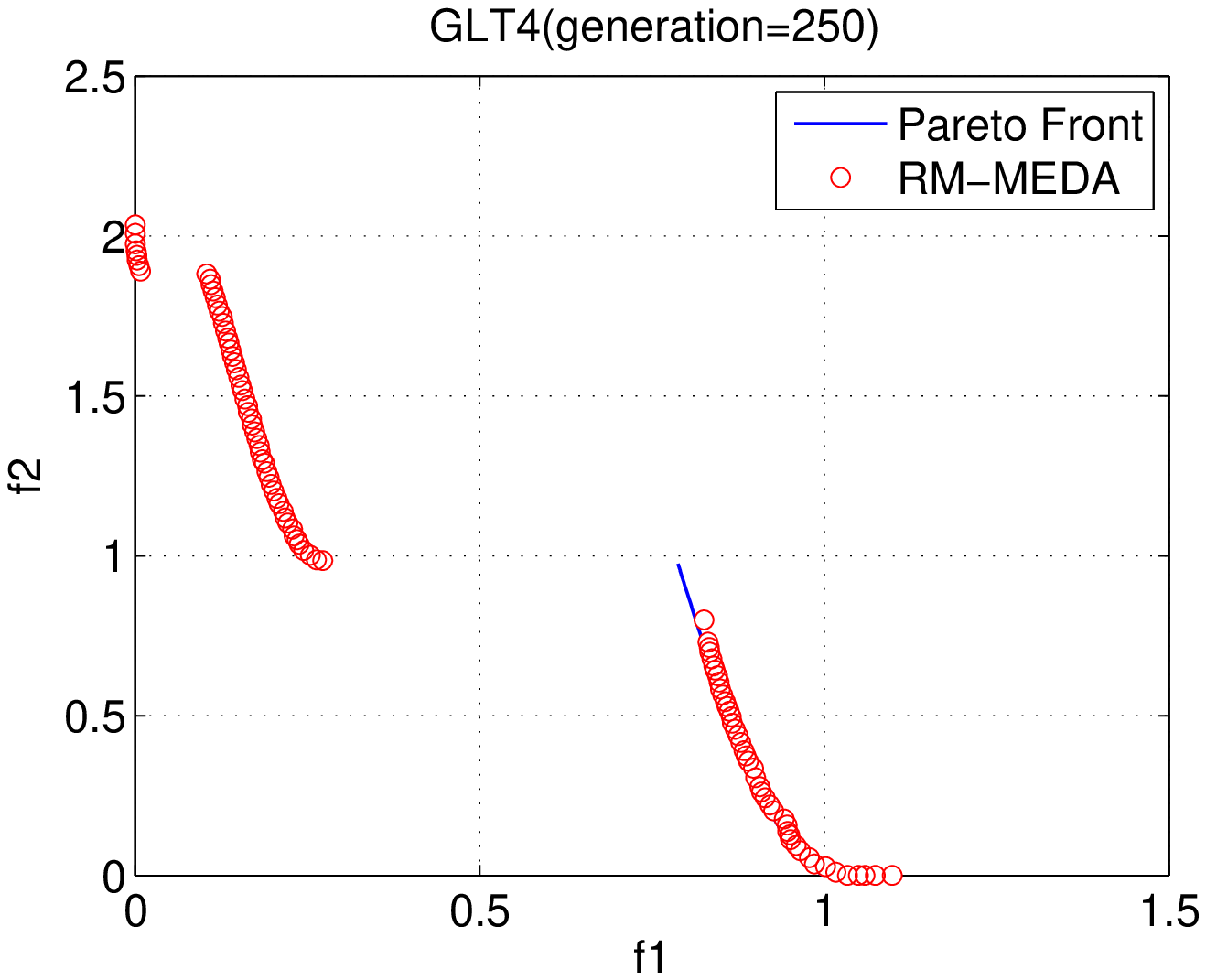}
\includegraphics[width=0.24\textwidth]{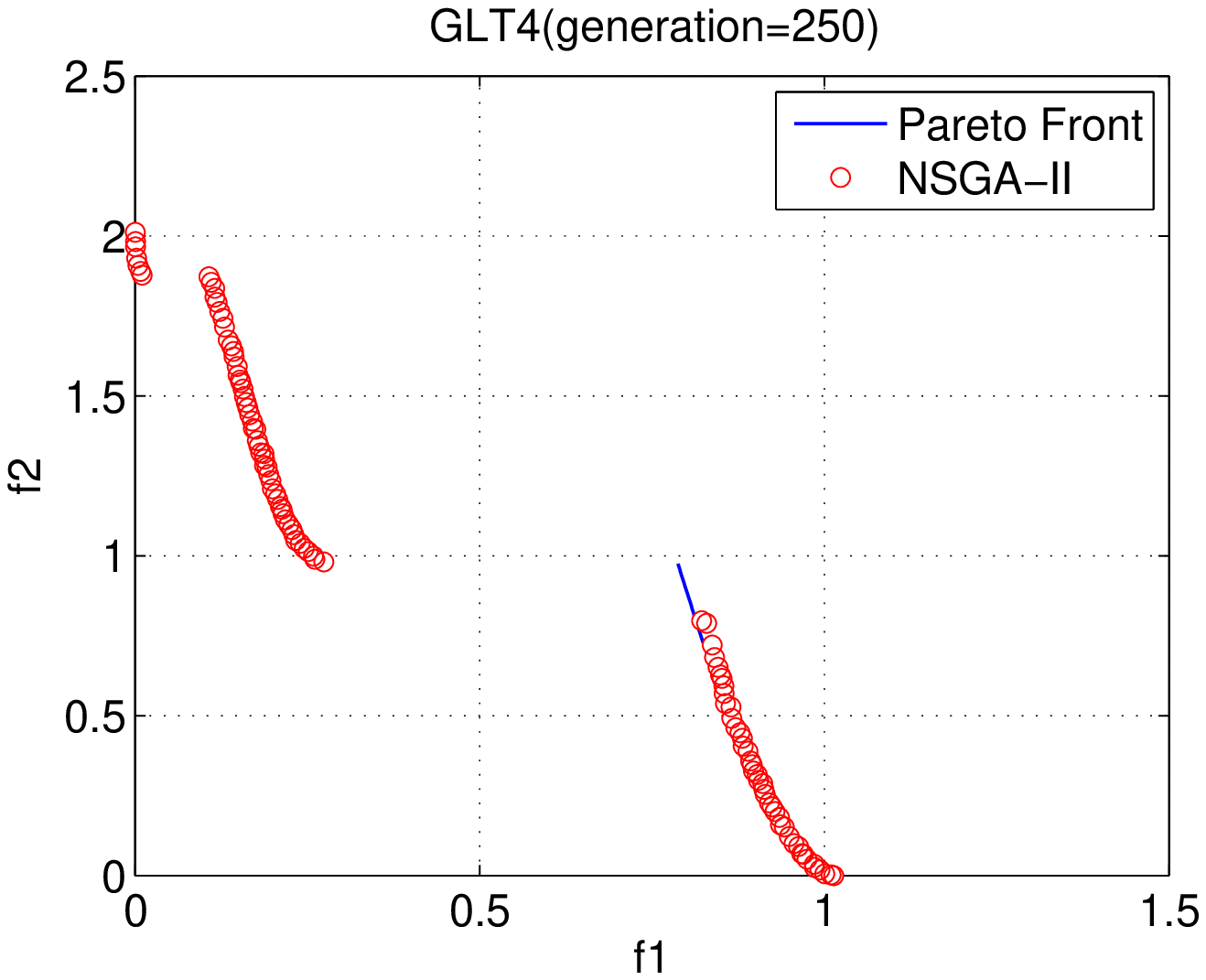}
\includegraphics[width=0.24\textwidth]{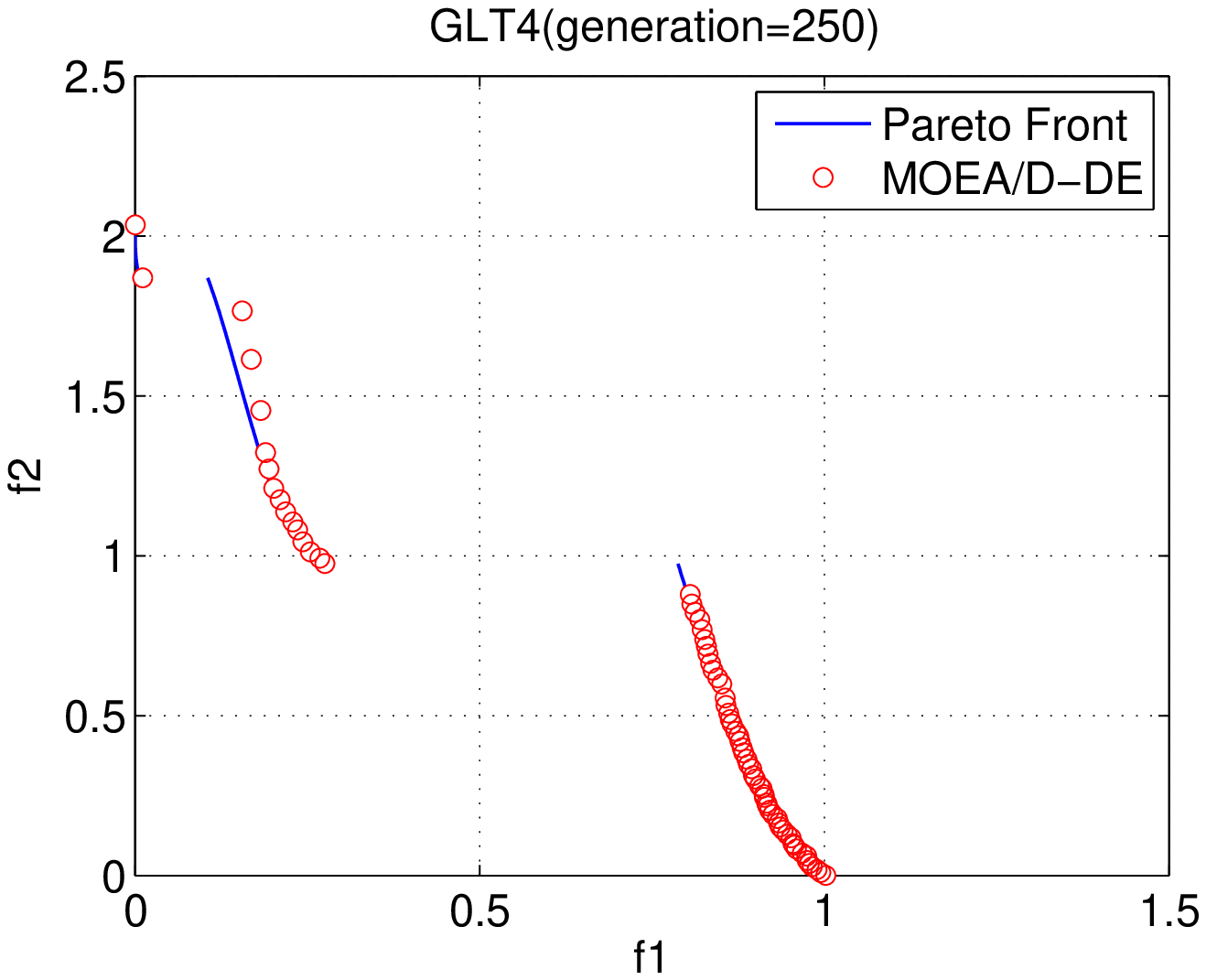}
\includegraphics[width=0.24\textwidth]{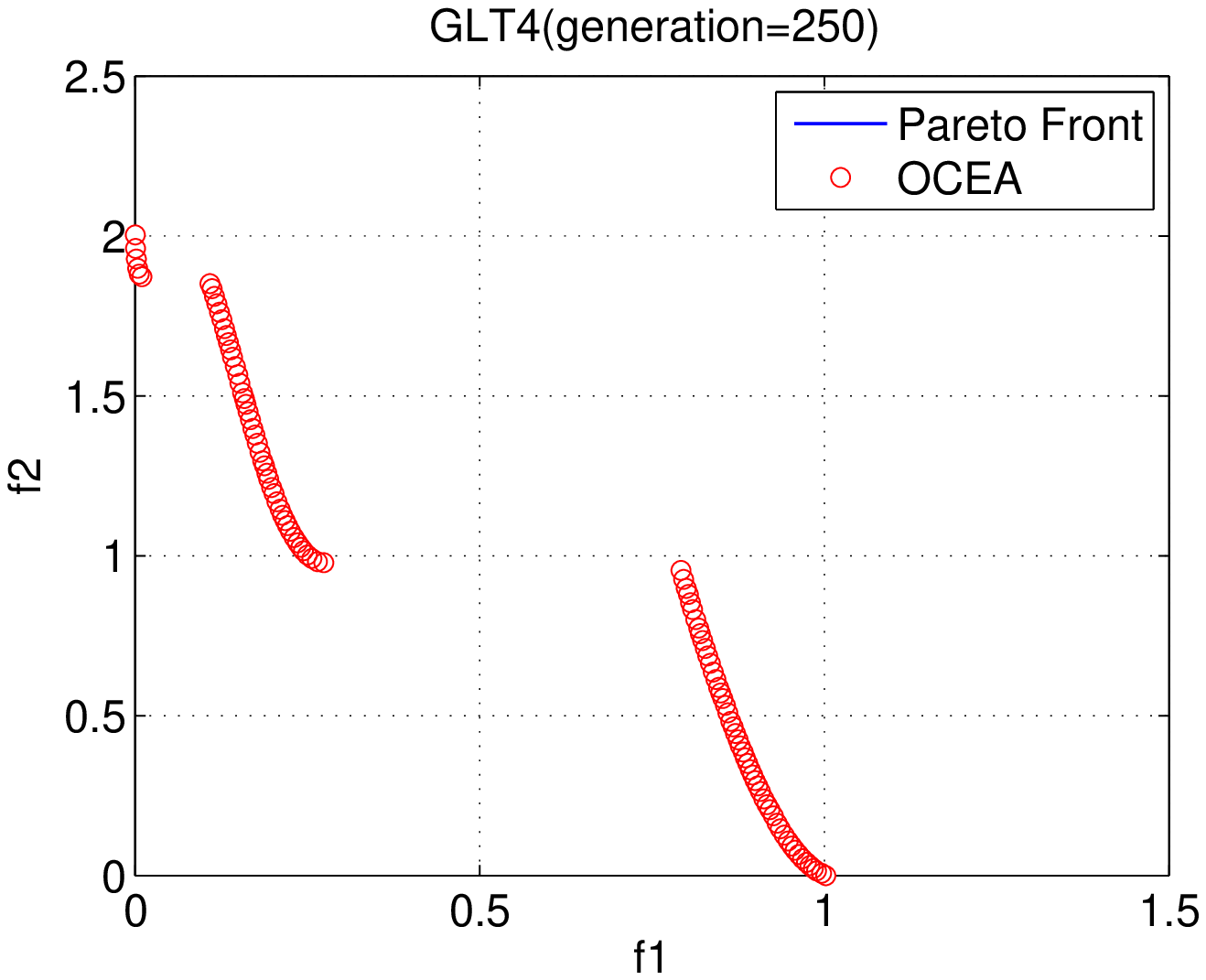}\\
\subfigure[RM-MEDA] {\includegraphics[width=0.24\textwidth]{GLT4_RMMEDA_m6.eps}\label{a}}
\subfigure[NSGA-II] {\includegraphics[width=0.24\textwidth]{GLT4_NSGAIIDE_m6.eps}\label{b}}
\subfigure[MOEA/D-DE] {\includegraphics[width=0.24\textwidth]{GLT4_MOEADDE_m6.eps}\label{c}}
\subfigure[OCEA] {\includegraphics[width=0.24\textwidth]{GLT4_OACDE_m6.eps}\label{d}}
\caption{Evolution of the approximated fronts obtained by RM-MEDA, NSGA-II, MOEA/D-DE and OCEA on GLT4}\label{AFEvo}
\end{figure*}

To reveal the search processes, Fig.~\ref{AFEvo} plots the evolution of the approximated fronts obtained by RM-MEDA, NSGA-II, MOEA/D-DE and OCEA on GLT4. It is noted that the evolution of the approximated front obtained by each algorithm plotted in the figure is representative. The representative evolution of an algorithm here indicates the final approximated front yielded by the evolution is with the median IGD metric value in 33 independent runs. It can be seen from the figure that, at the 100th generation, the approximated front yielded by OCEA has reached the PF completely, and almost covered the whole PF. After 300 generations, it has reached the approximated front with excellent convergence and diversity. On the other hand, after 300 generations, the final approximated fronts obtained by RM-MEDA, NSGA-II, MOEA/D-DE still cannot cover the whole PF, are not distributed unevenly. Fig.~\ref{AFEvo} shows that OCEA can indeed greatly improve the search efficiency.

\begin{figure*}[htbp]
\centering
\includegraphics[width=0.24\textwidth]{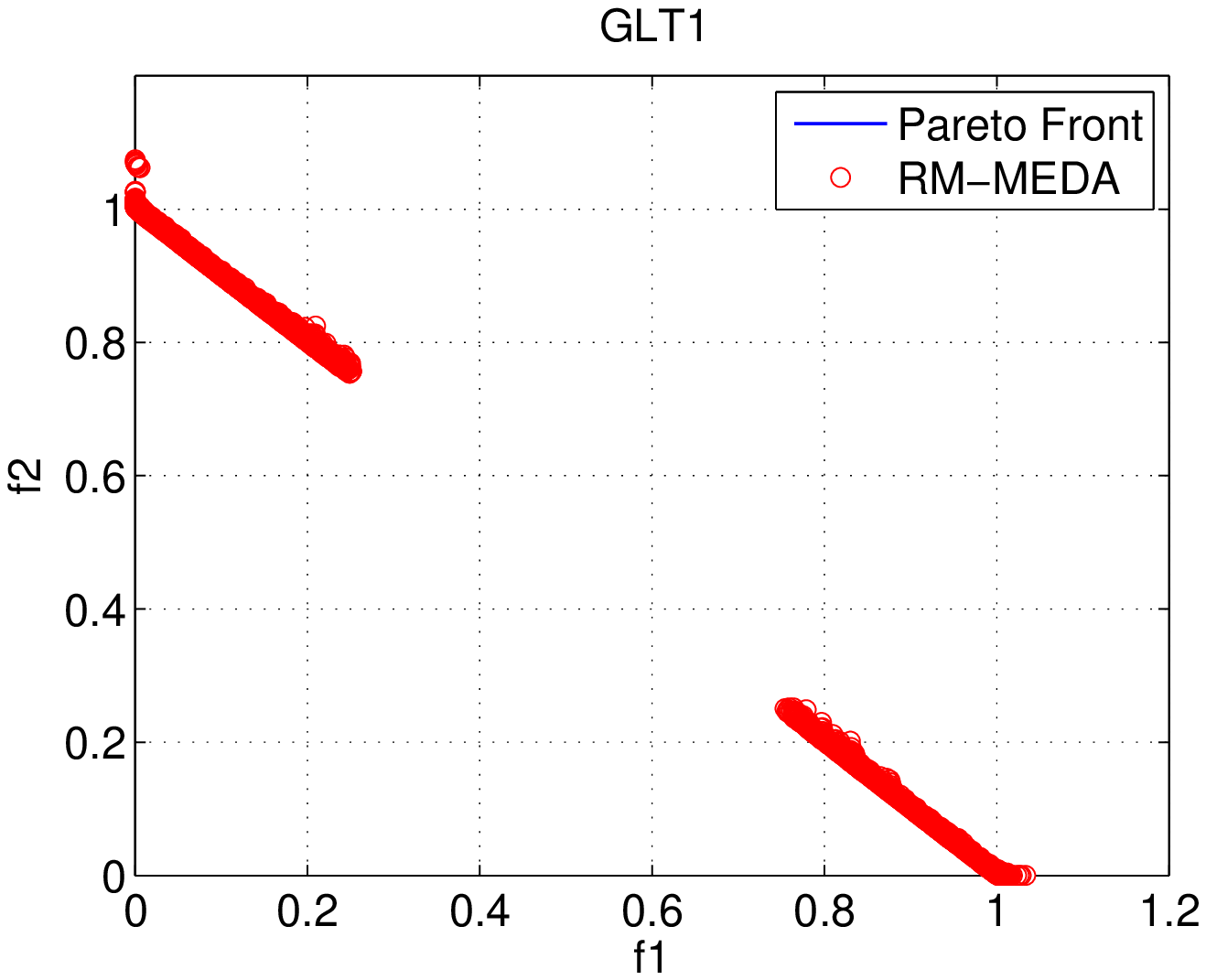}
\includegraphics[width=0.24\textwidth]{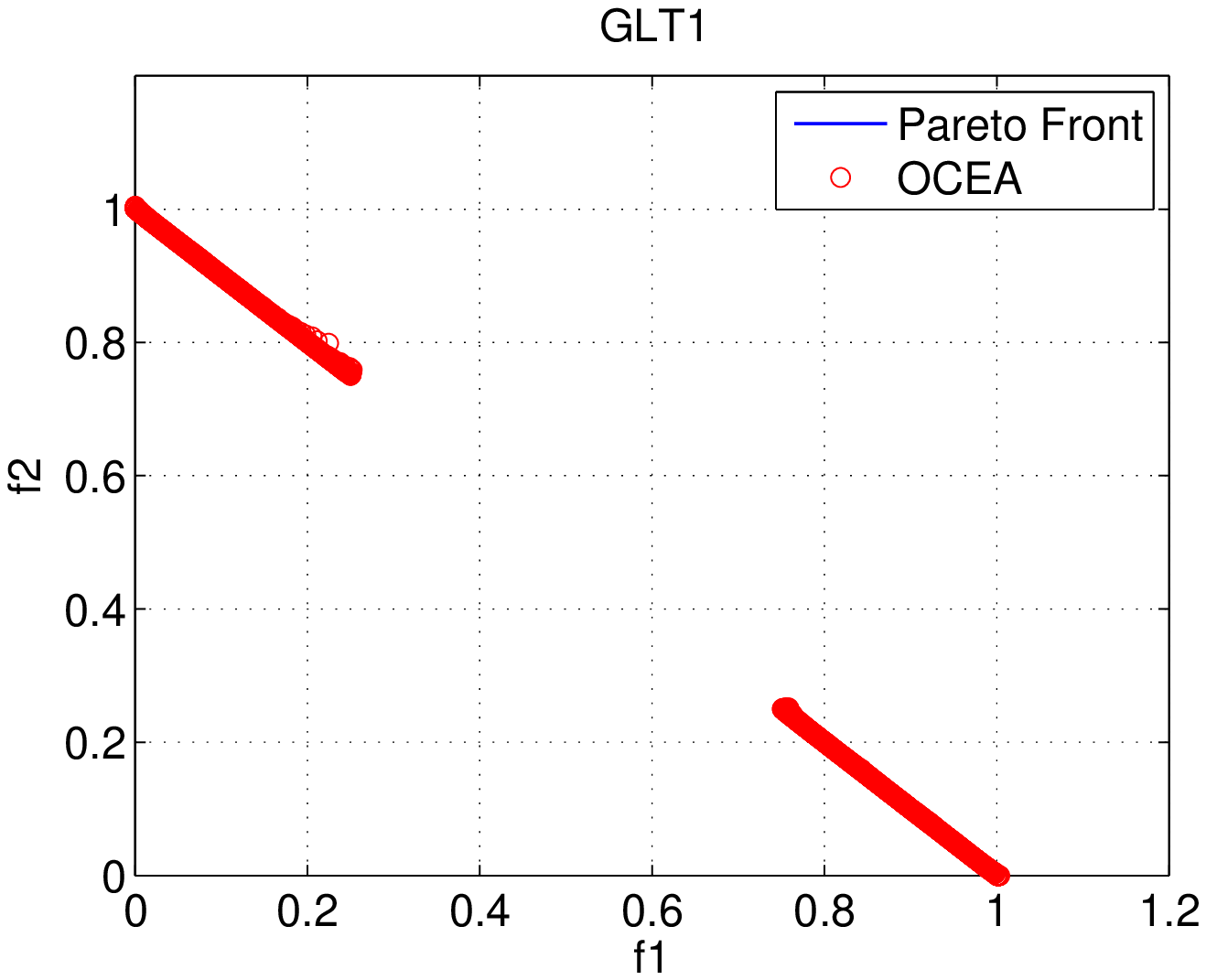}
\includegraphics[width=0.24\textwidth]{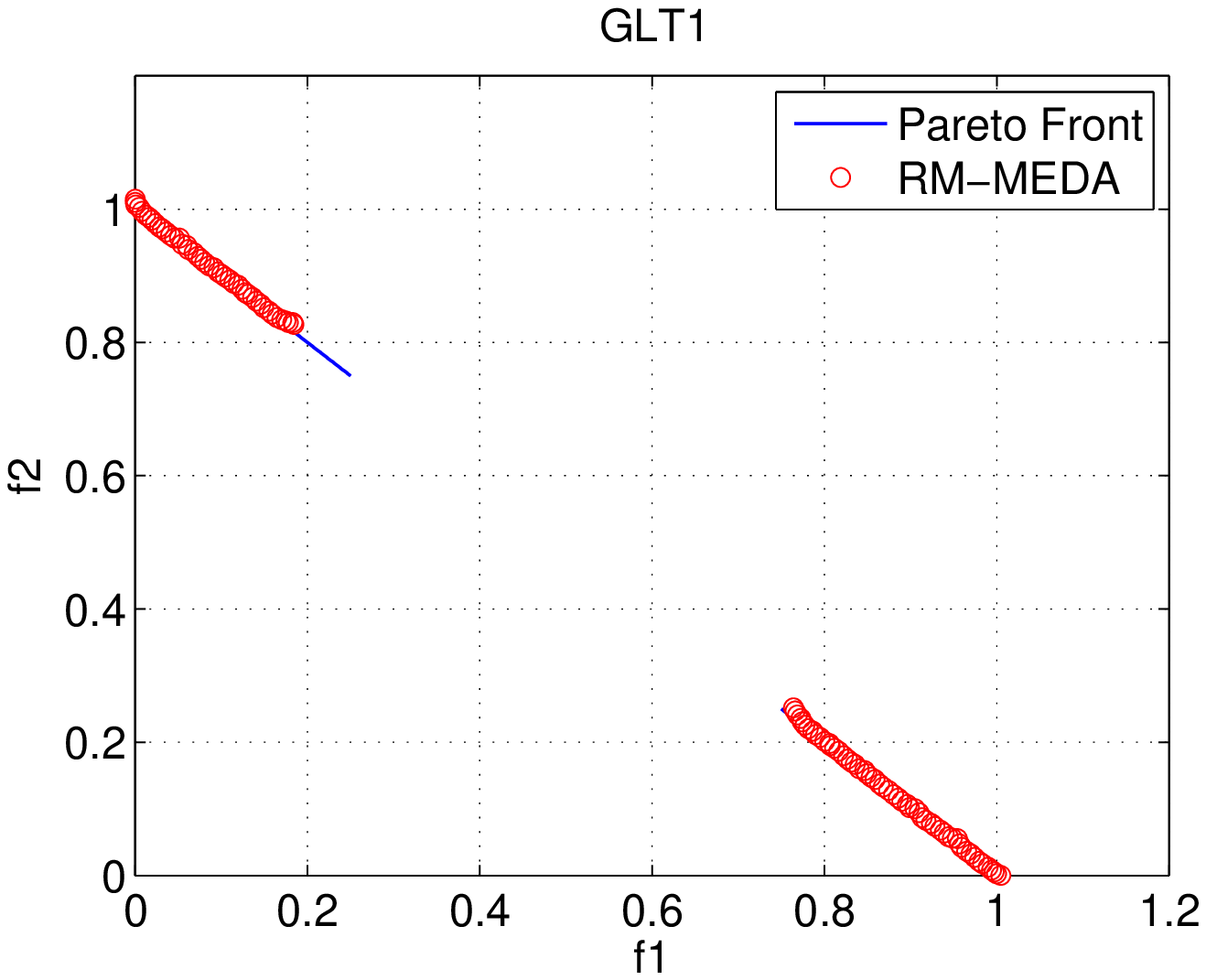}
\includegraphics[width=0.24\textwidth]{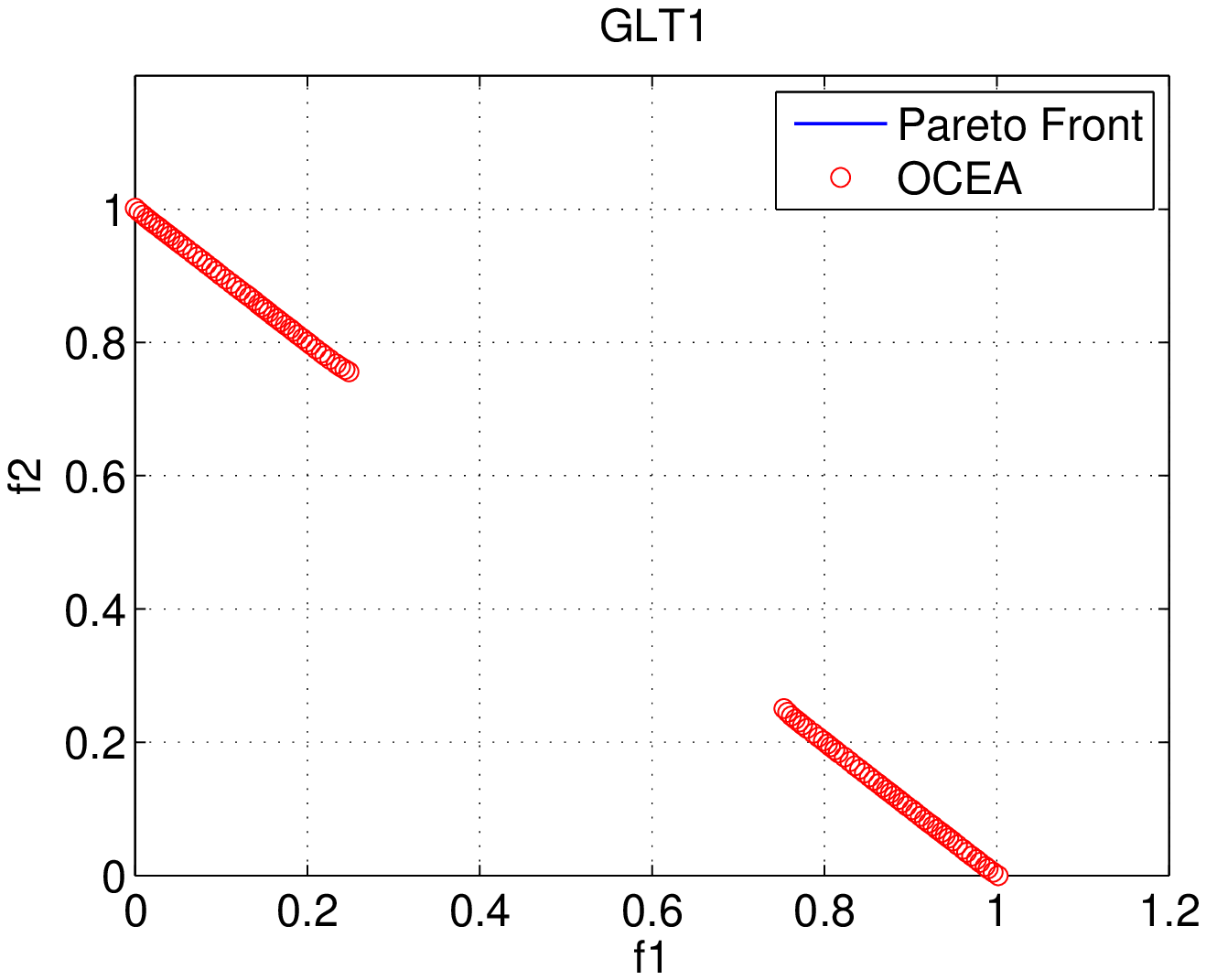}\\
\includegraphics[width=0.24\textwidth]{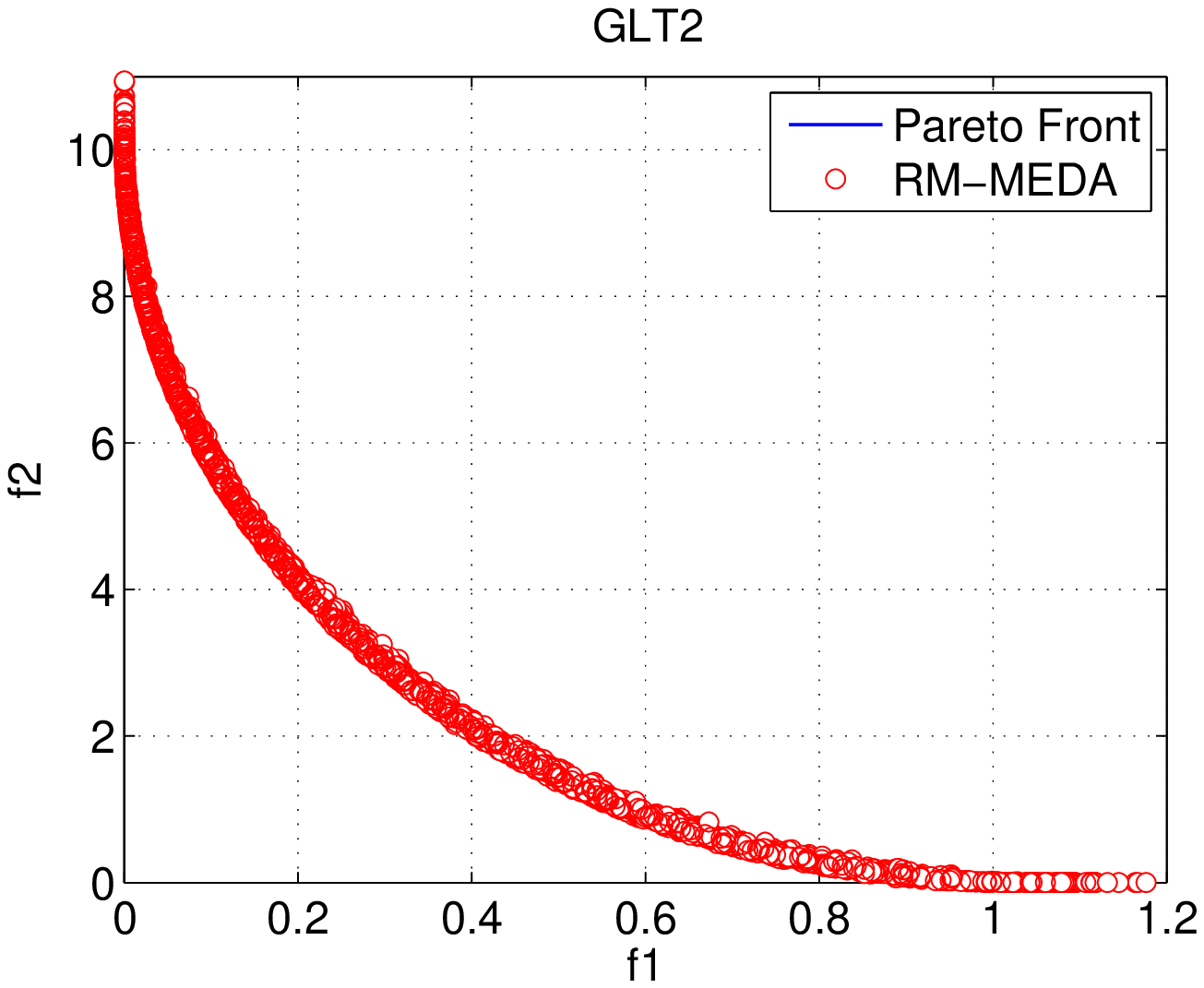}
\includegraphics[width=0.24\textwidth]{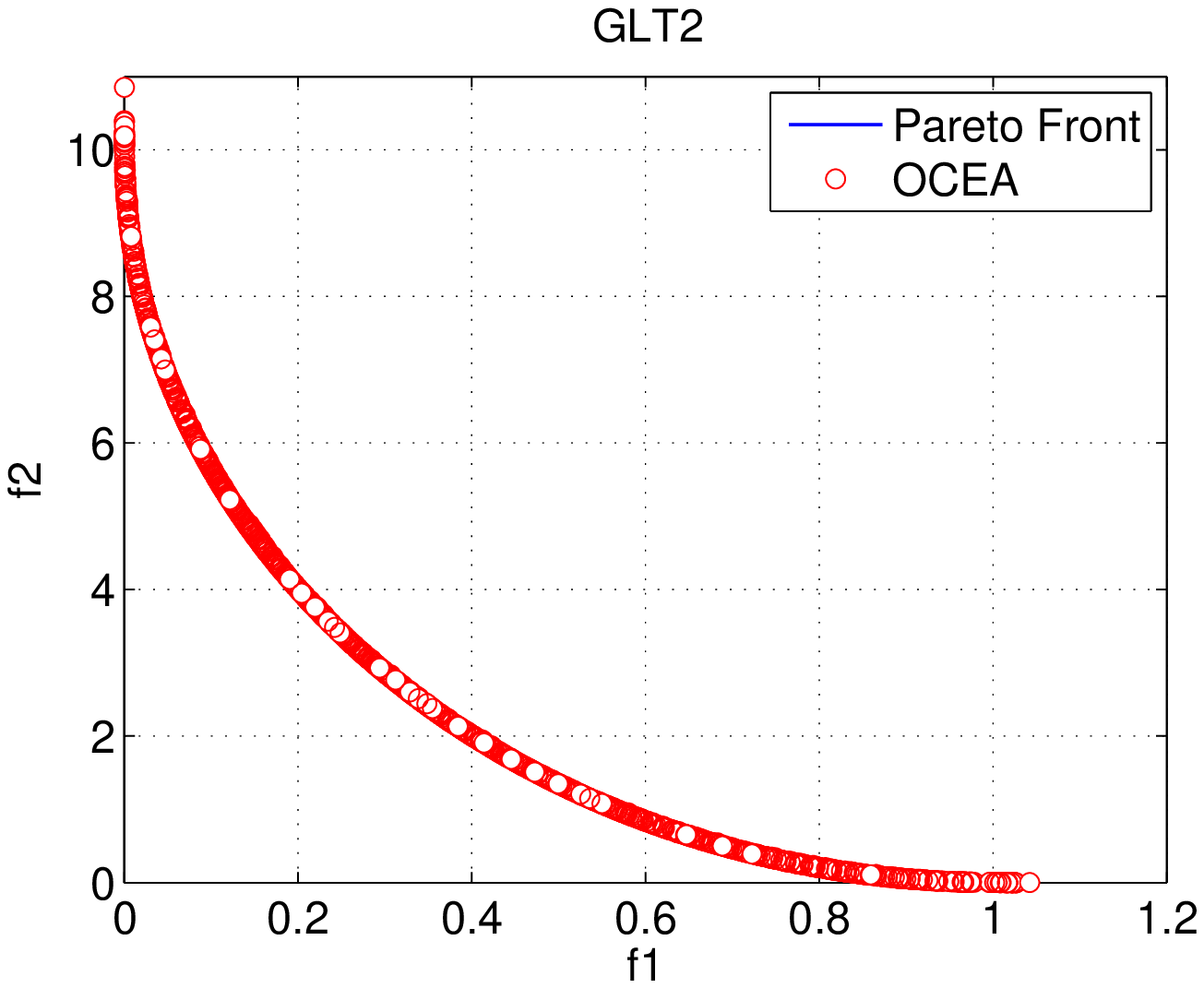}
\includegraphics[width=0.24\textwidth]{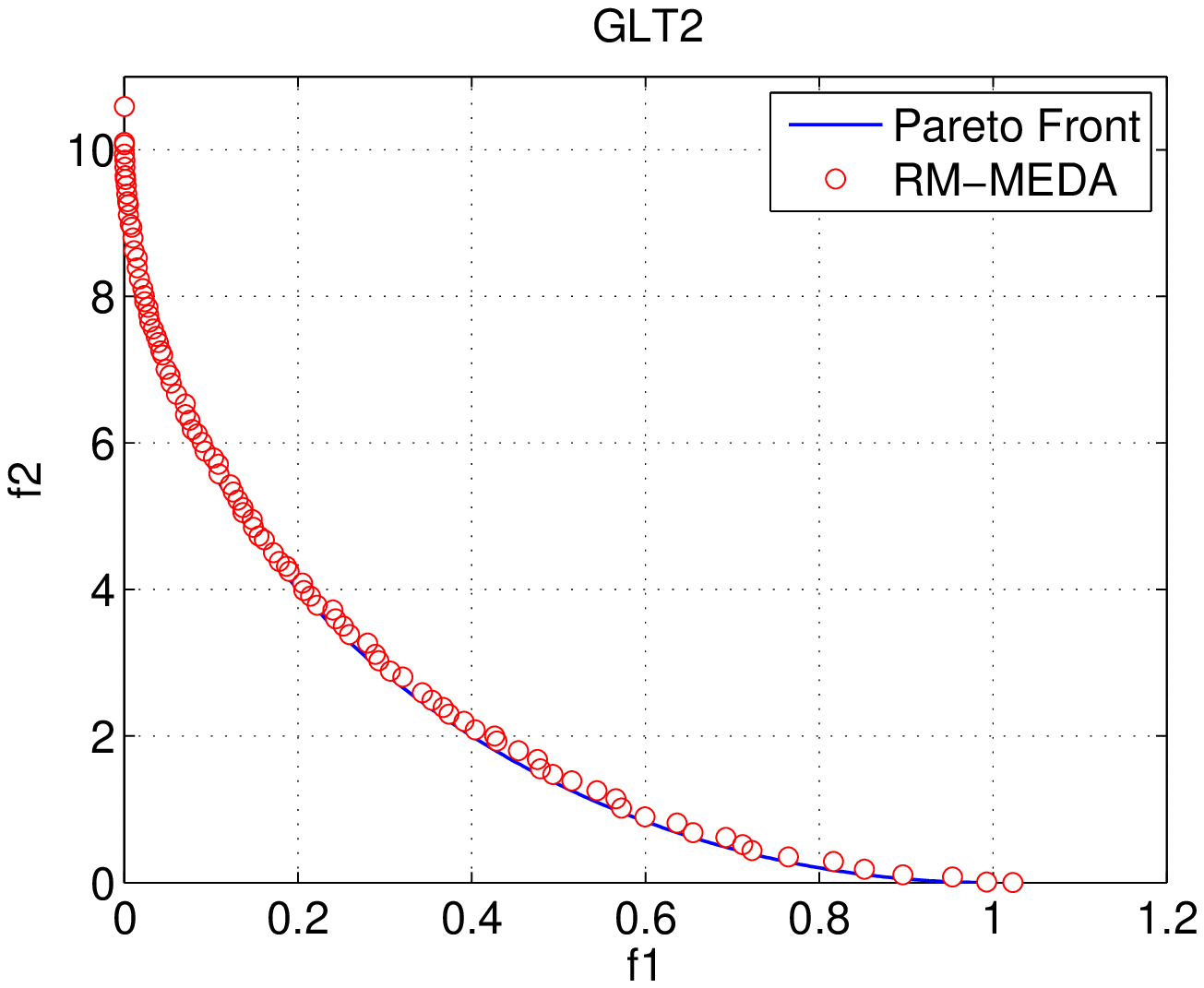}
\includegraphics[width=0.24\textwidth]{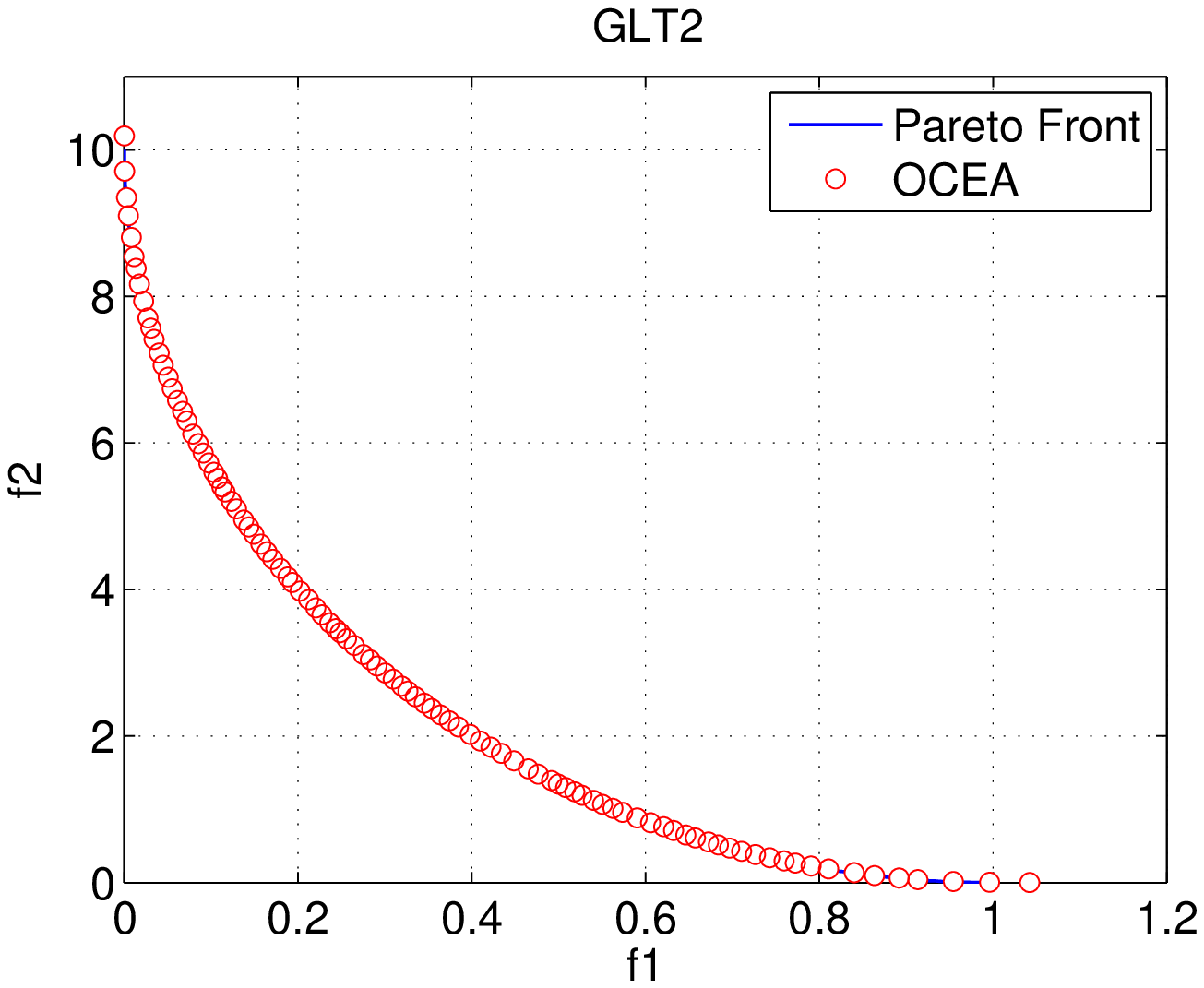}\\
\includegraphics[width=0.24\textwidth]{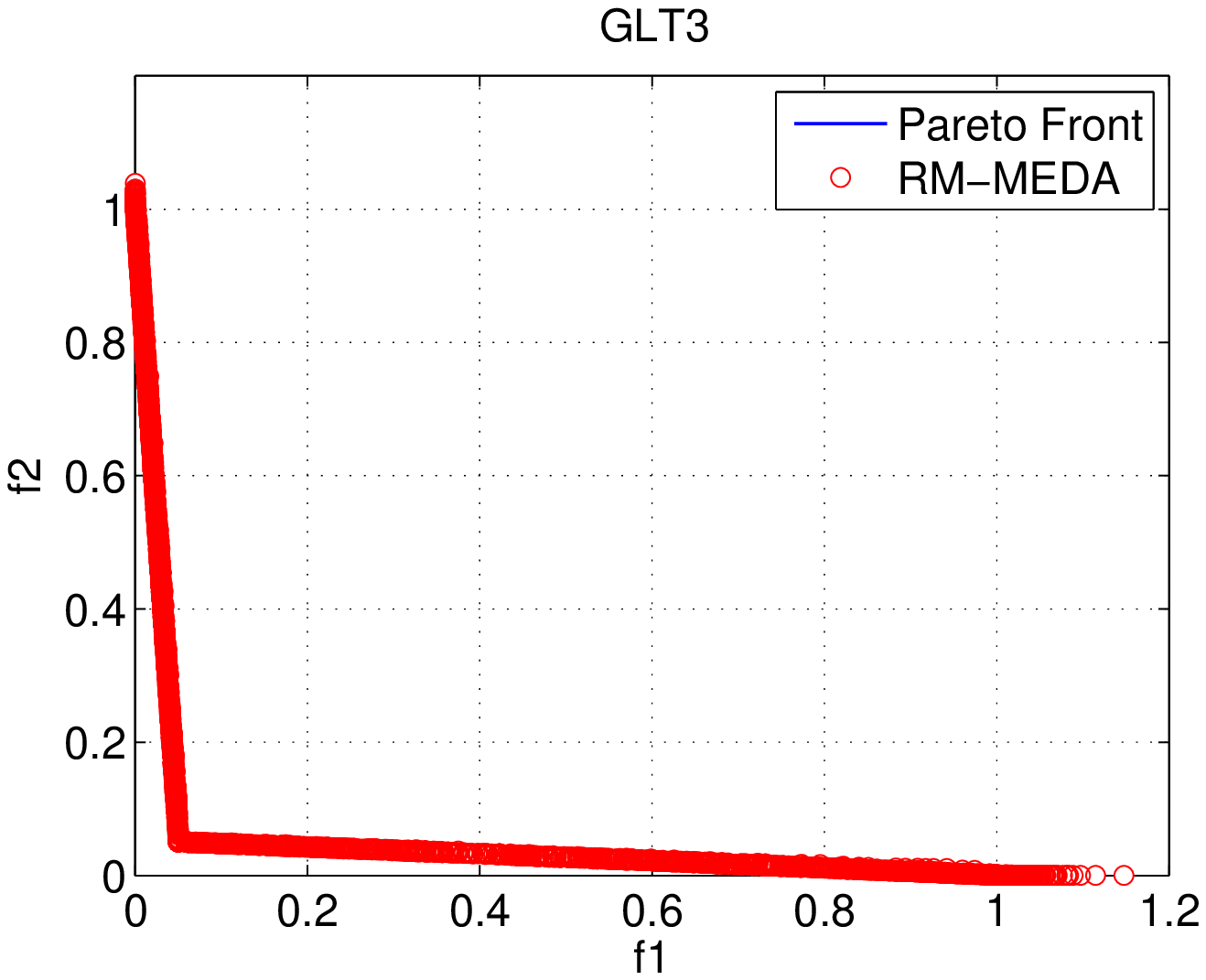}
\includegraphics[width=0.24\textwidth]{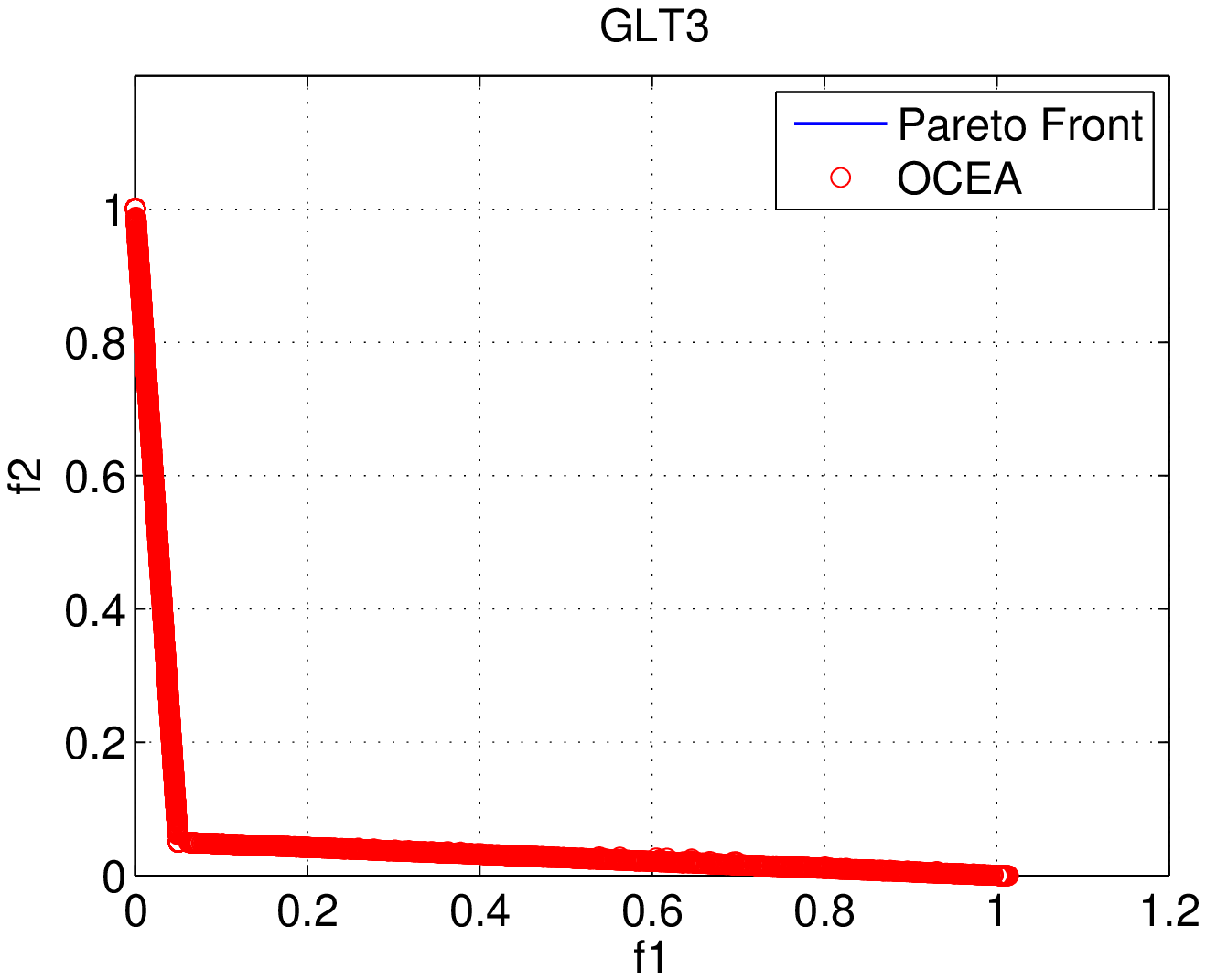}
\includegraphics[width=0.24\textwidth]{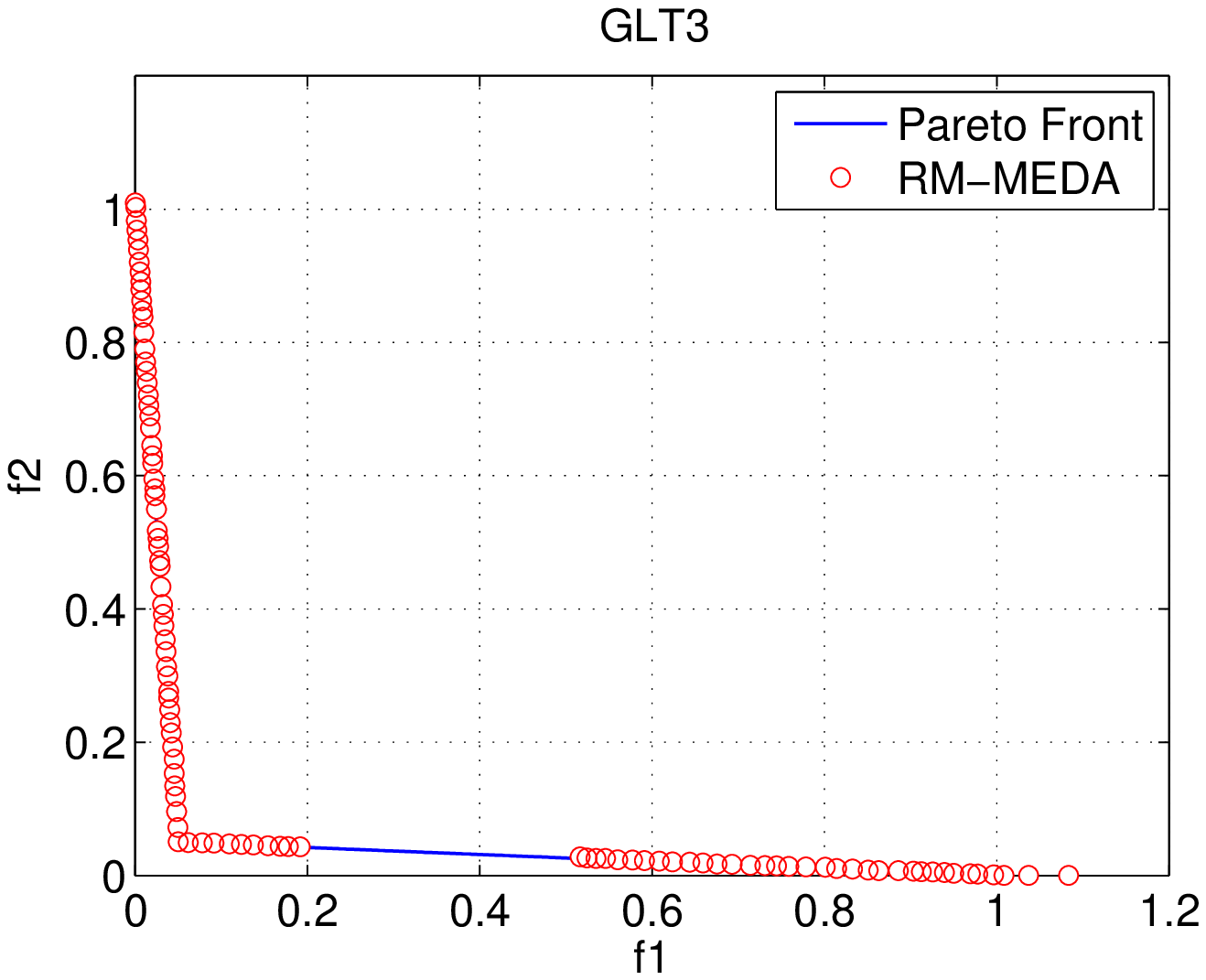}
\includegraphics[width=0.24\textwidth]{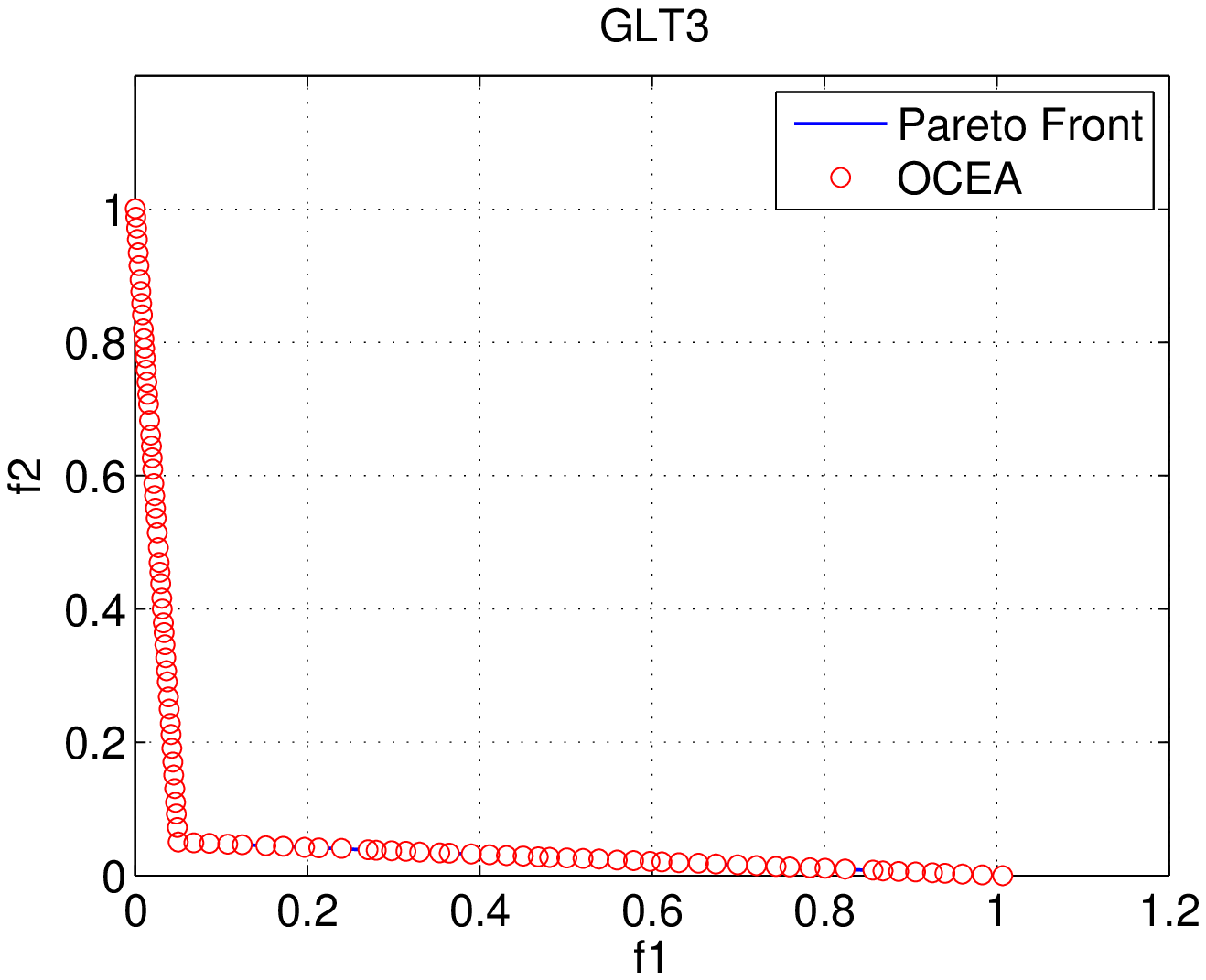}\\
\includegraphics[width=0.24\textwidth]{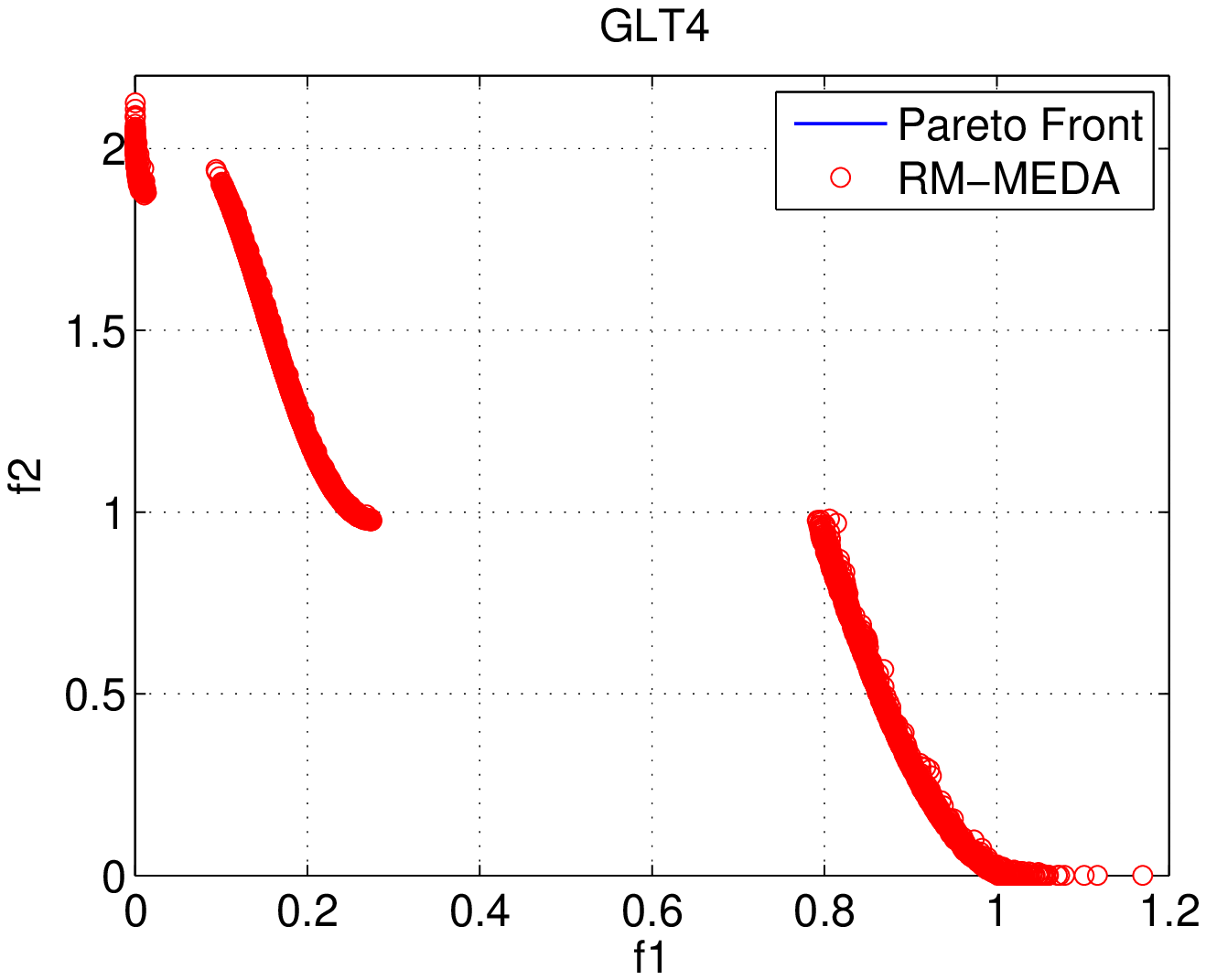}
\includegraphics[width=0.24\textwidth]{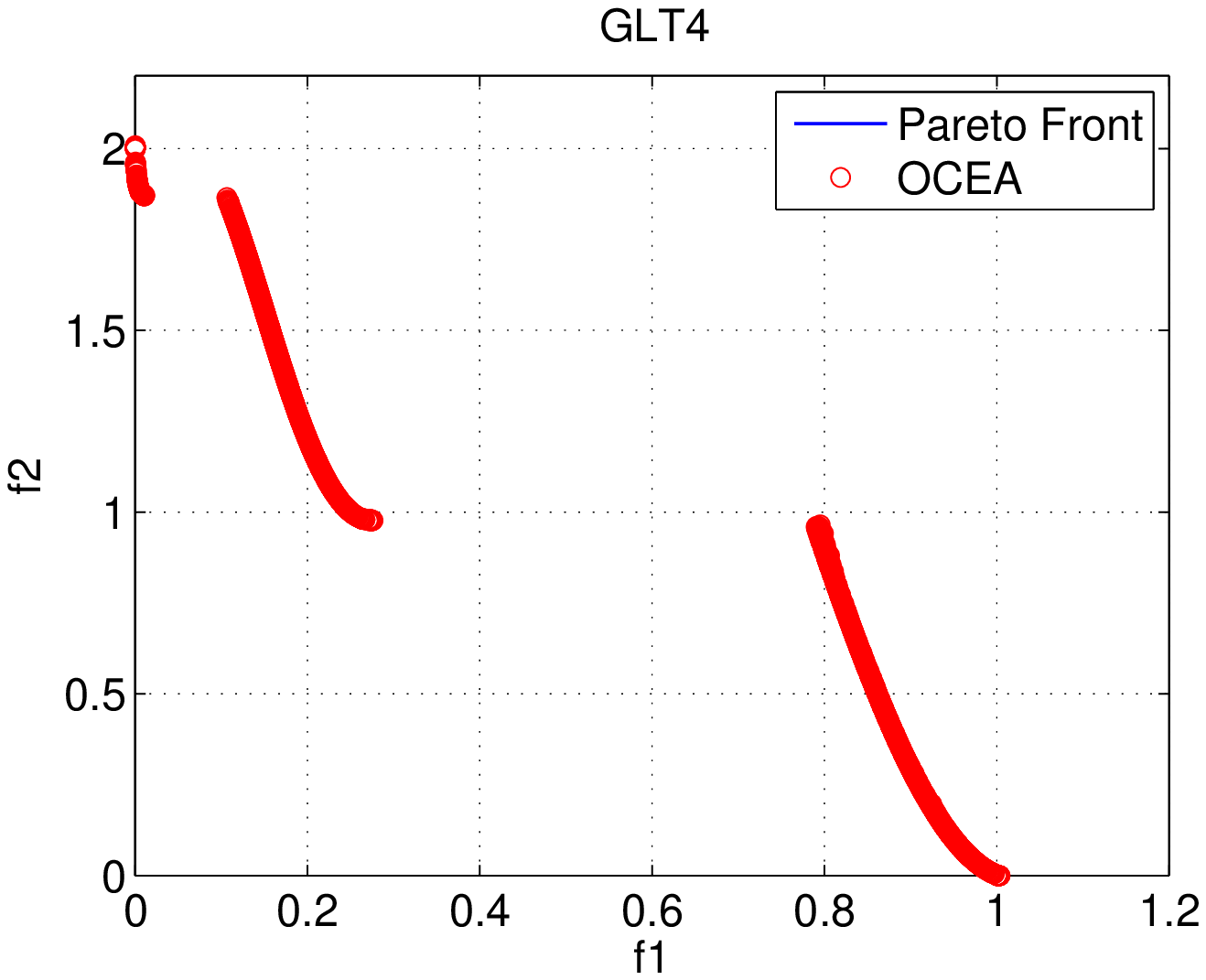}
\includegraphics[width=0.24\textwidth]{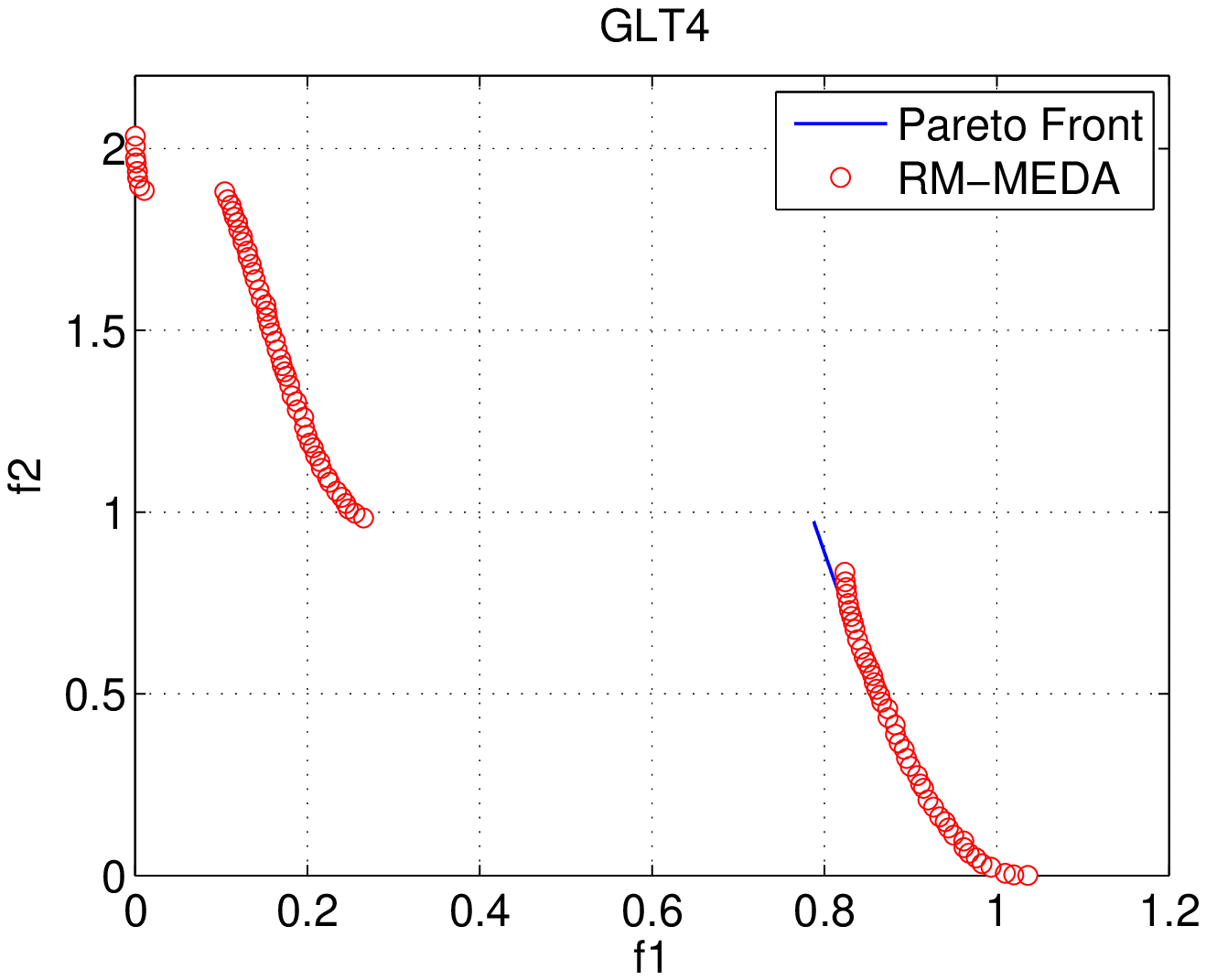}
\includegraphics[width=0.24\textwidth]{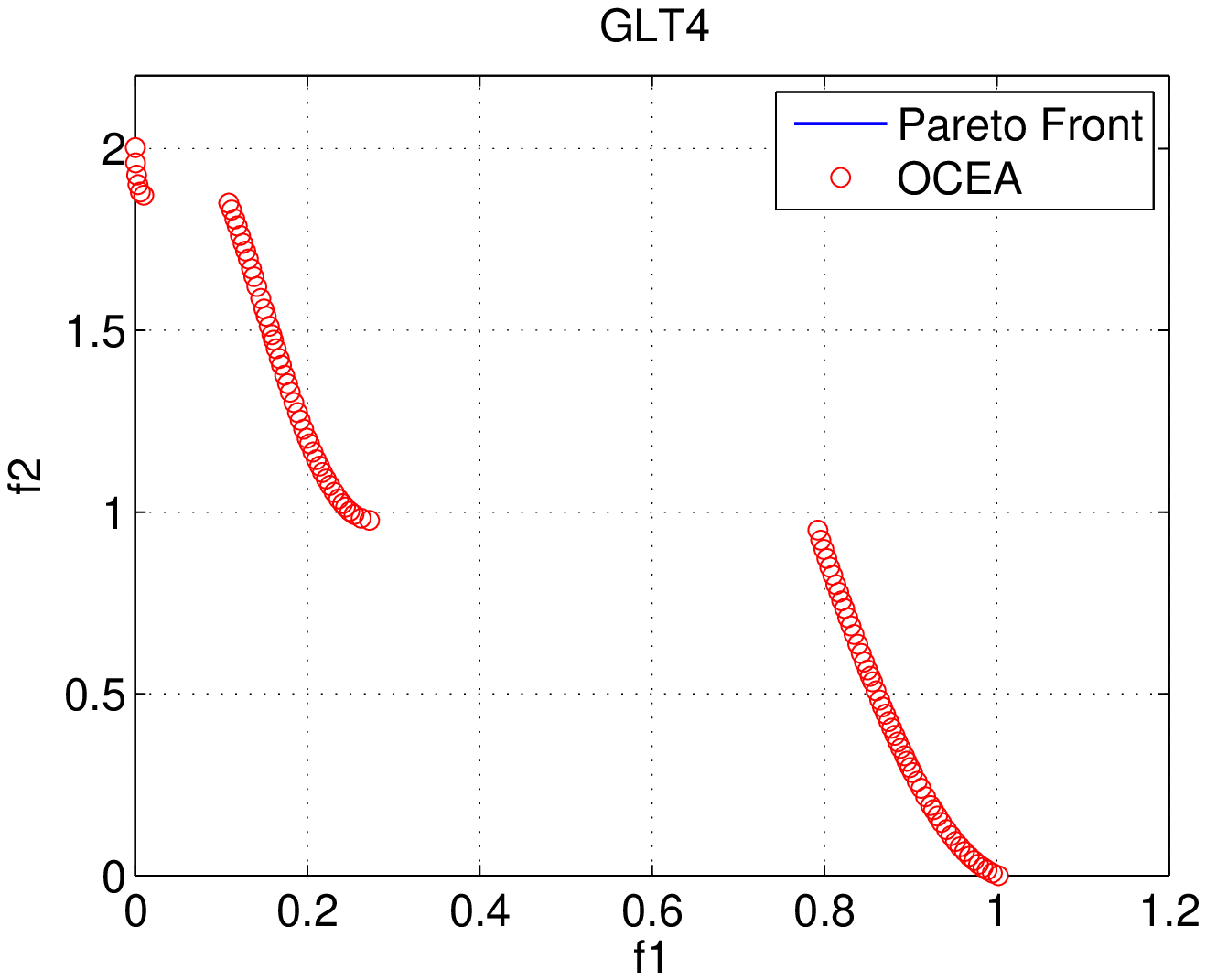}\\
\includegraphics[width=0.24\textwidth]{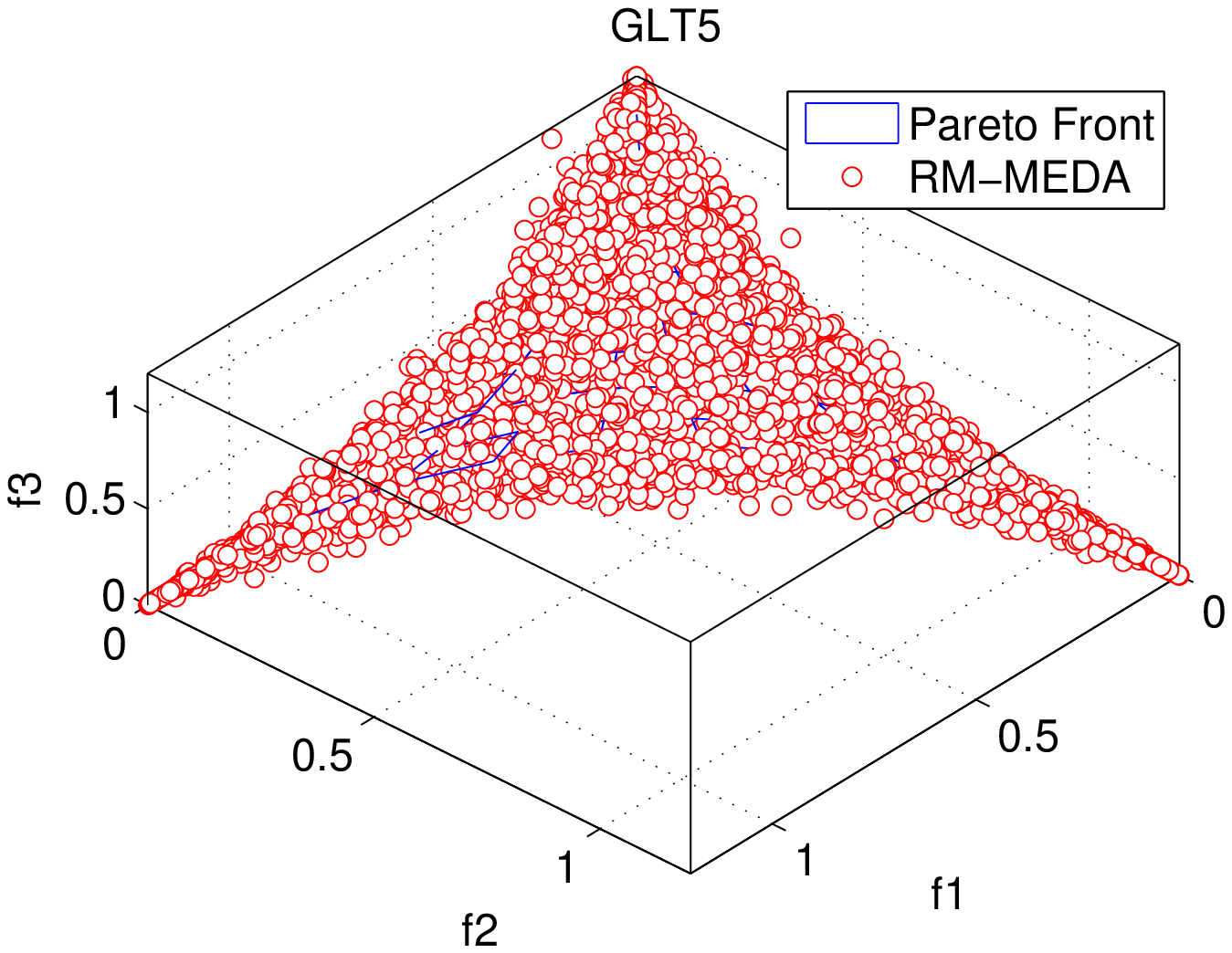}
\includegraphics[width=0.24\textwidth]{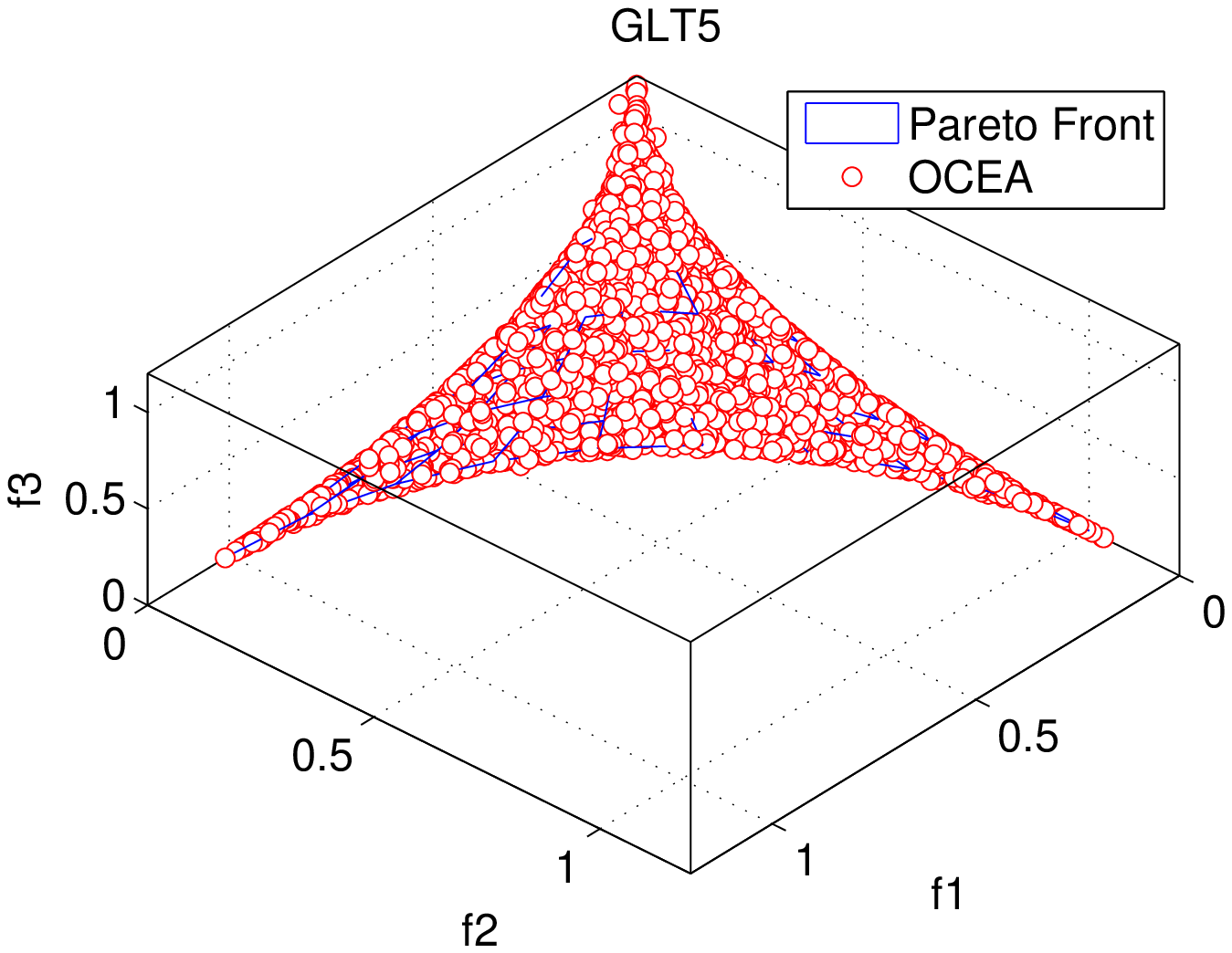}
\includegraphics[width=0.24\textwidth]{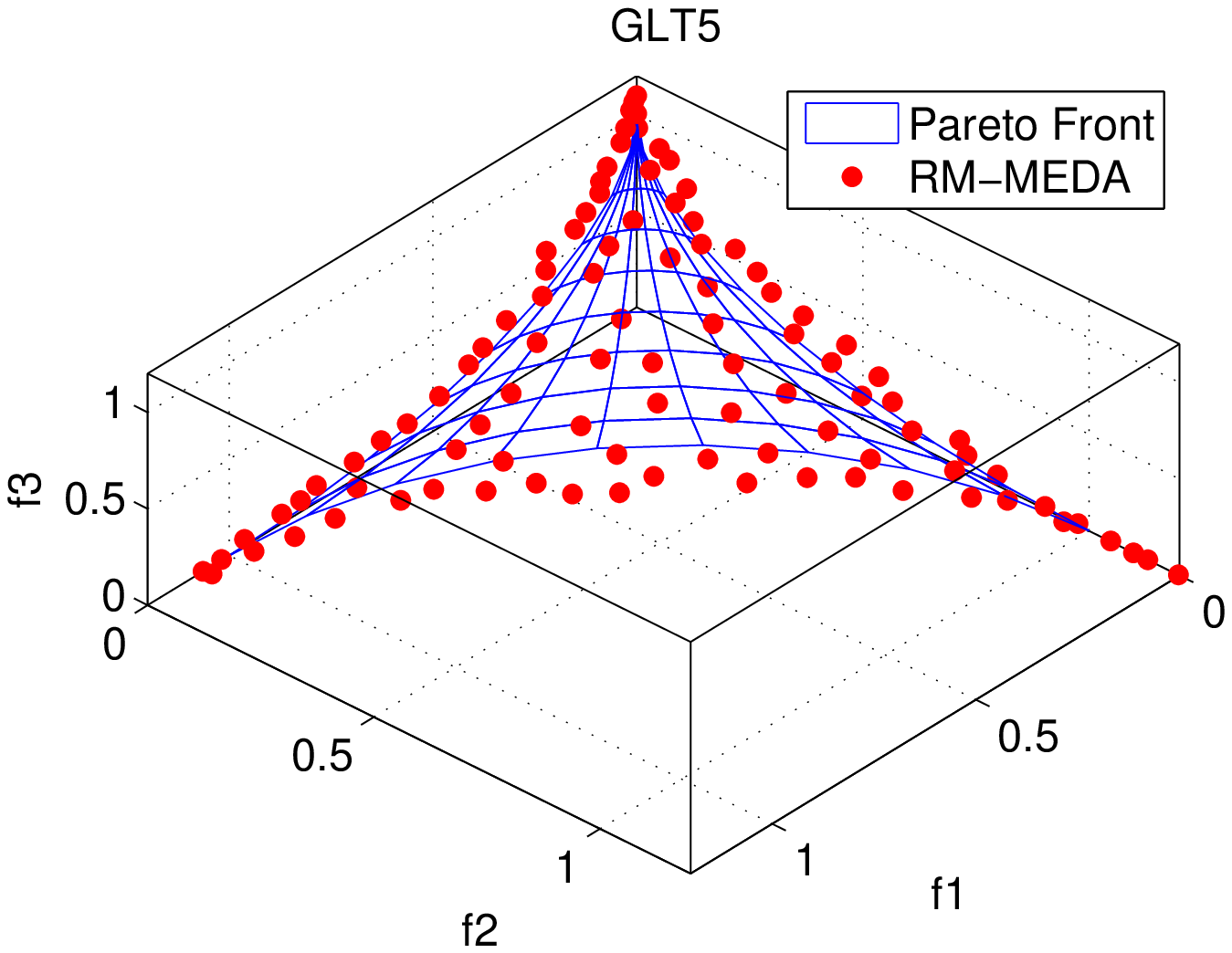}
\includegraphics[width=0.24\textwidth]{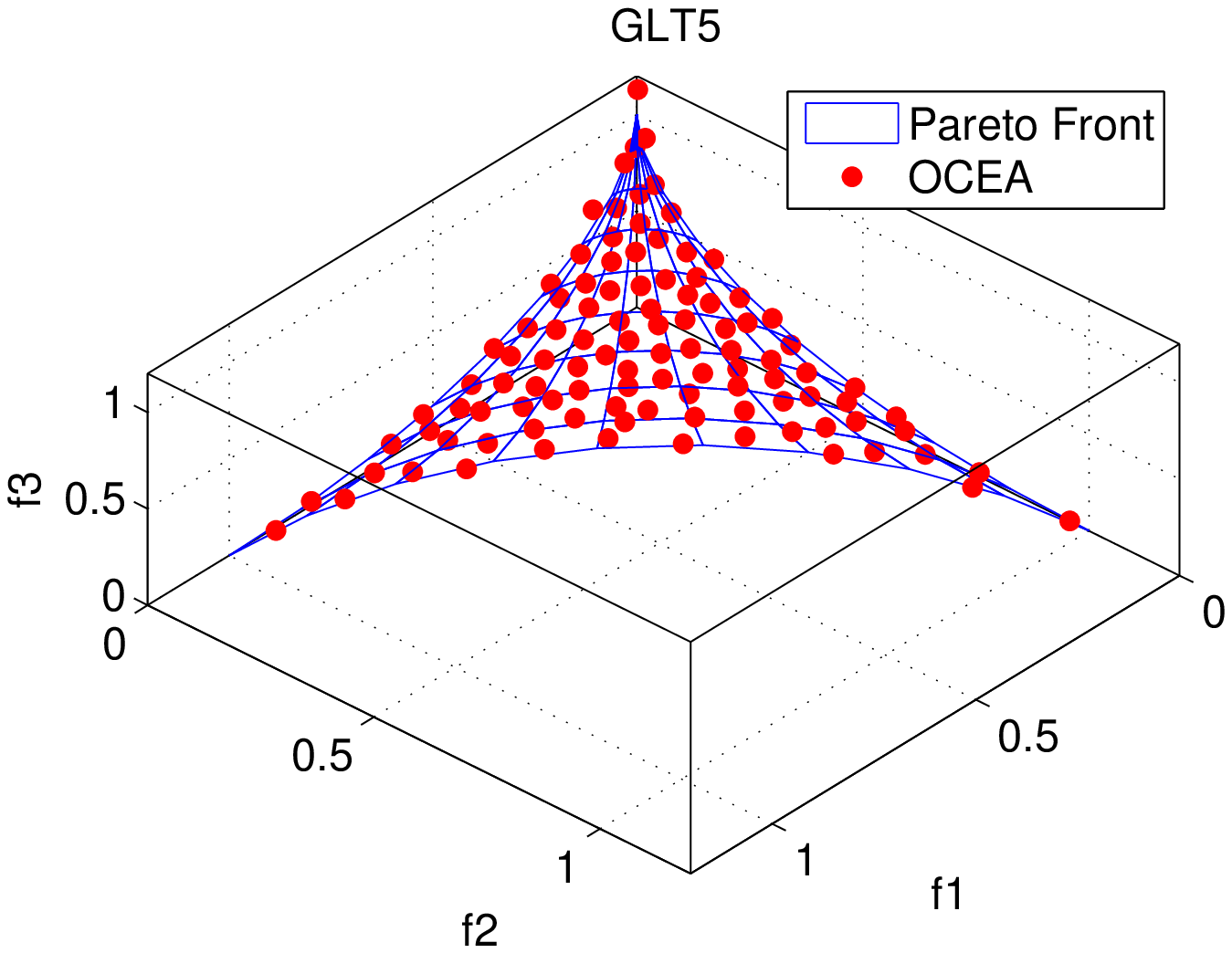}\\
\subfigure[overall fronts] {\includegraphics[width=0.24\textwidth]{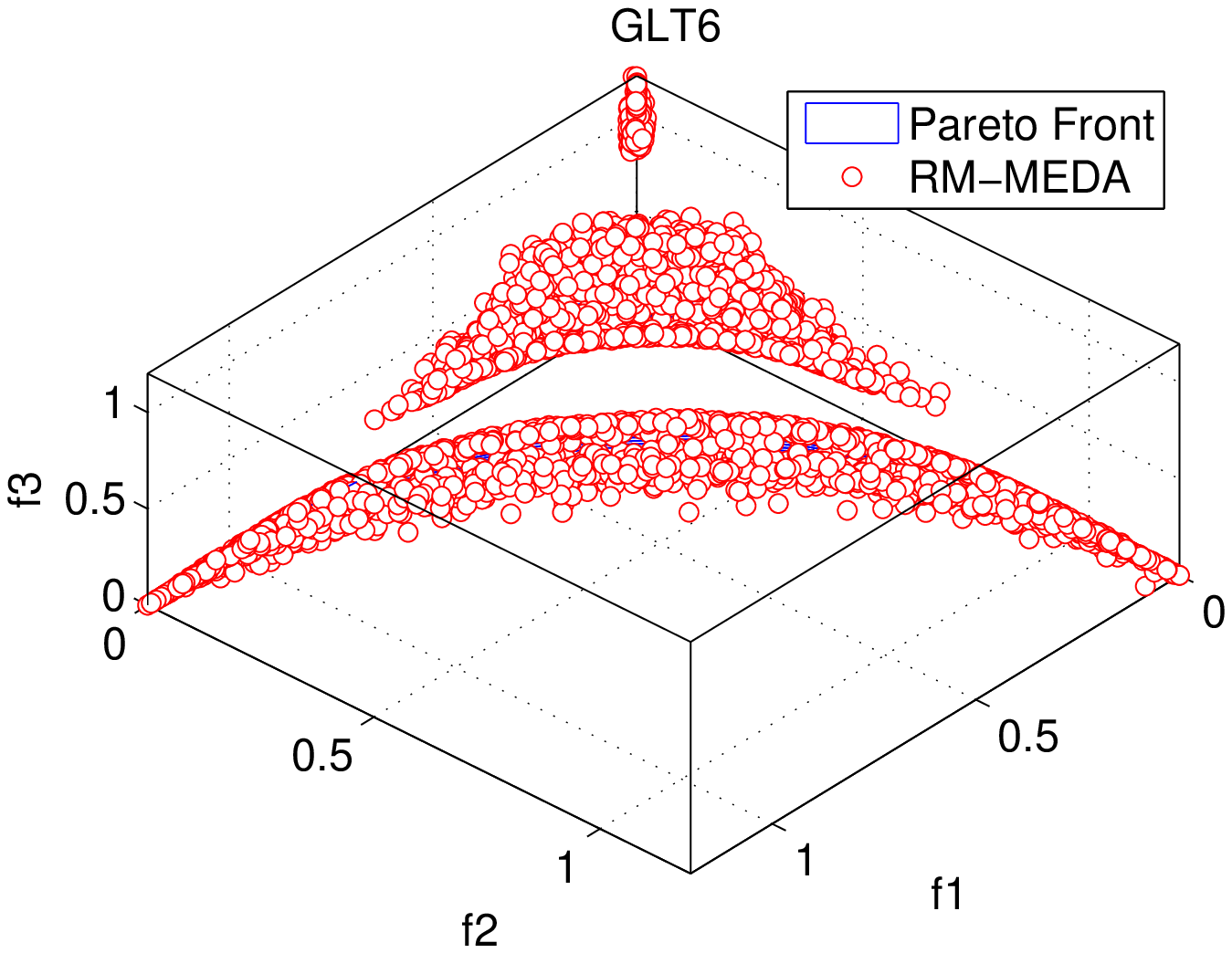}\label{a}}
\subfigure[overall fronts] {\includegraphics[width=0.24\textwidth]{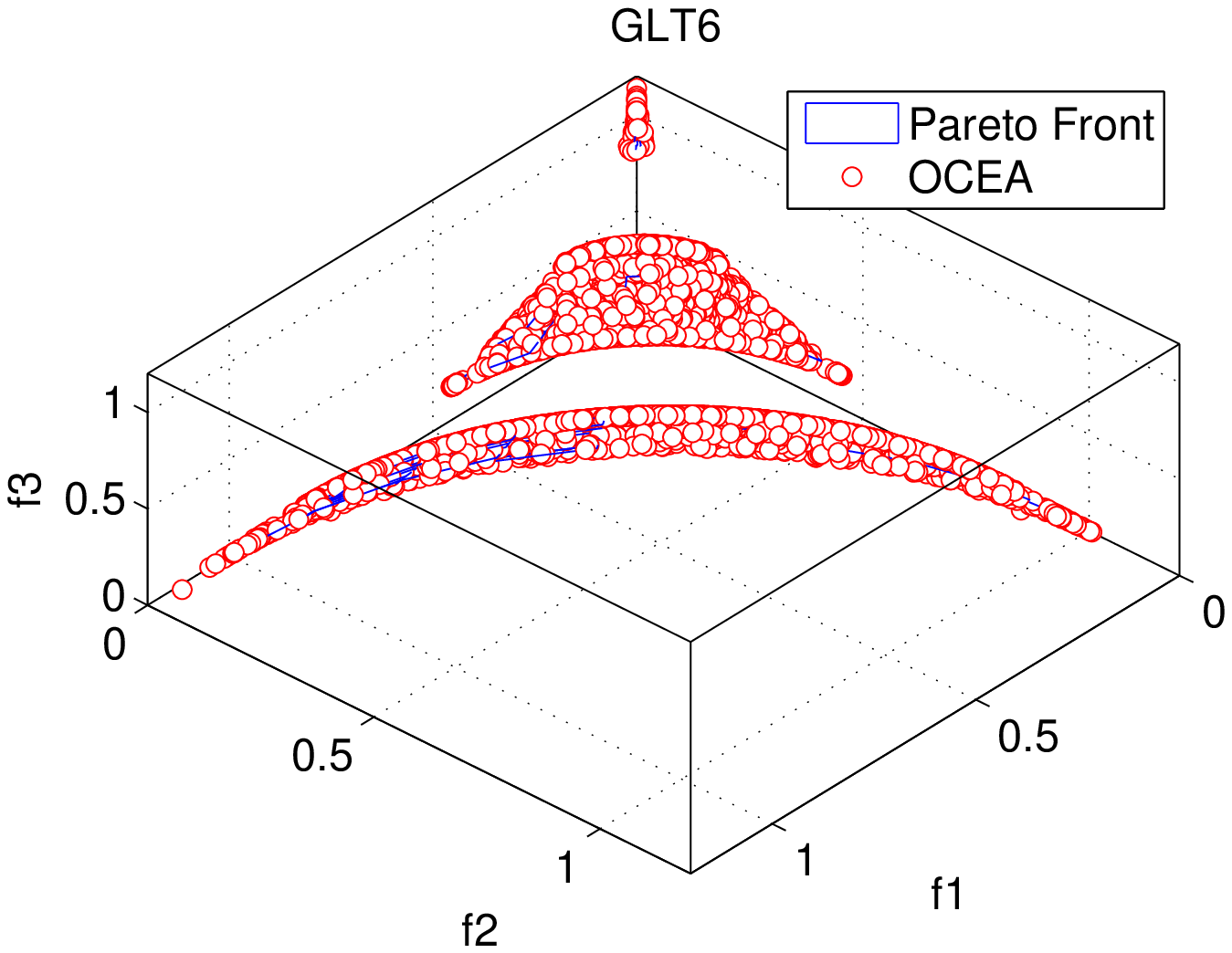}\label{b}}
\subfigure[representative fronts] {\includegraphics[width=0.24\textwidth]{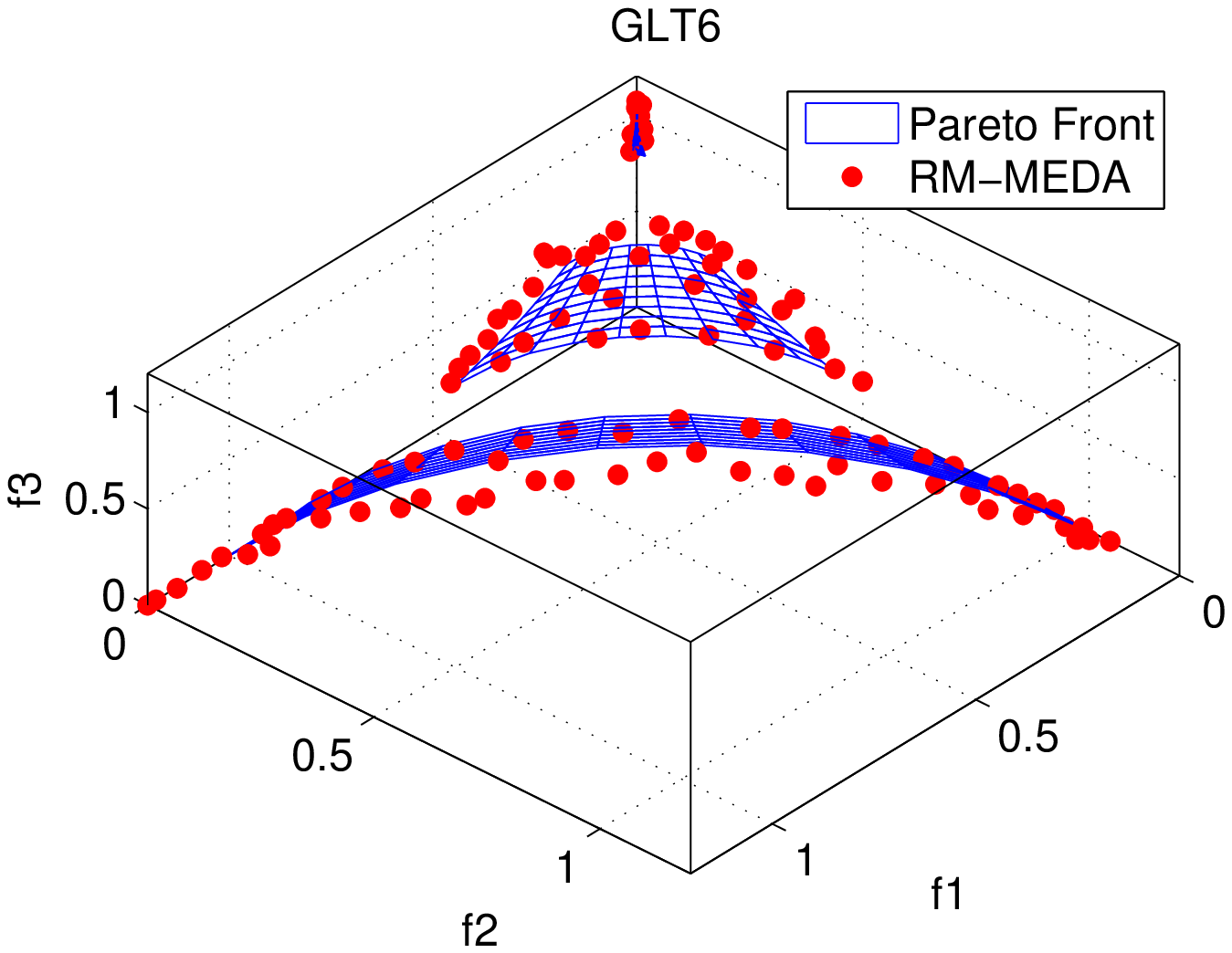}\label{c}}
\subfigure[representative fronts] {\includegraphics[width=0.24\textwidth]{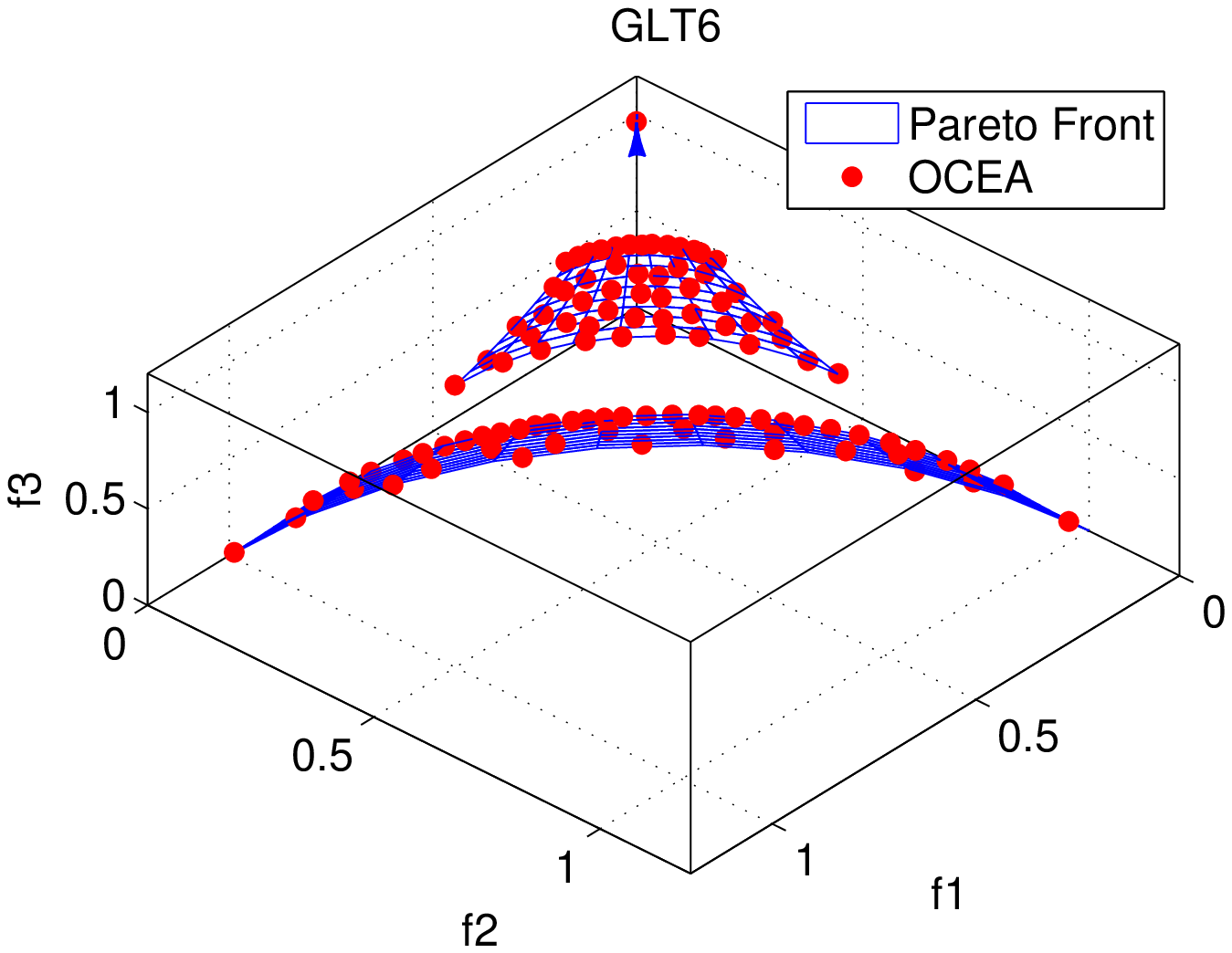}\label{d}}
\caption{Final approximated fronts obtained by OCEA and RM-MEDA}\label{PFcmp}
\end{figure*}

To further investigate the effect of OCEA, Fig.~\ref{PFcmp} plots the final approximated fronts obtained by RM-MEDA and OCEA on GLT1-GLT6. All the final approximated fronts of each instance obtained by RM-MEDA and OCEA, are plotted in Fig.~\ref{a} and~\ref{b}. The final approximated front of each instance with median IGD metric value (called representative front) obtained by RM-MEDA and OCEA, respectively, over 33 independent runs are plotted in Fig.~\ref{c} and~\ref{d}. From Fig.~\ref{a} and \ref{b}, it can be seen that through 33 independent runs, the final approximated fronts of each instance achieved by RM-MEDA and OCEA, respectively, both can cover the whole PF of that instance. However, compared with RM-MEDA, OCEA performs more stably. From Fig.~\ref{c} and \ref{d}, it is observed that the representative fronts of GLT5-GLT6 yielded by RM-MEDA do not reach the PFs. For GLT1-GLT4, although the representative fronts yielded by RM-MEDA all reach the PFs, the PFs are not completely covered. By contrast, the representative fonts obtained by OCEA for each instance all converge to the PFs and distributed well over them. Fig.~\ref{PFcmp} implies that for the GLT test instances, OCEA is stable and robust in terms of convergence and diversity.

In summary, we may conclude that OCEA has shown an excellent performance for dealing with MOPs with complicated PSs and complex PFs.

\section{Further Discussions}\label{furtherDis}

\subsection{Performance on WFG test suite}

To deeply understand the performance of OCEA, OCEA is also applied to the WFG test suite~\cite{WFG} and compared with the five algorithms mentioned above. It is well known that the WFG test instances have complex PFs and are with various complicated characteristics, such as nonseparable, multimodal, degenerate, deceptive, etc. In this section, 9 bi-objective WFG test instances with 30 dimensional decision variables are taken as the test-bed. The maximum evolutionary generation is set as 450. Through preliminary optimisation over parameters, part of the parameter settings of these algorithms are listed in Table~\ref{WFGParameter}; while the rest is the same as in Section~\ref{expSet}. Again 33 independent runs of these algorithms are carried out on each test instance. Table~\ref{WFG_Test} shows the statistics of the IGD and HV metric values obtained by MOEA/D-DE, TMOEA/D, RM-MEDA, NSGA-II, SMS-EMOA and OCEA on the WFG test instances over 33 independent runs.

\begin{table}[htbp]\scriptsize
\centering
\caption{Parameter settings for MOEA/D-DE, TMOEA/D, RM-MEDA, NSGA-II, SMS-EMOA and OCEA on the WFG test suite}\label{WFGParameter}
\begin{tabular}{ll}\toprule
Algs.& Parameters\\\midrule
MOEA/D-DE&$NS=20, \beta=0.9, F=0.5, CR=0.2$\\
TMOEA/D&$F=0.5, CR=0.2$\\
RM-MEDA&$K=5$\\
NSGA-II&$F=0.3, CR=0.2$\\
SMS-EMOA&$F=0.5, CR=0.2$\\
OCEA&$K_{\max}=4, F=0.3, CR=0.6, \beta=0.9$\\
\bottomrule
\end{tabular}
\end{table}

\begin{table*}[htbp]\scriptsize
\centering
\centering \caption{Statistics (mean(std. dev.)[rank]) of the IGD and HV metric values of final approximated fronts obtained by MOEA/D-DE, TMOEA/D, RM-MEDA, NSGA-II, SMS-EMOA and OCEA algorithms over 33 independent runs on the WFG test suite.}\label{WFG_Test}
\begin{tabular}{lcccccc}\toprule
test instances &MOEA/D-DE&TMOEA/D&RM-MEDA&NSGA-II&SMS-EMOA&OCEA\\
\cmidrule{2-7}
&\multicolumn{6}{c}{IGD}\\
\midrule
WFG1&1.105e+00$^\S_{1.15e-02}$[2]&\cellcolor{gray25}\textbf{9.327e-01}$^\S_{1.68e-02}$[1]&1.169e+00$^\S_{7.95e-03}$[3]&1.435e+00$^\dag_{7.17e-02}$[5]&1.446e+00$^\dag_{6.26e-02}$[6]&1.306e+00$_{6.37e-02}$[4]\\
WFG2&3.781e-02$^\dag_{6.18e-04}$[6]&2.860e-02$^\dag_{3.40e-03}$[4]&3.030e-02$^\dag_{4.58e-03}$[5]&1.431e-02$^\approx_{8.48e-04}$[3]&\cellcolor{gray25}\textbf{1.357e-02}$^\S_{8.05e-04}$[1]&1.429e-02$_{8.94e-04}$[2]\\
WFG3&1.456e-02$^\dag_{3.21e-04}$[4]&1.237e-02$^\dag_{5.17e-04}$[2]&3.399e-02$^\dag_{4.46e-03}$[6]&1.869e-02$^\dag_{7.05e-04}$[5]&1.245e-02$^\dag_{2.46e-04}$[3]&\cellcolor{gray25}\textbf{1.184e-02}$_{1.11e-04}$[1]\\
WFG4&\cellcolor{gray25}\textbf{3.400e-02}$^\S_{5.46e-03}$[1]&4.921e-02$^\S_{1.09e-02}$[3]&9.385e-02$^\dag_{2.92e-03}$[6]&5.692e-02$^\approx_{5.76e-03}$[5]&4.581e-02$^\S_{4.01e-03}$[2]&5.512e-02$_{7.60e-03}$[4]\\
WFG5&6.762e-02$^\dag_{2.51e-04}$[3]&7.776e-02$^\dag_{9.13e-03}$[5]&1.116e-01$^\dag_{1.38e-02}$[6]&6.895e-02$^\dag_{3.92e-04}$[4]&\cellcolor{gray25}\textbf{6.656e-02}$^\approx_{7.22e-05}$[1]&6.657e-02$_{7.97e-05}$[2]\\
WFG6&3.325e-01$^\dag_{1.33e-02}$[5]&3.531e-01$^\dag_{2.96e-02}$[6]&3.249e-01$^\dag_{2.54e-03}$[2]&3.321e-01$^\dag_{5.92e-03}$[4]&3.318e-01$^\dag_{1.65e-02}$[3]&\cellcolor{gray25}\textbf{3.249e-01}$_{1.15e-02}$[1]\\
WFG7&1.873e-02$^\dag_{4.30e-04}$[3]&4.113e-02$^\dag_{2.13e-02}$[6]&3.935e-02$^\dag_{7.71e-03}$[5]&2.860e-02$^\dag_{1.54e-03}$[4]&1.192e-02$^\dag_{2.62e-04}$[2]&\cellcolor{gray25}\textbf{1.137e-02}$_{2.45e-04}$[1]\\
WFG8&4.930e-02$^\dag_{7.11e-03}$[3]&4.484e-02$^\approx_{1.73e-02}$[2]&1.628e-01$^\dag_{1.36e-02}$[6]&6.190e-02$^\dag_{8.42e-03}$[5]&5.474e-02$^\dag_{6.73e-03}$[4]&\cellcolor{gray25}\textbf{3.763e-02}$_{1.12e-02}$[1]\\
WFG9&2.451e-01$^\dag_{3.14e-02}$[3]&2.687e-01$^\dag_{2.66e-02}$[5]&\cellcolor{gray25}\textbf{2.073e-01}$^\approx_{2.51e-03}$[1]&2.673e-01$^\dag_{1.95e-02}$[4]&2.703e-01$^\approx_{1.10e-02}$[6]&2.422e-01$_{3.56e-02}$[2]\\
\toprule
&\multicolumn{6}{c}{HV}\\
\cmidrule{2-7}
WFG1&5.942e+00$^\S_{4.52e-02}$[2]&\cellcolor{gray25}\textbf{6.718e+00}$^\S_{8.16e-02}$[1]&5.607e+00$^\S_{4.38e-02}$[3]&4.427e+00$^\dag_{2.63e-01}$[5]&4.398e+00$^\dag_{2.33e-01}$[6]&4.889e+00$_{2.22e-01}$[4]\\
WFG2&1.144e+01$^\S_{1.19e-03}$[2]&\cellcolor{gray25}\textbf{1.145e+01}$^\S_{4.20e-03}$[1]&1.127e+01$^\dag_{3.44e-02}$[6]&1.141e+01$^\dag_{4.67e-03}$[5]&1.143e+01$^\S_{3.66e-03}$[3]&1.142e+01$_{6.53e-03}$[4]\\
WFG3&1.092e+01$^\dag_{3.32e-03}$[4]&1.093e+01$^\dag_{1.26e-02}$[3]&1.077e+01$^\dag_{2.81e-02}$[6]&1.089e+01$^\dag_{4.86e-03}$[5]&1.094e+01$^\dag_{2.98e-03}$[2]&\cellcolor{gray25}\textbf{1.094e+01}$_{1.53e-03}$[1]\\
WFG4&\cellcolor{gray25}\textbf{8.527e+00}$^\S_{3.47e-02}$[1]&8.337e+00$^\approx_{1.23e-01}$[5]&8.107e+00$^\dag_{1.68e-02}$[6]&8.341e+00$^\approx_{7.72e-02}$[4]&8.420e+00$^\S_{2.49e-02}$[2]&8.350e+00$_{5.13e-02}$[3]\\
WFG5&8.140e+00$^\dag_{3.32e-02}$[4]&7.815e+00$^\dag_{2.38e-01}$[6]&7.962e+00$^\dag_{1.02e-01}$[5]&8.144e+00$^\dag_{4.56e-02}$[3]&8.197e+00$^\approx_{6.18e-02}$[2]&\cellcolor{gray25}\textbf{8.199e+00}$_{5.41e-02}$[1]\\
WFG6&6.351e+00$^\dag_{6.78e-02}$[5]&6.245e+00$^\dag_{1.10e-01}$[6]&6.388e+00$^\dag_{1.40e-02}$[2]&6.351e+00$^\dag_{3.08e-02}$[4]&6.355e+00$^\dag_{8.39e-02}$[3]&\cellcolor{gray25}\textbf{6.391e+00}$_{5.88e-02}$[1]\\
WFG7&8.643e+00$^\dag_{3.73e-03}$[3]&7.979e+00$^\dag_{5.89e-01}$[6]&8.459e+00$^\dag_{5.19e-02}$[5]&8.579e+00$^\dag_{8.67e-03}$[4]&8.666e+00$^\dag_{2.52e-03}$[2]&\cellcolor{gray25}\textbf{8.672e+00}$_{1.91e-03}$[1]\\
WFG8&8.450e+00$^\approx_{4.27e-02}$[2]&8.274e+00$^\dag_{2.26e-01}$[5]&7.674e+00$^\dag_{7.98e-02}$[6]&8.329e+00$^\dag_{5.08e-02}$[4]&8.373e+00$^\dag_{4.01e-02}$[3]&\cellcolor{gray25}\textbf{8.463e+00}$_{7.02e-02}$[1]\\
WFG9&6.181e+00$^\dag_{1.61e-01}$[3]&6.038e+00$^\dag_{1.32e-01}$[6]&\cellcolor{gray25}\textbf{6.411e+00}$^\S_{2.17e-02}$[1]&6.119e+00$^\dag_{1.05e-01}$[4]&6.114e+00$^\dag_{5.25e-02}$[5]&6.246e+00$_{1.68e-01}$[2]\\
\midrule
Mean Rank&3.111&4.056&4.444&4.278&3.111&2.000\\
$\dag$/$\S$/$\approx$&12/5/1&12/4/2&14/3/1&15/0/3&11/4/3&\\
\bottomrule
\end{tabular}
\end{table*}

Table~\ref{WFG_Test} shows that OCEA achieves 9 out of the 18 best mean metrics. The rest five algorithms obtain only 9. The performance of these algorithms ranked from the best to the worst is OCEA, MOEA/D-DE, SMS-EMOA, TMOEA/D, NSGA-II and RM-MEDA according to the mean ranks. The Wilcoxon's rank sum test suggests that OCEA performs better than MOEA/D-DE, TMOEA/D, RM-MEDA, NSGA-II and SMS-EMOA in 12, 12, 14, 15 and 11 out of the 18 mean metric values; performs worse in 5, 4, 3, 0 and 4; and similar in 1, 2, 1, 3 and 3. From Table~\ref{WFG_Test}, we may conclude that OCEA performs very well in solving the WFG test instances. It also indicates that OCEA is able to deal with MOPs with complex PFs and with complicated problem characteristics.

\section{Parameter Sensitivity Analysis}\label{senStudy}

The sensitivity of OCEA to its parameters is analysed in this section. The GLT test suite is used for the analysis.

\subsection{Maximum Number of Clusters}

\begin{figure*}[htbp]
\centering
\subfigure[$K_{\max}$] {\includegraphics[width=0.4\textwidth]{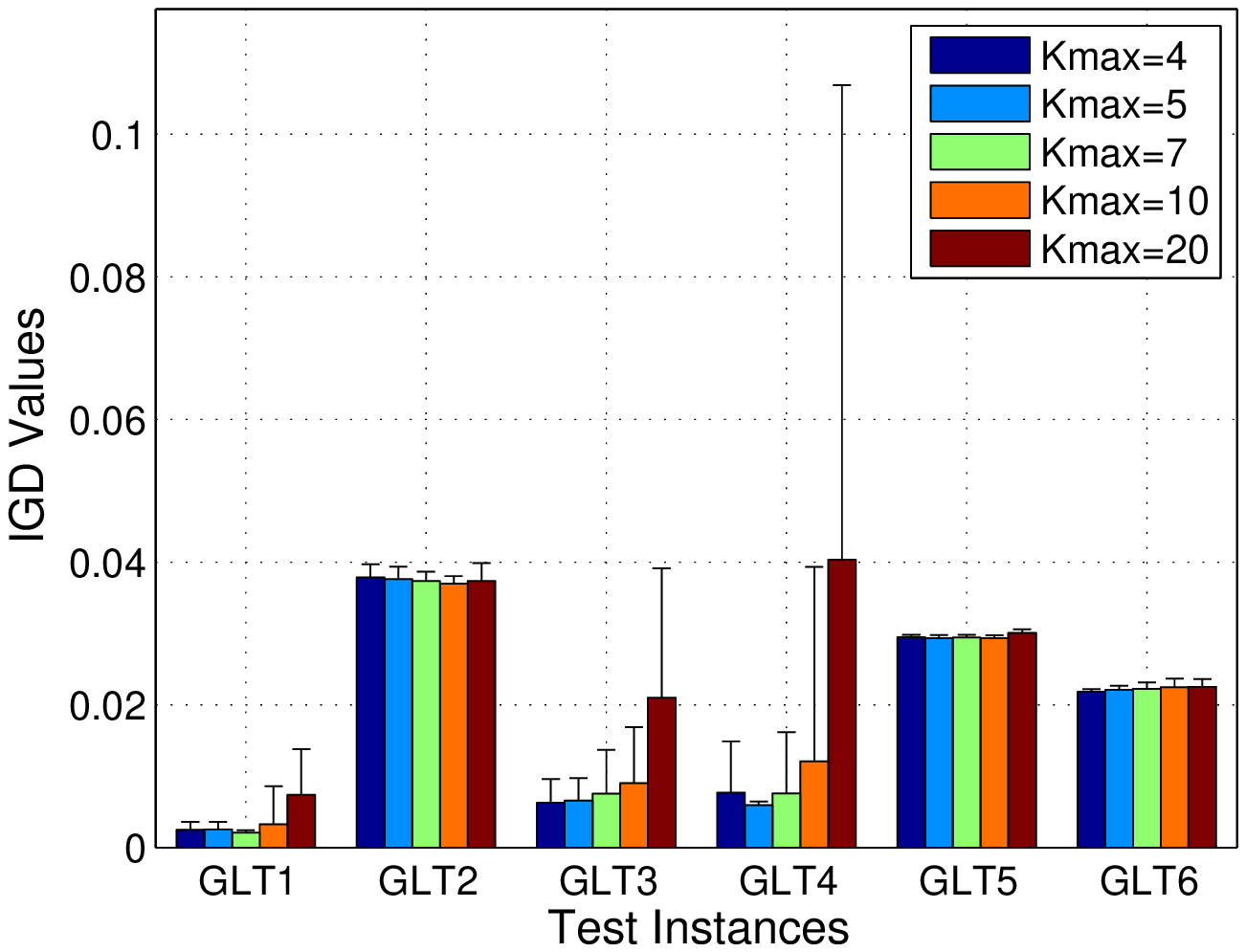} \label{Kmax}}
\subfigure[$\beta$] {\includegraphics[width=0.4\textwidth]{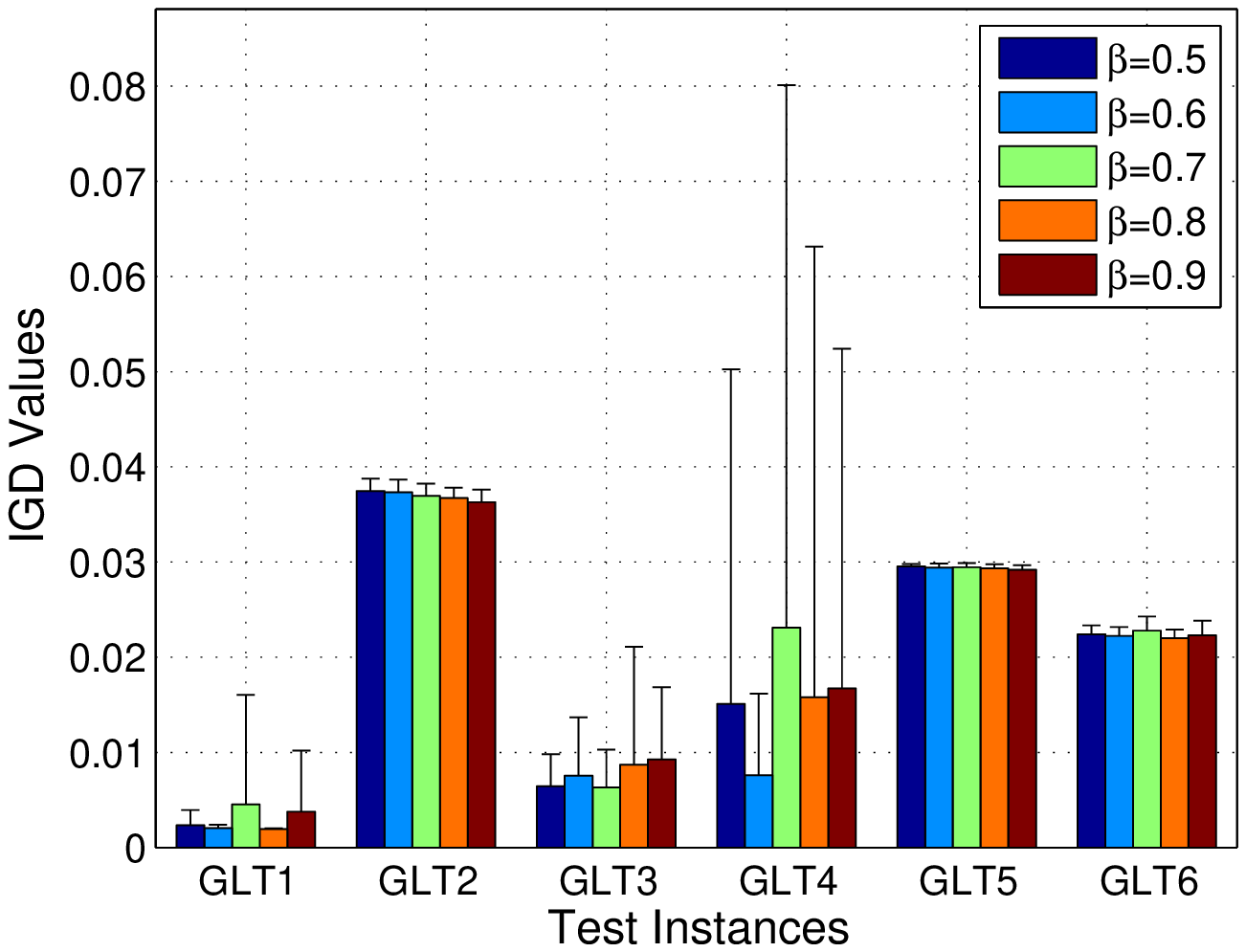} \label{beta}}\\
\subfigure[$F$] {\includegraphics[width=0.4\textwidth]{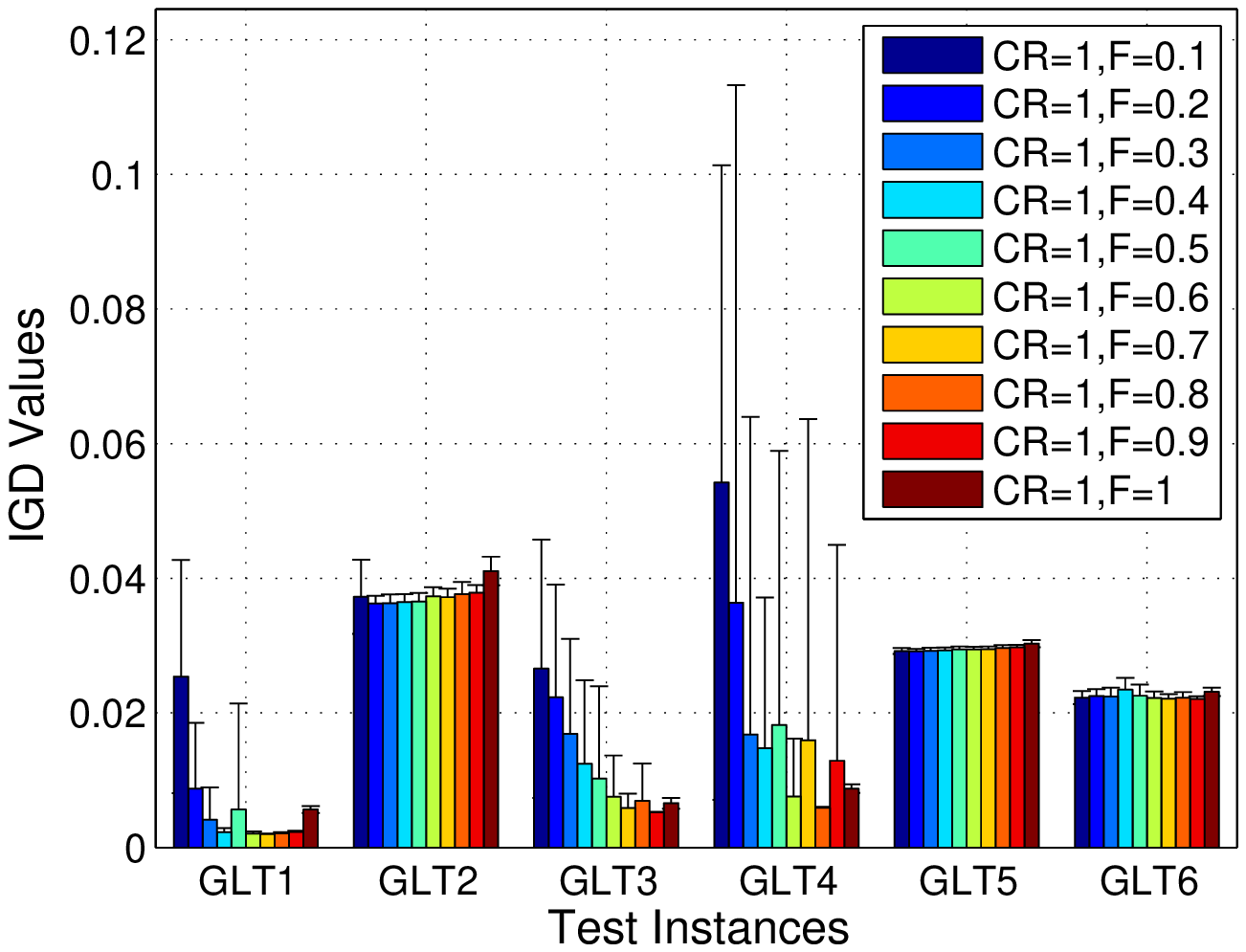} \label{F}}
\subfigure[$CR$] {\includegraphics[width=0.4\textwidth]{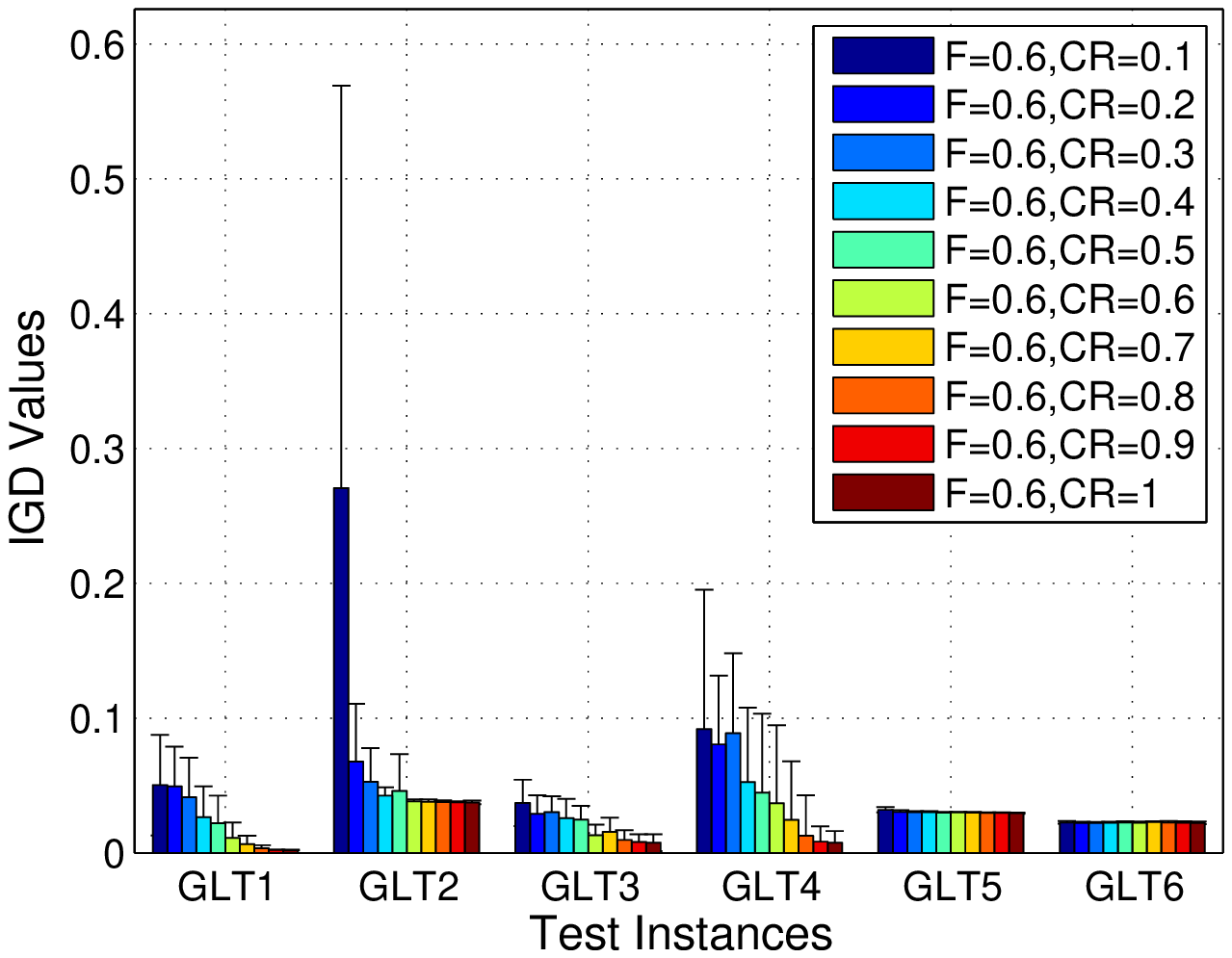} \label{CR}}
\caption{The mean values and standard deviations of the IGD metric values of approximated fronts obtained by OCEA with different $K_{\max}, \beta, F, CR$ values over 22 independent runs on GLT1-GLT6}
\end{figure*}

To test how $K_{\max}$ affects the performance of OCEA, $K_{\max}=\{$4,~5,~7,~10,~20$\}$ are chosen to do analysis. The rest parameters are the same as those in Section~\ref{expSet}. OCEA was run on each test instances independently 22 times with different $K_{\max}$ values. Fig.~\ref{Kmax} shows the mean and standard deviation values of the IGD metric values obtained by OCEA.

From Fig.~\ref{Kmax}, it can be seen that for GLT2, GLT5-GLT6, OCEA can always achieve similar performance robustly for different $K_{\max}$ values. But for GLT1, GLT3-GLT4, different $K_{\max}$ leads to relatively large performance differences. Especially, when $K_{\max}$ is large, the performance of OCEA is not well enough. In general, a small $K_{\max}$ can result in good search results by OCEA on the GLT test instances. This implies that OCEA is not very sensitive to the $K_{\max}$ values on the GLT test instances. Therefore, $K_{\max}=7$ is chosen in Section~\ref{expStu} to carry out the comparison. It should be noted that the optimal $K_{\max}$ depends on individual problem.

\subsection{Clustering Effectiveness Analysis}

\begin{figure*}[htbp]
\centering
\includegraphics[width=0.32\textwidth]{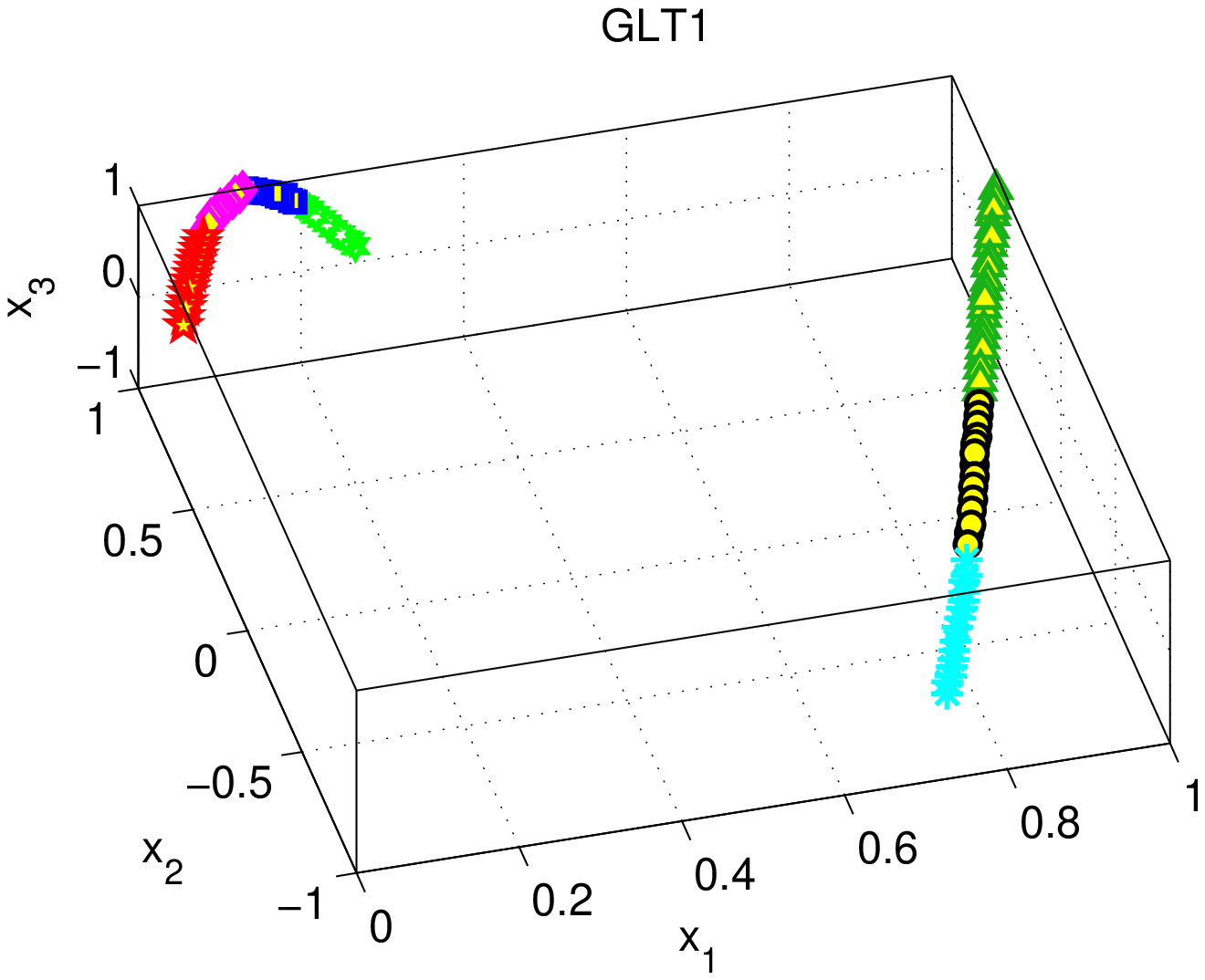}
\includegraphics[width=0.32\textwidth]{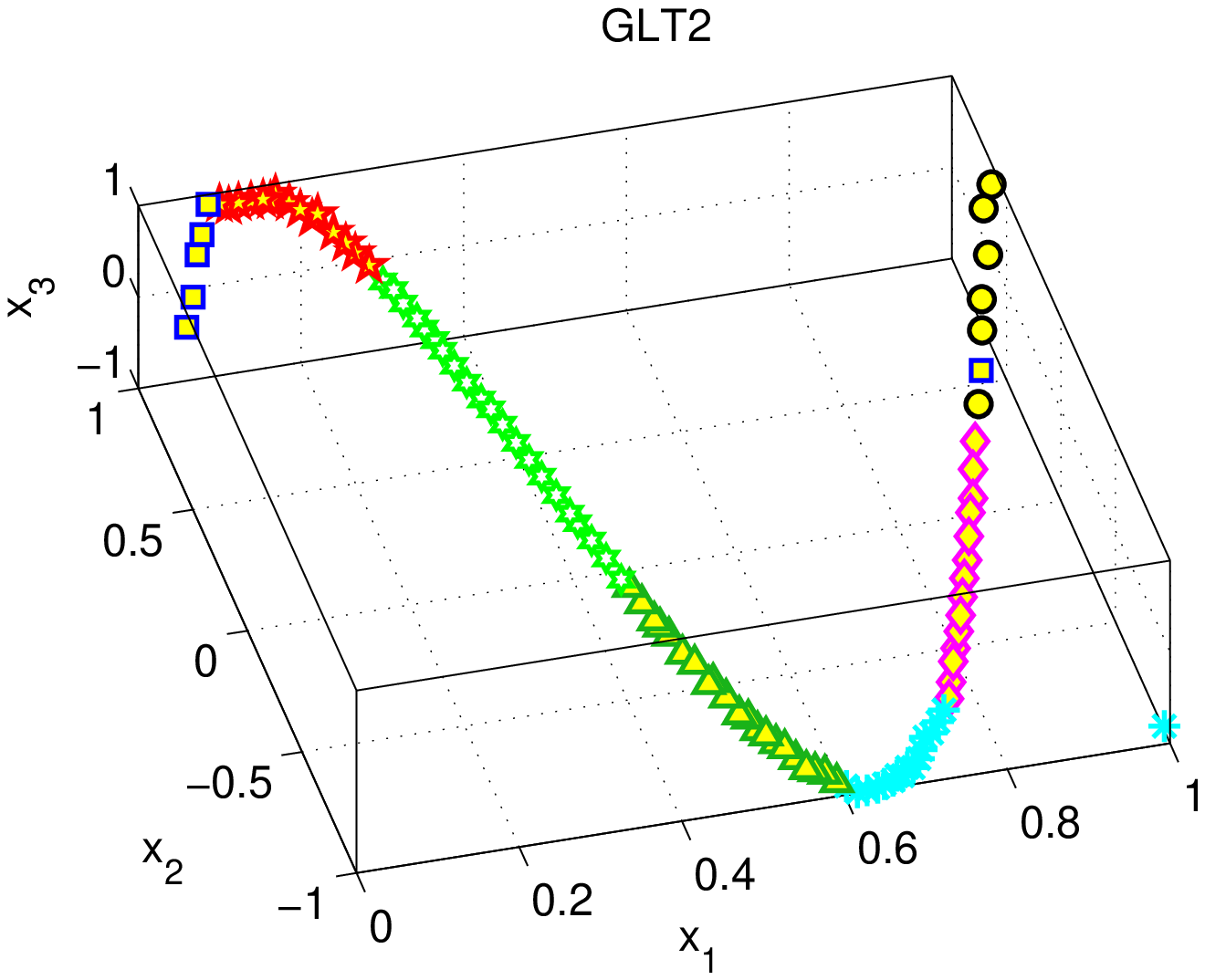}
\includegraphics[width=0.32\textwidth]{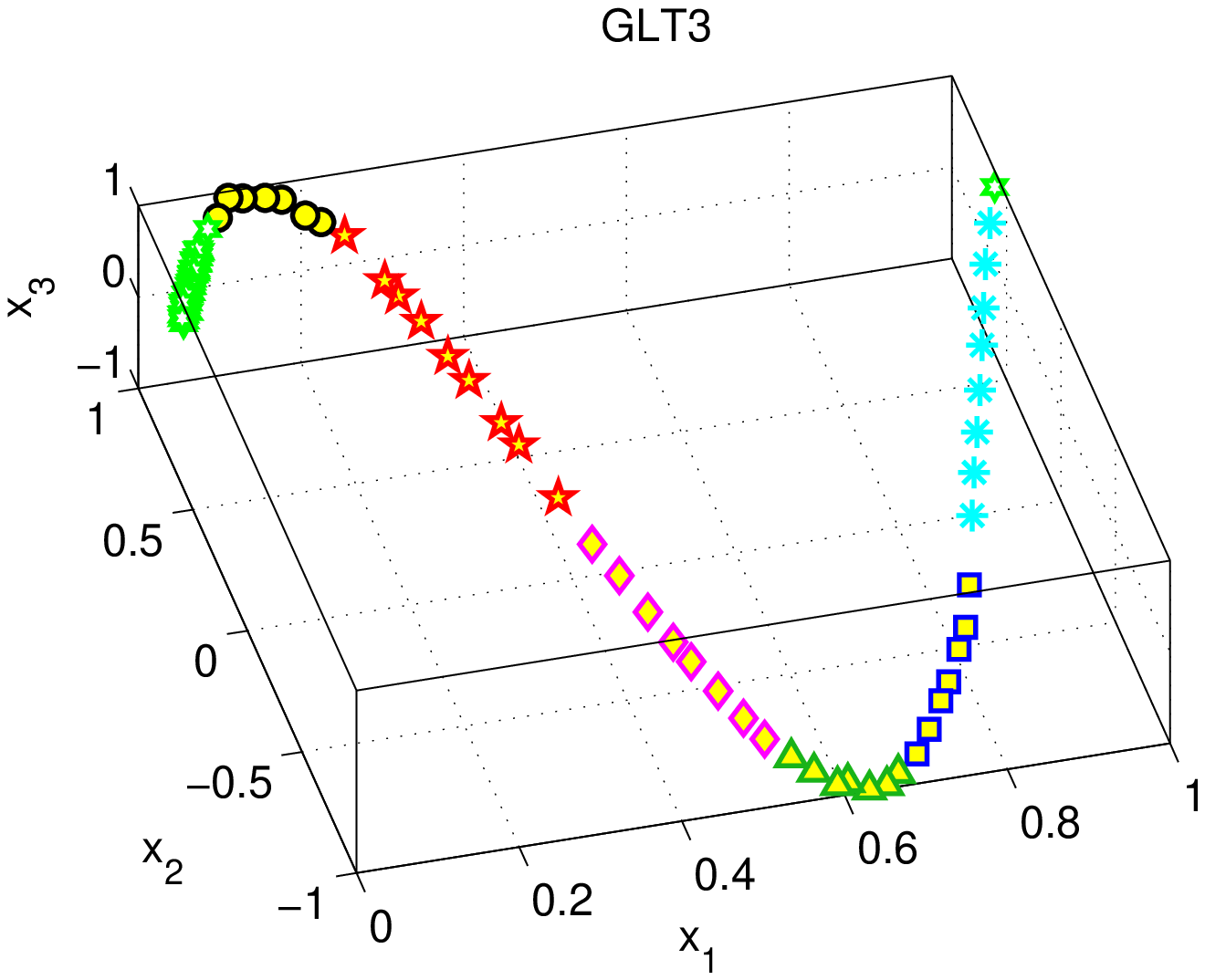}\\
\includegraphics[width=0.32\textwidth]{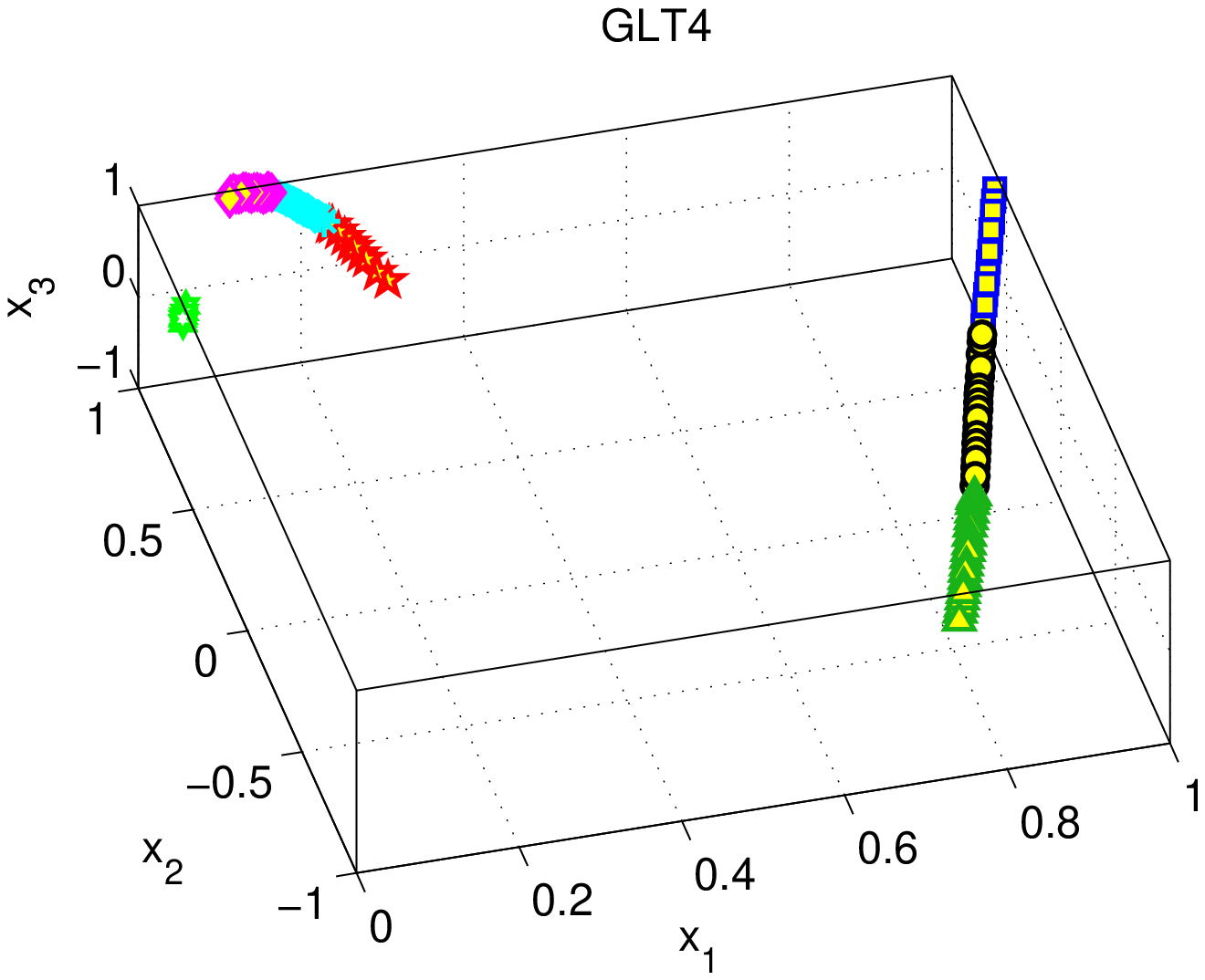}
\includegraphics[width=0.32\textwidth]{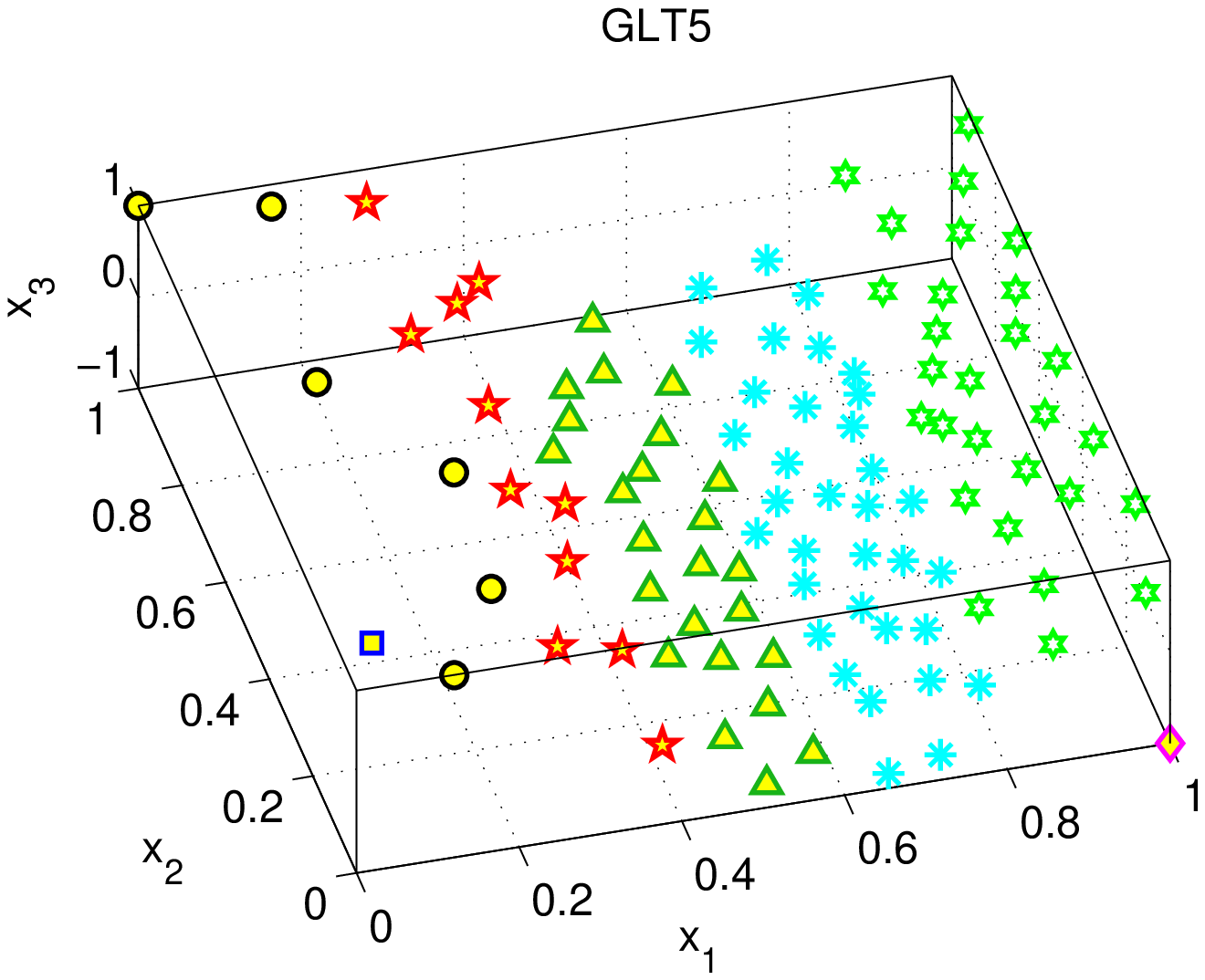}
\includegraphics[width=0.32\textwidth]{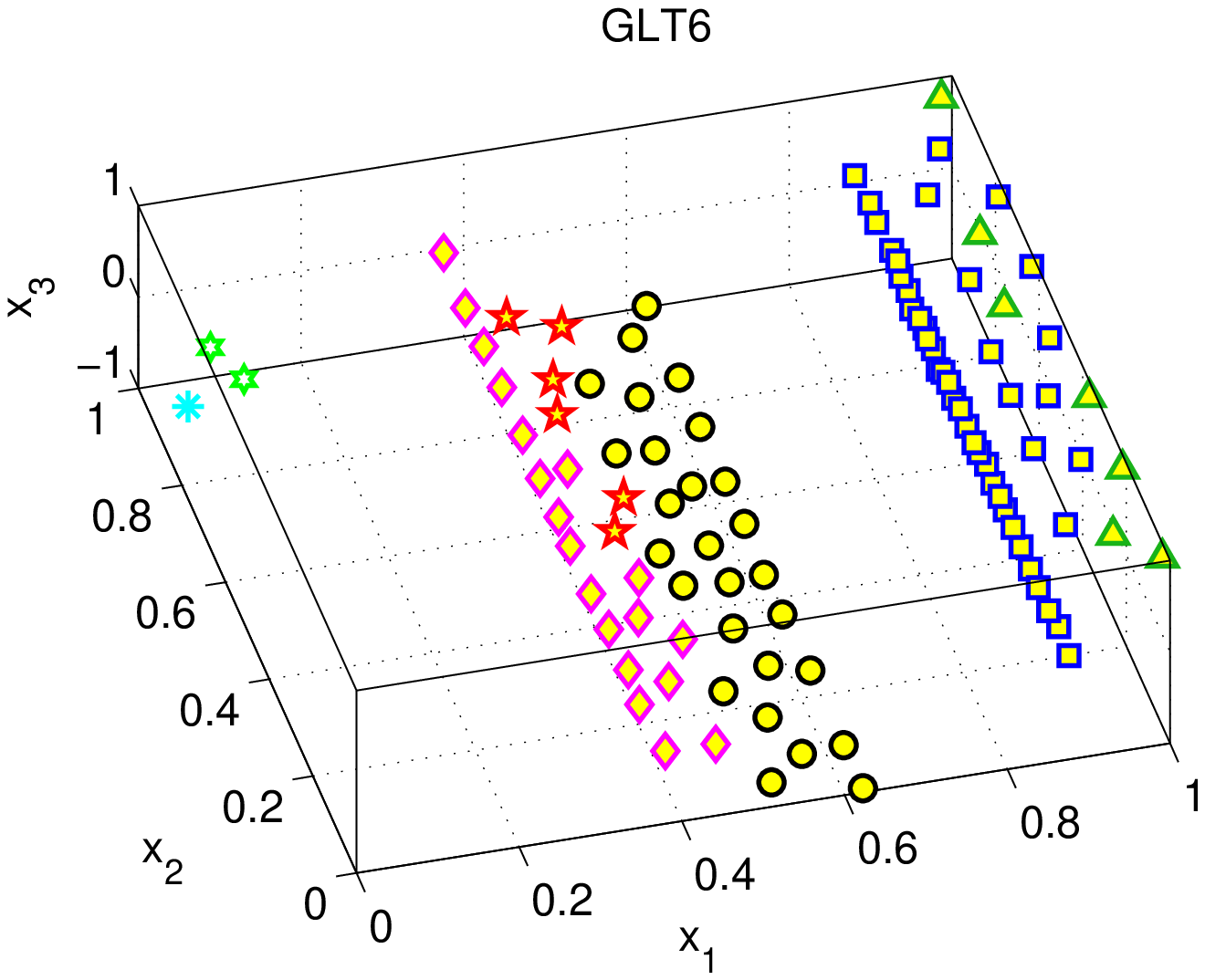}
\caption{The clusters of final approximated sets obtained by OCEA for GLT1-GLT6}\label{clustanalysis}
\end{figure*}

The evolution procedure couples naturally with the online clustering procedure in OCEA. It is expected that the approximated set will present a clustering effect when the evolution procedure has converged. To justify the effectiveness of the online clustering, Fig.~\ref{clustanalysis} plots the clustering results in the first 3-dimensional search space on the GLT1-GLT6 test instances. In the figure, the solutions in each different cluster are marked with different colors and symbols. It can be seen that the final approximated sets are partitioned into 7 clusters clearly (note that $K_{\max}$ is set as 7). This figure indicates that OCEA can indeed approximate the clustering structure effectively.

\subsection{Mating Restriction Probability}

To test the sensitivity of the OCEA's performance to the mating control parameter $\beta$, $\beta=\{$0.5,~0.6,~0.7,~0.8,~0.9$\}$ are used for the analysis. The rest parameters are the same as those in Section~\ref{expSet}. Again, for different $\beta$ value, OCEA independently run 22 time on the test instances. Fig.~\ref{beta} shows the statistics of the obtained IGD metric values.

From Fig.~\ref{beta}, it is observable that for GLT5 and GLT6, different $\beta$ values bring a similar performance for OCEA; but for GLT1-GLT4, OCEA with different $\beta$ values performs very differently. Nevertheless, when $\beta=0.6$, OCEA has relatively better performance for all the instances. The observation in Fig.~\ref{beta} indicates that OCEA is not so sensitive to the setting of $\beta$ in solving the GLT test instances. Therefore, $\beta=0.6$ is chosen in Section~\ref{expStu} for the controlled comparison experiments. Again, it is necessary to point out that an optimal $\beta$ setting should depend on the problem characteristics.

\subsection{Control Parameters of Differential Evolution Operator}

The effect of the DE parameters, i.e., $F$ and $CR$, are to be evaluated in this section. $F~(CR)=\{$0.1, 0.2, 0.3, 0.4, 0.5, 0.6, 0.7, 0.8, 0.9, 1$\}$ are chosen to proceed analysis. The rest parameters are the same as in Section~\ref{expSet}. When different $F$ ($CR$) values are set, $CR$ ($F$) is set as 1 (0.6). The mean and standard deviation values of the IGD metric values obtained by OCEA with different $F$ or $CR$ over 22 independent runs are shown in Figs.~\ref{F} and \ref{CR}.

Fig.~\ref{F} shows that the $F$ value has a crucial effect on the OCEA performance for GLT1, GLT3-GLT4, and a large $F$ value can lead to a good OCEA performance. However, for GLT2, GLT5-GLT6, different $F$ settings do not affect the OCEA performance acutely. Fig.~\ref{CR} shows that the $CR$ value has a significant effect on OCEA for GLT1-GLT4, and a small $CR$ value is better. But OCEA performs rather stably for GLT5-GLT6 with different $CR$ values. In case $F=\{0.6,~0.8\}$ ($CR=1$) and $CR=\{0.9,~1\}$ ($F=0.6$), OCEA can always find good IGD metric values for all the GLT test instances. In general, Figs.~\ref{F} and \ref{CR} denote that OCEA is not very sensitive to the $F$ and $CR$ settings.

\section{Conclusion}\label{conclusion}

This paper presented a first-ever MOEA that incorporate an {\em online clustering} to address the non-stationary nature of the evolutionary search. The underlying consideration is 1) to learn the manifold structure of the PS (i.e. the so-called regularity property of MOPs) through clustering; and 2) to adapt to the non-stationary search dynamics. The online agglomerative clustering approach developed in~\cite{guedalia1999line} is modified to accommodate the evolution search dynamics. Experimental study has shown that the online clustering can address the non-stationary search process well, and is able to adaptively learn the clustering structure of the PS. The comparison against five well-known MOEAs has also shown that the structures learned adaptively by the online clustering can indeed improve the search efficiency (in terms of search speed) and effectiveness (in terms of the quality of the final approximated sets and fronts). Future work includes 1) the development of intelligent recombination operators that can be well fitted in the online learning mechanism; 2) the development and/or incorporation of other online learning strategies; and 3) the study of the developed framework for many-objective optimisation problems.

%\bibliography{reference}

\begin{thebibliography}{10}
\providecommand{\url}[1]{#1}
\csname url@samestyle\endcsname
\providecommand{\newblock}{\relax}
\providecommand{\bibinfo}[2]{#2}
\providecommand{\BIBentrySTDinterwordspacing}{\spaceskip=0pt\relax}
\providecommand{\BIBentryALTinterwordstretchfactor}{4}
\providecommand{\BIBentryALTinterwordspacing}{\spaceskip=\fontdimen2\font plus
\BIBentryALTinterwordstretchfactor\fontdimen3\font minus
  \fontdimen4\font\relax}
\providecommand{\BIBforeignlanguage}[2]{{%
\expandafter\ifx\csname l@#1\endcsname\relax
\typeout{** WARNING: IEEEtran.bst: No hyphenation pattern has been}%
\typeout{** loaded for the language `#1'. Using the pattern for}%
\typeout{** the default language instead.}%
\else
\language=\csname l@#1\endcsname
\fi
#2}}
\providecommand{\BIBdecl}{\relax}
\BIBdecl

\bibitem{1999Miettinen}
K.~Miettinen, \emph{Nonlinear Multiobjective Optimization}.\hskip 1em plus
  0.5em minus 0.4em\relax Boston, USA: Kluwer Academic Publishers, 1999.

\bibitem{2001Deb}
K.~Deb, \emph{Multi-Objective Optimization Using Evolutionary
  Algorithms}.\hskip 1em plus 0.5em minus 0.4em\relax New York, USA: John Wiley
  {\&} Sons, 2001.

\bibitem{2011Zhou}
A.~Zhou, B.-Y. Qu, H.~Li, S.-Z. Zhao, P.~N. Suganthan, and Q.~Zhang,
  ``Multiobjective evolutionary algorithms: A survey of the state of the art,''
  \emph{Swarm and Evolutionary Computation}, vol.~1, no.~1, pp. 32--49, 2011.

\bibitem{zhang2015self}
H.~Zhang, S.~Song, and X.~Z. Gao, ``Self-organizing multiobjective optimization
  based on decomposition with neighborhood ensemble,'' \emph{Neurocomputing},
  2015.

\bibitem{yu2010introduction}
X.~Yu and M.~Gen, \emph{Introduction to Evolutionary Algorithms}.\hskip 1em
  plus 0.5em minus 0.4em\relax London, UK: Springer-Verlag, 2010.

\bibitem{2008ZhangZJ}
Q.~Zhang, A.~Zhou, and Y.~Jin, ``{RM-MEDA}: A regularity model based
  multiobjective estimation of distribution algorithm,'' \emph{IEEE
  Transactions on Evolutionary Computation}, vol.~12, no.~1, pp. 41--63, 2008.

\bibitem{regularity}
C.~Hillermeier, \emph{Nonlinear Multiobjective Optimization-A Generalized
  Homotopy Approach}.\hskip 1em plus 0.5em minus 0.4em\relax Basel,
  Switzerland: Birkh\"{a}user Verlag, 2001.

\bibitem{wang2012regularity}
Y.~Wang, J.~Xiang, and Z.~Cai, ``A regularity model-based multiobjective
  estimation of distribution algorithm with reducing redundant cluster
  operator,'' \emph{Applied Soft Computing}, vol.~12, no.~11, pp. 3526--3538,
  2012.

\bibitem{guedalia1999line}
I.~D. Guedalia, M.~London, and M.~Werman, ``An on-line agglomerative clustering
  method for nonstationary data,'' \emph{Neural computation}, vol.~11, no.~2,
  pp. 521--540, 1999.

\bibitem{2002DebPAM}
K.~Deb, A.~Pratap, S.~Agarwal, and T.~Meyarivan, ``A fast and elitist
  multiobjective genetic algorithm: {NSGA-II},'' \emph{IEEE Transactions on
  Evolutionary Computation}, vol.~6, no.~2, pp. 182--197, 2002.

\bibitem{zitzler2001spea2}
E.~Zitzler, M.~Laumanns, and L.~Thiele, ``{SPEA2}: Improving the strength
  pareto evolutionary algorithm,'' Swiss Federal Institute Technology, Zurich,
  Switzerland, Tech. Rep., 2001.

\bibitem{corne2001pesa}
D.~W. Corne, N.~R. Jerram, J.~D. Knowles, M.~J. Oates, and J.~Martin,
  ``{PESA-II}: Region-based selection in evolutionary multiobjective
  optimization,'' in \emph{Proceedings of the 2nd Annual Genetic and
  Evolutionary Computation Conference}.\hskip 1em plus 0.5em minus 0.4em\relax
  San Francisco, California, USA: Morgan Kaufmann Publishers, 2001, pp.
  283--290.

\bibitem{deb2014evolutionary}
K.~Deb and H.~Jain, ``An evolutionary many-objective optimization algorithm
  using reference-point-based nondominated sorting approach, part i: Solving
  problems with box constraints,'' \emph{IEEE Transactions on Evolutionary
  Computation}, vol.~18, no.~4, pp. 577--601, 2014.

\bibitem{beume2007sms}
N.~Beume, B.~Naujoks, and M.~Emmerich, ``{SMS-EMOA:} multiobjective selection
  based on dominated hypervolume,'' \emph{European Journal of Operational
  Research}, vol. 181, no.~3, pp. 1653--1669, 2007.

\bibitem{2011BaderZ}
J.~Bader and E.~Zitzler, ``{HypE}: An algorithm for fast hypervolume-based
  many-objective optimization,'' \emph{Evolutionary Computation}, vol.~19,
  no.~1, pp. 45--76, 2011.

\bibitem{phan2013r2}
D.~Phan and J.~Suzuki, ``{R2-IBEA}: R2 indicator based evolutionary algorithm
  for multiobjective optimization,'' in \emph{Proceedings of the 2013 IEEE
  Congress on Evolutionary Computation}.\hskip 1em plus 0.5em minus 0.4em\relax
  New York, USA: IEEE, 2013, pp. 1836--1845.

\bibitem{rodriguez2012new}
C.~A. Rodr{\'\i}guez~Villalobos and C.~A. Coello~Coello, ``A new
  multi-objective evolutionary algorithm based on a performance assessment
  indicator,'' in \emph{Proceedings of the 14th Annual Genetic and Evolutionary
  Computation Conference}.\hskip 1em plus 0.5em minus 0.4em\relax New York,
  USA: ACM, 2012, pp. 505--512.

\bibitem{2009LiZ}
H.~Li and Q.~Zhang, ``Multiobjective optimization problems with complicated
  {Pareto} sets, {MOEA/D} and {NSGA-II},'' \emph{IEEE Transactions on
  Evolutionary Computation}, vol.~13, no.~2, pp. 284--302, 2009.

\bibitem{Liu2013}
H.-L. Liu, F.~Gu, and Q.~Zhang, ``Decomposition of a multiobjective
  optimization problem into a number of simple multiobjective subproblems,''
  \emph{IEEE Transactions on Evolutionary Computation}, vol.~18, no.~3, pp.
  450--455, 2014.

\bibitem{2007ZhangL}
Q.~Zhang and H.~Li, ``{MOEA/D}: A multiobjective evolutionary algorithm based
  on decomposition,'' \emph{IEEE Transactions on Evolutionary Computation},
  vol.~11, no.~6, pp. 712--731, 2007.

\bibitem{li2014stable}
K.~Li, Q.~Zhang, S.~Kwong, M.~Li, and R.~Wang, ``Stable matching-based
  selection in evolutionary multiobjective optimization,'' \emph{IEEE
  Transactions on Evolutionary Computation}, vol.~18, no.~6, pp. 909--923,
  2014.

\bibitem{zhou2009approximating}
A.~Zhou, Q.~Zhang, and Y.~Jin, ``Approximating the set of pareto-optimal
  solutions in both the decision and objective spaces by an estimation of
  distribution algorithm,'' \emph{IEEE Transactions on Evolutionary
  Computation}, vol.~13, no.~5, pp. 1167--1189, 2009.

\bibitem{li2013improved}
Y.~Li, X.~Xu, P.~Li, and L.~Jiao, ``Improved {RM-MEDA} with local learning,''
  \emph{Soft Computing}, vol.~18, no.~7, pp. 1--15, 2013.

\bibitem{Li2014learning}
K.~Li and S.~Kwong, ``A general framework for evolutionary multiobjective
  optimization via manifold learning,'' \emph{Neurocomputing}, vol. 146, pp.
  65--74, 2014.

\bibitem{bacharoglou10}
A.~Bacharoglou, ``Approximation of probability distributions by convex mixtures
  of gaussian measures,'' \emph{Proceedings of the American Mathematical
  Society}, vol. 138, no.~7, pp. 2619--2628, 2010.

\bibitem{price2006differential}
K.~Price, R.~M. Storn, and J.~A. Lampinen, \emph{Differential Evolution: A
  Practical Approach to Global Optimization}.\hskip 1em plus 0.5em minus
  0.4em\relax Heidelberg, Berlin, Germany: Springer-Verlag, 2006.

\bibitem{XQiu2015}
X.~Qiu, J.~Xu, K.~Tan, and H.~Abbass, ``Adaptive cross-generation differential
  evolution operators for multi-objective optimization,'' \emph{IEEE
  Transactions on Evolutionary Computation}, vol.~PP, no.~99, pp. 1--1, 2015.

\bibitem{zitzler2003performance}
E.~Zitzler, L.~Thiele, M.~Laumanns, C.~M. Fonseca, and V.~G. Da~Fonseca,
  ``Performance assessment of multiobjective optimizers: An analysis and
  review,'' \emph{IEEE Transactions on Evolutionary Computation}, vol.~7,
  no.~2, pp. 117--132, 2003.

\bibitem{liu2010t}
H.-L. Liu, F.~Gu, and Y.~Cheung, ``{T-MOEA/D}: {MOEA/D} with objective
  transform in multi-objective problems,'' in \emph{Proceedings of the 2010
  International Conference of Information Science and Management Engineering},
  vol.~2.\hskip 1em plus 0.5em minus 0.4em\relax New York, USA: IEEE, 2010, pp.
  282--285.

\bibitem{zitzler1999multiobjective}
E.~Zitzler and L.~Thiele, ``Multiobjective evolutionary algorithms: A
  comparative case study and the strength pareto approach,'' \emph{IEEE
  Transactions on Evolutionary Computation}, vol.~3, no.~4, pp. 257--271, 1999.

\bibitem{WFG}
S.~Huband, L.~Barone, L.~While, and P.~Hingston, ``A scalable multi-objective
  test problem toolkit,'' in \emph{Proceedings of the 3rd International
  Conference on Evolutionary Multi-Criterion Optimization}.\hskip 1em plus
  0.5em minus 0.4em\relax Heidelberg, Berlin, Germany: Springer-Verlag, 2005,
  pp. 280--295.

\end{thebibliography}

\end{document}